\documentclass{article} %
\usepackage{car_preprint,times}
\usepackage[T1]{fontenc}

\usepackage{amsmath,amsfonts,bm}

\def\eqref#1{equation~\ref{#1}}

\def\1{\bm{1}}

\DeclareMathAlphabet{\mathsfit}{\encodingdefault}{\sfdefault}{m}{sl}
\SetMathAlphabet{\mathsfit}{bold}{\encodingdefault}{\sfdefault}{bx}{n}

\usepackage{hyperref}
\usepackage{url}

\usepackage{caption}
\usepackage{subcaption}                   \usepackage{algorithm}

\usepackage{graphicx}
\usepackage{multirow}
\usepackage[table]{xcolor}
\usepackage{colortbl}
\usepackage{booktabs}
\usepackage{array}
\usepackage{tabularx}
\usepackage{amsmath}
\usepackage{amssymb}
\usepackage{algpseudocode}
\usepackage{pifont}
\newcommand{\rnum}[1]{\ding{\the\numexpr171+#1\relax}}

\definecolor{carrow}{RGB}{234,242,251}    %
\definecolor{carbar}{RGB}{128,174,231}    %
\definecolor{basebar}{RGB}{192,198,208}   %
\definecolor{refrow}{gray}{0.965}         %
\definecolor{budgetrow}{gray}{0.915}      %
\definecolor{venuec}{gray}{0.42}          %
\newlength{\relw}                         %
\newlength{\relnumw}                      %

\newtheorem{proposition}{Proposition}   %

\title{CAR: Reducing Decoder Output Shift in Visual Token Pruning via Coverage
Selection and Attention Reallocation}

\author{%
\textbf{Xixi Jiang\textsuperscript{1}, Pingcheng Dong\textsuperscript{1}, Chen Yang\textsuperscript{1}, Xin Yang\textsuperscript{2}, Kwang-Ting Cheng\textsuperscript{1}} \\[0.5em]
{\normalfont\small \textsuperscript{1}The Hong Kong University of Science and Technology} \\
{\normalfont\small \textsuperscript{2}Huazhong University of Science and Technology}
}
\date{}

\hypersetup{%
  hidelinks,
  pdftitle={CAR: Reducing Decoder Output Shift in Visual Token Pruning via Coverage Selection and Attention Reallocation},
  pdfauthor={Xixi Jiang, Pingcheng Dong, Chen Yang, Xin Yang, Kwang-Ting Cheng},
  pdfsubject={Visual token pruning},
  pdfkeywords={visual token pruning, vision-language models, coverage selection, attention reallocation}
}

\begin{document}

\maketitle

\begin{abstract}

    Visual token pruning is an effective way to accelerate multimodal large language models
    (MLLMs). However, existing methods degrade substantially under aggressive pruning, especially
    on visual grounding (VG). We analyze the sources of this degradation from the perspective of the output shift
    that pruning causes in decoder layers, and decompose this shift in closed form into two terms, an image summary error and an attention
    allocation error. The summary error stems from the loss of visual information, and the allocation
    error from the drop in the share of attention assigned to visual tokens. To reduce both errors, we propose CAR (Coverage selection and Attention
    Reallocation), a token pruning framework that restores attention share while
    maintaining information. Specifically, to minimize the summary error, whose upper bound tightens as the retained
    tokens better cover all visual tokens, our coverage selection retains tokens that are both
    salient and non-redundant, with saliency from encoder heads chosen by a label-free probe. To
    minimize the allocation error, our attention reallocation reassigns the attention of the
    discarded tokens to the retained ones through a closed-form logit bias. On Qwen2.5-VL-7B with $90\%$
    of the visual tokens pruned, CAR retains $93.0\%$ of the unpruned accuracy on visual question answering (VQA) and $77.9\%$ on VG, against $26.5\%$ for the best existing method on VG; the logit
    bias alone raises the VG accuracy of existing methods by up to $24$ points.
    Experiments on multiple MLLMs across VQA, VG and video understanding show that CAR
    generalizes well and consistently outperforms existing methods.
    \end{abstract}

\section{Introduction}
\label{Sec:1}

Multimodal large language models
(MLLMs)~\citep{liu2024llava1.5, Bai2025qwen2.5-vl, zhu2025internvl3} have achieved strong
performance on vision-language tasks such as visual question
answering~\citep{singh2019towards, liu2024mmbench}, visual
grounding~\citep{yu2016modeling, mao2016generation} and video
understanding~\citep{fu2025video, li2024mvbench}. However, modern vision encoders represent a
single image as hundreds to thousands of patch
tokens~\citep{liu2024llavanext, chen2024internvl2}, which dominate prefilling at a quadratic
attention cost~\citep{chen2024fastv, xing2024pdrop} and hinder deployment under real-time or
resource constraints. Visual token pruning therefore reduces this cost by retaining only a subset of the visual tokens while maximally preserving performance.

\begin{figure}[!t]
  \centering
  \includegraphics[width=\linewidth]{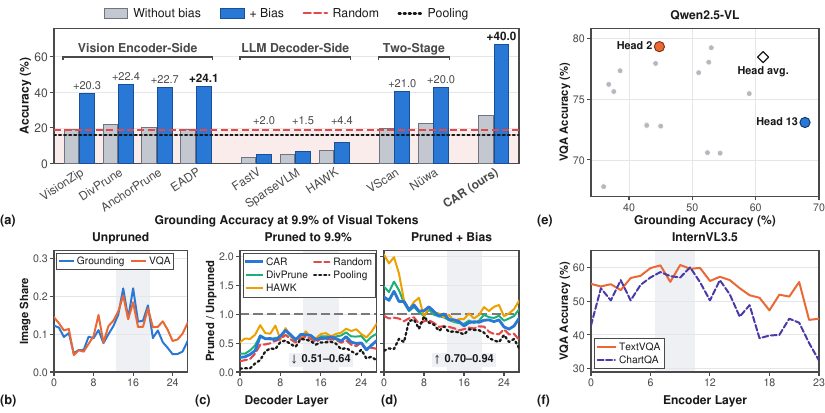}
  \vspace{-5mm}
  \caption{
  \textbf{(a)} Grounding accuracy of existing methods and two baselines (random
  sampling, pooling) on Qwen2.5-VL-7B at $9.9\%$ retention, averaged over eight
  RefCOCO/+/g splits, without and with the uniform bias $\log(N/k)$.
  \textbf{(b)} Visual attention share, the share of the text tokens' attention allocated to the image,
  at each decoder layer in the unpruned model.
  \textbf{(c, d)} Visual attention share after pruning to $9.9\%$, divided by its unpruned value on
  the same image, without (c) and with (d) the bias. The grey band in (b)--(d) marks layers $13$ to $19$.
   \textbf{(e)} Grounding and VQA accuracy with saliency from each head of the last
  encoder layer of Qwen2.5-VL-7B. \textbf{(f)} VQA accuracy on TextVQA and ChartQA with
  saliency from each encoder layer of InternVL3.5-8B. Setup
  in Appendix~\ref{sec:appendix:motivation}.}
  \label{fig:phenomena}
\end{figure}

We categorize mainstream visual token pruning methods into three types based on their
operational stage: Vision Encoder-Side
Pruning~\citep{yang2025visionzip, alvar2025divprune, zhangcdpruner, liu2026hiprune,
oh2026anchorprune, wang2026eadp}, LLM Decoder-Side
Pruning~\citep{chen2024fastv, zhang2024sparsevlm, xing2024pdrop, wen2025dart, zhu2026hawk}, and
Two-Stage Pruning~\citep{zhang2025vscan, huang2026nuwa}. These methods maintain their performance on
Visual Question Answering (VQA) but degrade sharply on Visual Grounding (VG) under aggressive
pruning~\citep{huang2026nuwa}. Figure~\ref{fig:phenomena}(a) compares these pruning methods with two simple baselines, random sampling and average pooling, on the eight splits of RefCOCO, RefCOCO+ and RefCOCOg~\citep{yu2016modeling, mao2016generation}. Counterintuitively, these methods hold only a small advantage over the simple baselines. This observation raises a question: \emph{Why do existing pruning methods exhibit such significant performance degradation?}

 We find that this performance degradation stems not only from the loss of information carried by
  discarded tokens, but also from a drop in the visual attention share, the share of attention that
  the text allocates to the image. In the unpruned model, the visual attention share is highest in
  the intermediate layers (Figure~\ref{fig:phenomena}(b)), consistent with the finding that these
  layers transfer question-relevant visual information into the text tokens~\citep{zhang2025crossmodalflow}.
  Since the softmax normalizes over all tokens, pruning removes the attention that the discarded
  visual tokens received, while the text tokens keep theirs. As a result, the visual attention share
  in these layers falls to only $51$--$64\%$ of its unpruned value regardless of the selection method,
  even under random sampling (Figure~\ref{fig:phenomena}(c)). As the share shrinks, the attention
  output relies less on the image and more on the text, weakening the visual evidence available for
  the answer. However, most existing methods focus on token selection while overlooking the loss of
  visual attention share.

To quantify these two sources of performance degradation, we formulate pruning as the problem of
  minimizing the shift in decoder layer outputs and derive a closed-form decomposition of this shift
  (Section~\ref{sec:method:setup}). This shift is the sum of two terms, an image summary error caused
  by the information loss and an attention allocation error caused by the drop in the visual
  attention share. We further derive an upper bound on the image summary error, which decreases as
  the retained tokens better cover all visual tokens, especially those that receive more attention
  from the query. We minimize the attention allocation error with an additive logit bias on the
  retained visual tokens to restore the visual attention share, which requires the $k$ retained
  tokens to collectively carry the total attention originally held by all $N$ visual tokens. This
  condition is underdetermined with respect to the $k$ bias terms, resulting in multiple possible
  solutions; we present two approximate solutions in closed form: a uniform bias that assigns an
  identical bias to each retained token, determined solely by the retention ratio; and a per-cell
  bias that allocates to each retained token the portion of attention originally held by the group of
  tokens it represents.

\begin{figure}[t]
  \centering
 \includegraphics[width=\textwidth]{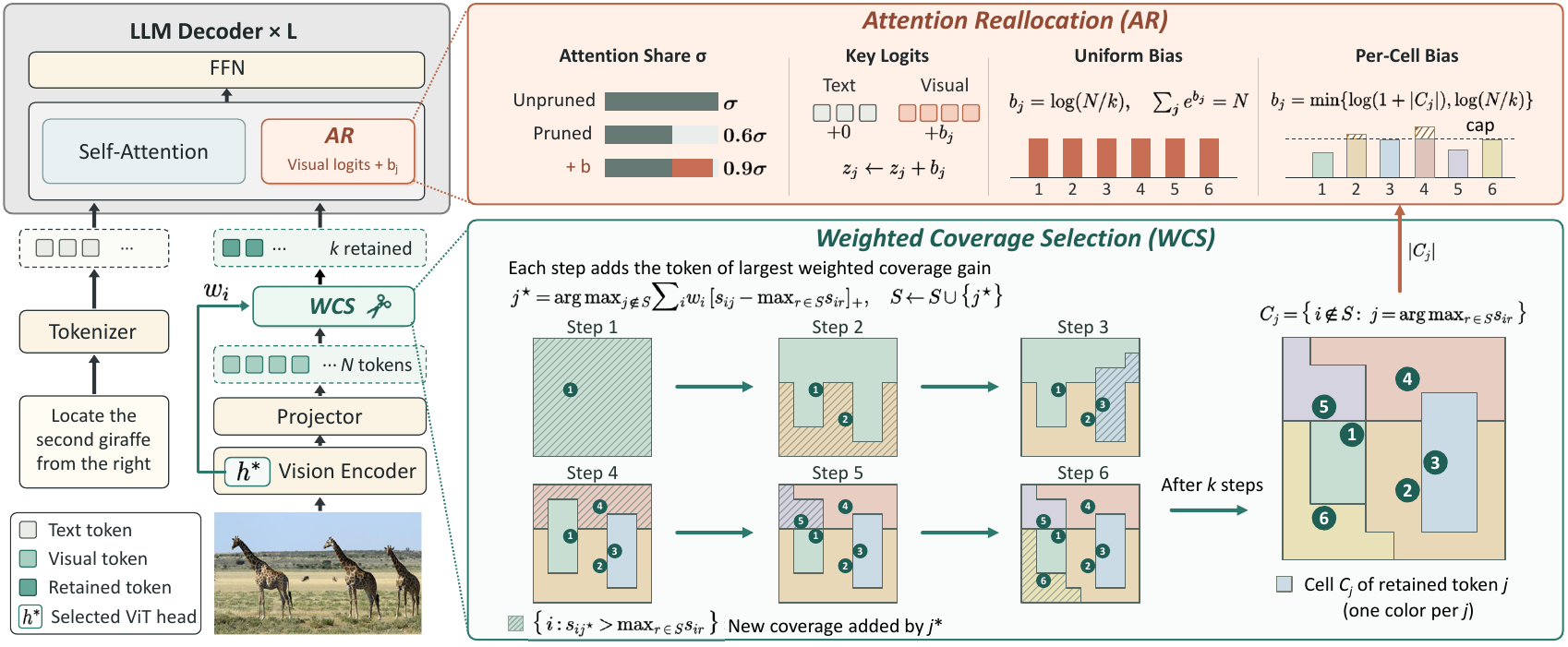}
  \caption{\textbf{Overview.}
  \textbf{Left:} {weighted coverage selection (WCS)} reduces the $N$ visual tokens to $k$ before the
  LLM, weighting the coverage by the saliency $w_i$ of the ViT head $h^{*}$ that the probe selects; AR adds a logit bias
  $b_j$ to the retained tokens in each decoder layer.
  \textbf{Bottom right:} greedy selection adds in each step the token with
  the largest new coverage; after $k$ steps, each discarded token is assigned to the most similar
  retained token, which forms the cells $C_j$.
  \textbf{Top right:} pruning lowers the image's attention share $\sigma$, and the bias restores it,
  uniformly for grounding and capped per cell for VQA.}
  \label{fig:framework}
\end{figure}

In this paper, we propose CAR, a novel framework that reduces the decoder output shift in visual
  token pruning through Coverage selection and Attention Reallocation. Specifically, as shown in
  Figure~\ref{fig:framework}, to minimize the image summary error, we introduce a weighted coverage
  selection method with a spatial kernel, which selects tokens that are both important and
  non-redundant so that they represent the entire set of visual tokens. Since the head or layer of
  the vision encoder that best supplies the token importance varies across tasks and backbones
  (Figure~\ref{fig:phenomena}(e,~f)), we further introduce a head probe that selects it by the
  consistency between the outputs of the pruned and the unpruned model. To minimize the attention
  allocation error, for grounding we apply the uniform bias, which scales the attention of all
  retained tokens by the same factor and thus keeps its spatial distribution, and for question
  answering the per-cell bias, which returns to each retained token the attention of the tokens it
  represents. The uniform bias depends only on the retention ratio and leaves the retained tokens
  unchanged, so it can be attached to any existing pruning method. Applying it alone to existing
  pruning methods brings their visual attention share closer to that of the unpruned model
  (Figure~\ref{fig:phenomena}(d)) and improves their grounding accuracy by up to $24$ points
  (Figure~\ref{fig:phenomena}(a)).
Our main contributions are summarized as follows:
\begin{list}{}{%
  \setlength{\leftmargin}{0pt}\setlength{\itemindent}{0pt}\setlength{\labelwidth}{0pt}%
  \setlength{\labelsep}{0.4em}\setlength{\listparindent}{0pt}%
  \setlength{\topsep}{3pt}\setlength{\parsep}{0pt}\setlength{\itemsep}{4pt}}
  \item[\ding{182}] We present a closed-form decomposition of the output shift caused by visual
  token pruning, together with a practical form for each of its two terms: an upper bound that
  turns the image summary error into a coverage criterion, and an exact condition that turns the
  attention allocation error into a constraint on the biases of the retained tokens.

  \item[\ding{183}] We propose Coverage selection and Attention Reallocation (CAR) for
  training-free visual token pruning, which minimizes the two errors via a saliency-weighted
  coverage objective with a label-free head probe and an additive logit bias on the retained
  tokens, yielding a retained set that represents the image and carries the attention share the
  image held before pruning.

  \item[\ding{184}] Extensive experiments show that CAR outperforms prior state-of-the-art methods across multiple MLLMs and benchmarks, leading the strongest baseline by $44.0$ points in grounding accuracy at a retention ratio of $9.9\%$, and generalizes well across VQA, grounding, and video understanding.
\end{list}

\section{Method}
\label{sec:method}

\subsection{Formulating Token Pruning as Output Shift Minimization}
\label{sec:method:setup}

To measure the effect of pruning, we consider the output of one attention head of a decoder layer at a text query, through which visual information is aggregated into the text positions where the answer is generated.
Let $T$ index the text tokens and $V$ the $N$ visual tokens, let $k_u$ and $v_u$ be the key
and the value of token $u\in T\cup V$, and let $z_u=q^{\top}k_u/\sqrt{d_h}$ be the logit with
which a text query $q$ attends to it, where $d_h$ is the head dimension. 
The unpruned attention output is
\begin{equation}
O \;=\; \sum_{u\in T\cup V} p_u\, v_u ,
\qquad
p_u \;=\; \operatorname{softmax}_{T\cup V}(z)_u \;=\; \frac{e^{z_u}}{\sum_{w\in T\cup V} e^{z_w}} .
\label{eq:attn}
\end{equation}
Token pruning retains a subset $S\subseteq V$ of size $k$ and discards the remainder, leaving the text tokens intact, so that the softmax loses terms on its visual side only. To let the retained tokens compensate for the lost terms, we allow an additive logit bias $b_u$ on each retained visual token, which scales its softmax term by $e^{b_u}$, with $b_t=0$ for $t\in T$, so that the pruned output is

\begin{equation}
O' \;=\; \sum_{u\in T\cup S} p'_u\, v_u ,
\qquad
p'_u \;=\; \operatorname{softmax}_{T\cup S}(z+b)_u \;=\; \frac{e^{z_u+b_u}}{\sum_{w\in T\cup S} e^{z_w+b_w}} .
\label{eq:pruned}
\end{equation}
Our objective is to minimize the expected output shift of the layer,
\begin{equation}
\min_{|S|=k,\;b}\ \ \mathcal{E}(S,b) \;=\; \mathbb{E}\big[\lVert O'-O\rVert^2\big],
\label{eq:objective}
\end{equation}
where the expectation is over queries and images. Two factors control this error: the retained set $S$, and the attention the decoder pays to it through $b$. Evaluating Eq.~\ref{eq:objective} requires the unpruned output $O$, however, grouping the attention sum by modality writes each output as a mixture of an image summary and a text summary (Appendix~\ref{sec:appendix:notation}),
\begin{equation}
O \;=\; \sigma\,\bar v_{\mathrm{img}} + (1-\sigma)\,\bar v_{\mathrm{txt}} ,
\qquad
O' \;=\; \sigma'\,\bar v_{S} + (1-\sigma')\,\bar v_{\mathrm{txt}} .
\label{eq:blend}
\end{equation}
Here $\sigma=\sum_{u\in V}p_u$ is the share of its attention that the query spends on the image, and
$\bar v_{\mathrm{img}}$ and $\bar v_{\mathrm{txt}}$ are the attention-weighted mean values of
the two groups; after pruning, $\sigma'$ and $\bar v_S$ are the same two quantities on the
retained set, while the text summary is unchanged. Subtracting the two outputs and splitting $\sigma'\bar v_S$ into $\sigma\bar v_S+(\sigma'-\sigma)\bar v_S$ separates the change in the summary from the change in the share:
\begin{equation}
O'-O \;=\;
\underbrace{\sigma\big(\bar v_S-\bar v_{\mathrm{img}}\big)}_{\textstyle\mathcal{E}_{\mathrm{sum}}\ (\text{image summary error})}
\;+\;
\underbrace{(\sigma'-\sigma)\big(\bar v_S-\bar v_{\mathrm{txt}}\big)}_{\textstyle\mathcal{E}_{\mathrm{alloc}}\ (\text{attention allocation error})} .
\label{eq:decomp}
\end{equation}

By the triangle inequality, 
$\|O'-O\|^2\le 2\|\mathcal{E}_{\mathrm{sum}}\|^2+2\|\mathcal{E}_{\mathrm{alloc}}\|^2$, so we minimize Eq.~\ref{eq:objective} through this upper bound, reducing the two terms separately. The image summary error captures the gap between
the summaries formed from $S$ and from $V$. The attention allocation error reflects attention
shifting from the image to the text; it vanishes exactly when $\sigma'=\sigma$.

\subsection{Minimizing the Image Summary Error}
\label{sec:method:select}
\label{sec:method:proxy}
\label{sec:method:head}

Minimizing $\mathcal{E}_{\mathrm{sum}}=\sigma(\bar v_S-\bar v_{\mathrm{img}})$ directly is
infeasible: we cannot obtain $\bar v_{\mathrm{img}}$, and choosing $S$ is a search over
$\binom{N}{k}$ subsets. It is, however, bounded by how well the retained set covers the image.

\begin{proposition}[Coverage bounds the image summary error]
\label{prop:bound}
Let $a_i=e^{z_i}/Z_V$, with $Z_V=\sum_{u\in V}e^{z_u}$, be the share of the image's attention that visual token $i$ receives from the query, so that $\sum_{i\in V}a_i=1$, and let $\rho=\sum_{j\in S}a_j$ be the share held by the retained
set. If the value vectors have equal norm $R$, and if the attention of each discarded token is
carried by the retained token with the most similar value, then
\begin{equation}
\big\lVert\bar v_S-\bar v_{\mathrm{img}}\big\rVert
\;\le\;\sqrt{2}\,R\,\sqrt{1-\rho}\,\sqrt{1-F_v(S)} ,
\qquad
F_v(S)=\sum_{i\in V}a_i\max_{j\in S}\cos(v_i,v_j) .
\label{eq:bound}
\end{equation}
\end{proposition}

\noindent
The proof is in Appendix~\ref{sec:appendix:proof}. To minimize the image summary error, the bound therefore suggests maximizing $F_v(S)$, the attention-weighted coverage of the image by the retained set. The bound asks that \emph{each visual
token be covered by a retained token similar to it, and that the coverage of a token contribute in
proportion to the attention it receives}. The shares $a_i$ are unknown at selection time, since they depend on how later layers attend to the visual tokens and vary with the layer and the query. We therefore maximize a computable surrogate of $F_v$.

\paragraph{Weighted coverage selection.}
To have the retained set well represent the original visual tokens, we formulate the token selection as a
weighted facility-location objective~\citep{iyer2021submodular},
\begin{equation}
F(S)=\sum_{i=1}^{N} w_i \max_{j\in S}s_{ij},
\qquad
s_{ij}=\big[\cos(x_i,x_j)\big]_+ ,
\label{eq:fl-ours}
\end{equation}
where $x_i$ is the encoder feature of token $i$, $w_i$ its saliency, the attention it receives
from all the tokens in the vision encoder, and $s_{ij}$ the cosine similarity between two
features with negative values clipped to zero. The inner maximum
is how well token $i$ is represented in $S$, and the sum asks the retained set to represent the
whole image, giving more weight to the salient tokens. 
The similarity of Eq.~\ref{eq:fl-ours} does not account for spatial structure, so a token can
be covered by a distant one of similar appearance, and its region may keep no token at all.
Since grounding depends on the coordinates of the target region, for grounding we multiply the
feature similarity by a spatial kernel on the token grid,
\begin{equation}
s_{ij}=\big[\cos(x_i,x_j)\big]_+\!\cdot
\exp\!\Big(\!-\frac{\lVert g_i-g_j\rVert^{2}}{2\,(N/k)}\Big),
\label{eq:fl-kernel}
\end{equation}
where $g_i$ is the grid position of token $i$. The kernel discounts the feature
similarity by the distance on the grid. The scale of the
kernel is the area each of the $k$ representatives covers, about $N/k$ cells, so it widens as
the budget tightens. Question answering, which concerns semantic information, uses
the similarity of Eq.~\ref{eq:fl-ours}, without the spatial kernel. {Unlike the bound of Proposition~\ref{prop:bound}, which uses decoder quantities, $F$ is computed in the encoder. Appendix~\ref{sec:appendix:enc-dec} shows that the two are correlated, and that a higher $F$ corresponds to a lower image summary error.}

\paragraph{Head probe.}
The heads of the vision encoder attend to different aspects of the image, and they are
therefore unequal as estimates of $a_i$. A head that disagrees with the decoder weights the
wrong tokens, and the retained set then covers parts of the image that the query does not
attend to. Accuracy varies widely with the head, and the best source differs between backbones
and task families (Figure~\ref{fig:phenomena}(e,~f)). Since $a_i$ is not observable, we select the
source with a probe: each
candidate $h$, the heads of the final block and their average, is scored by the agreement
between the pruned model's output and the unpruned model's,
\begin{equation}
\mathrm{score}(h)\;=\;\frac{1}{n}\sum_{i=1}^{n}
\mathrm{sim}\big(y_i^{(h)},\,y_i^{\mathrm{full}}\big) ,
\label{eq:head-probe}
\end{equation}
where $\mathrm{sim}$ is the task's own evaluation metric, exact match for question answering
and the IoU between the predicted boxes for grounding. The probe is run once offline for each
backbone and task family, and the best-scoring candidate $h^{*}$ is then fixed
(Appendix~\ref{sec:appendix:probe}). On InternVL3.5-8B and LLaVA-1.5-7B, whose late layers exhibit
attention sinks~\citep{darcet2024vision}, the probe selects the encoder layer as well
(Appendix~\ref{sec:appendix:internvl-attn}).

\paragraph{Greedy selection.}
$F$ is monotone and submodular, so although maximizing it is NP-hard, greedy selection
reaches at least $1-1/e$ of the optimum~\citep{nemhauser1978analysis}
(Appendix~\ref{sec:appendix:submod}). It builds $S$
one token at a time until $|S|=k$, each step adding the token of largest marginal coverage,
\begin{equation}
\arg\max_{j\notin S}\ \sum_{i=1}^{N} w_i\,\Big[\,s_{ij}-\max_{r\in S}s_{ir}\,\Big]_+ ,
\label{eq:greedy}
\end{equation}
where $\max_{r\in S}s_{ir}$ is how well token $i$ is already covered and the bracket is the
coverage that $j$ adds for it, weighted by $w_i$. Once $S$ is fixed, each discarded token is
assigned to the retained token most similar to it, and the tokens assigned to $j$ form its
cell $C_j$. 

\subsection{Minimizing the Attention Allocation Error}
\label{sec:method:bias}

Pruning lowers the share of the attention that the image receives from the query
(Figure~\ref{fig:phenomena}(c)). The allocation error
$\mathcal{E}_{\mathrm{alloc}}=(\sigma'-\sigma)(\bar v_S-\bar v_{\mathrm{txt}})$ of
Eq.~\ref{eq:decomp} is the product of the change in the share $\sigma$ and the gap between the
image summary and the text summary. The gap does not vanish in general, so the error vanishes
only when the share is restored, $\sigma'=\sigma$. We restore it with the logit bias $b_j$ of
Eq.~\ref{eq:pruned}, which raises the logits of the retained tokens.

The two shares, $\sigma$ before pruning and $\sigma'$ after it, differ only in their visual part:
both have the form $x/(Z_T+x)$ with the same text part $Z_T=\sum_{t\in T}e^{z_t}$, while $x$ is
$Z_V=\sum_{i\in V}e^{z_i}$ before pruning and $\sum_{j\in S}e^{z_j+b_j}$ after
(Appendix~\ref{sec:appendix:notation}). Since $x\mapsto x/(Z_T+x)$ is strictly increasing, the
image keeps its share exactly when the biased visual part again equals $Z_V$. With $m_j=e^{b_j}$,
\begin{equation}
\sum_{j\in S} e^{z_j}\,m_j \;=\; Z_V .
\label{eq:mass-condition}
\end{equation}
In words, the bias lets the $k$ retained tokens carry the attention mass that all $N$ visual
tokens carried before pruning. This is an equation involving $k$ unknowns, meaning multiple solutions exist, and the specific choice of solution is crucial. Since $\bar{v}_S$ is constructed using biased weights, if the $m_j$ values corresponding to the chosen solution vary across the retained tokens, the image summary will change, thereby affecting $\mathcal{E}_{\mathrm{sum}}$.

\paragraph{Uniform bias.}
The simplest solution gives each retained token the same $m_j=m$.
Equation~\ref{eq:mass-condition} then reads $m\sum_{j\in S}e^{z_j}=Z_V$, and since
$\sum_{j\in S}e^{z_j}=\rho Z_V$, where $\rho$ is the share of the visual attention held by the
retained set (Proposition~\ref{prop:bound}), this is $m\rho=1$ and $m=1/\rho$. Computing $\rho$ needs the attention of all $N$
visual tokens at a decoder layer, and pruning has already discarded $N-k$ of them. We therefore
replace $\rho$ by its expectation over queries, treating the visual tokens as exchangeable under
the query distribution, so that each receives the same attention in expectation. Then
$\mathbb{E}[\rho]=k/N$ whichever $k$ are retained, and
\begin{equation}
m_j\;=\;\frac{N}{k} ,
\qquad
b_j\;=\;\log\frac{N}{k} .
\label{eq:bias-uniform}
\end{equation}
The substitution solves Eq.~\ref{eq:mass-condition} in expectation and not per query, and the
gap grows with the variance of $\rho$. Equation~\ref{eq:bias-uniform} depends only on
the retention ratio $k/N$, and it applies a uniform bias to the retained tokens, which leaves
$\bar v_S$ and therefore $\mathcal{E}_{\mathrm{sum}}$ unchanged (Appendix~\ref{sec:appendix:derivation}).

\paragraph{Per-cell bias.}
The selector assigns each discarded token to the retained token most similar to it, and those
assigned to $j$ form its cell $C_j$ (Section~\ref{sec:method:select}). A second solution gives
each retained token the mass of the tokens it represents,
$e^{z_j}m_j=e^{z_j}+\sum_{i\in C_j}e^{z_i}$. Each discarded token belongs to exactly one cell, so
summing over $j\in S$ returns $Z_V$ and Eq.~\ref{eq:mass-condition} holds. The weight of a retained token in $\bar v_S$ is then the
attention of the tokens it represents, which is what Proposition~\ref{prop:bound} assumes. The
sum $\sum_{i\in C_j}e^{z_i}$ is not observable, and setting $e^{z_i}\approx e^{z_j}$
inside a cell gives
\begin{equation}
m_j \;=\; 1+|C_j| ,
\qquad
b_j \;=\; \log\big(1+|C_j|\big) .
\label{eq:bias-accounting}
\end{equation}
The assumption is exchangeability inside a cell, which the selector built by similarity. Since $m_j$ grows with the size of the cell, a token
that stands for many discarded ones is weighted up, and $\bar v_S$ changes with it. Inside a
cell $e^{z_i}<e^{z_j}$ is the common case, so the count overstates what the cell carried, the
more so the larger the cell. To limit it we cap the count at its mean $N/k$,
\begin{equation}
m_j \;=\; \min\Big\{1+|C_j|,\ \frac{N}{k}\Big\} ,
\qquad
b_j \;=\; \min\Big\{\log\big(1+|C_j|\big),\ \log\frac{N}{k}\Big\} ,
\label{eq:bias-capped}
\end{equation}
which restores less share than pruning removed and perturbs $\bar v_S$ less
(Appendix~\ref{sec:appendix:percell}).

\paragraph{Choice of the bias form.}
The uniform bias assigns the same value to each retained token, while the per-cell forms give
more to the tokens that stand for more discarded ones, so what they restore follows the
redundancy of each region. Grounding asks \emph{where}, so the answer depends on which part of
the image the attention falls on, and the uniform bias scales each retained token by the same
factor and leaves that unchanged. Question answering asks
\emph{what}, and the answer depends on the content of the image, which the per-cell forms
recover more faithfully by returning to each retained token the attention of the tokens it
represents. We therefore use the uniform bias for grounding and the capped per-cell bias for
question answering (Section~\ref{sec:exp:ablation}).

\section{Experiments}
\label{sec:exp}

\paragraph{Experimental setup.}
\label{sec:exp:setup}

To verify the generality and effectiveness of our method, we evaluate it on image understanding,
visual grounding and video understanding across several MLLMs and benchmarks. The backbones
are Qwen2.5-VL-7B~\citep{Bai2025qwen2.5-vl}, Qwen3-VL-8B~\citep{bai2025qwen3},
InternVL3.5-8B~\citep{wang2025internvl3} and LLaVA-1.5-7B~\citep{liu2024llava1.5}. We use eight VQA benchmarks (e.g.,
TextVQA~\citep{singh2019towards} and MME~\citep{fu2023mme}), the eight splits of RefCOCO, RefCOCO+~\citep{yu2016modeling} and RefCOCOg~\citep{mao2016generation} for visual grounding, and three video benchmarks on Qwen3-VL-8B
(e.g., Video-MME~\citep{fu2025video}). All experiments run on NVIDIA RTX 3090 GPUs. Detailed
configurations are given in Appendix~\ref{sec:appendix:impl}.

\subsection{Main Results}
\label{sec:exp:main}

\begin{table}[t]
  \centering
  \caption{VQA performance comparison on Qwen2.5-VL-7B. \textbf{Bold}: Best. \underline{Underline}: Second best.}
  \label{tab:qwen2.5-vl}
  \setlength{\tabcolsep}{4.5pt}  %
  \renewcommand{\arraystretch}{0.88}  %
  \resizebox{\linewidth}{!}{
    \begin{tabular}{l|cccccccc|cc}
    \toprule
    \textbf{Method} & \textbf{TextVQA} & \textbf{ChartQA} & \textbf{AI2D} & \textbf{OCRBench} & \textbf{HallBench} & \textbf{MME} & \textbf{MMB-EN} & \textbf{MMB-CN} & \textbf{Acc.} & \textbf{Rel.} \\
    \midrule
    \rowcolor{budgetrow}\multicolumn{11}{c}{\textit{Upper Bound, All 1{,}296 Tokens} (100\%)} \\
    \rowcolor{refrow} Qwen2.5-VL-7B & 84.3 & 87.1 & 81.6 & 867 & 56.2 & 2342.3 & 83.1 & 82.9 & 84.9 & 100.0\% \\
    \rowcolor{budgetrow}\multicolumn{11}{c}{\textit{256 Tokens on Average} ($\downarrow$80.2\%)} \\
    VisionZip{\footnotesize\color{venuec}\texttt{(CVPR25)}} & 76.1 & 77.2 & 79.3 & 722 & 51.3 & 2277.2 & 80.6 & 81.3 & 79.0 & 92.9\% \\
    DivPrune{\footnotesize\color{venuec}\texttt{(CVPR25)}} & 75.3 & 69.2 & 77.1 & 666 & 46.8 & 2182.9 & 79.7 & 79.2 & 75.4 & 88.5\% \\
    CDPruner{\footnotesize\color{venuec}\texttt{(NeurIPS25)}} & 77.4 & 60.5 & 79.6 & 669 & 50.5 & 2258.7 & 79.9 & 80.1 & 76.0 & 89.4\% \\
    HiPrune{\footnotesize\color{venuec}\texttt{(ACL26)}} & 77.9 & 80.4 & 79.7 & 738 & 51.0 & 2274.5 & 81.0 & \textbf{82.0} & 79.9 & 94.0\% \\
    AnchorPrune{\footnotesize\color{venuec}\texttt{(ECCV26)}} & 80.5 & \underline{82.7} & \textbf{80.4} & 804 & \underline{51.4} & 2290.4 & 81.1 & \underline{81.8} & 81.6 & 95.9\% \\
    EADP{\footnotesize\color{venuec}\texttt{(ECCV26)}} & 68.7 & 68.8 & 76.6 & 702 & 48.7 & 2217.3 & 80.7 & 79.1 & 75.5 & 88.6\% \\
    FastV{\footnotesize\color{venuec}\texttt{(ECCV24)}} & 78.8 & 66.0 & 75.0 & 631 & 48.2 & 2130.7 & 77.1 & 76.9 & 74.0 & 87.0\% \\
    SparseVLM{\footnotesize\color{venuec}\texttt{(ICML25)}} & 75.7 & 52.3 & 77.8 & 587 & 48.5 & 2255.0 & 80.5 & 79.9 & 73.3 & 86.1\% \\
    DART{\footnotesize\color{venuec}\texttt{(EMNLP25)}} & 70.5 & 63.6 & 74.8 & 623 & 44.6 & 2197.2 & 79.3 & 78.4 & 72.9 & 85.4\% \\
    HAWK{\footnotesize\color{venuec}\texttt{(CVPR26)}} & \textbf{82.7} & 81.7 & 79.7 & \underline{810} & 51.1 & \underline{2317.6} & \underline{81.7} & 81.4 & \underline{81.9} & \underline{96.2\%} \\
    VScan{\footnotesize\color{venuec}\texttt{(TMLR26)}} & 81.2 & 79.8 & 74.6 & 740 & 49.5 & 2235.5 & 77.0 & 77.9 & 78.2 & 91.9\% \\
    N\"uwa{\footnotesize\color{venuec}\texttt{(ICLR26)}} & 79.5 & 81.6 & 79.9 & 751 & 50.4 & 2286.6 & 81.0 & \textbf{82.0} & 80.5 & 94.5\% \\
    \rowcolor{carrow}\textbf{CAR} (Ours) & \underline{81.8} & \textbf{83.9} & \underline{80.2} & \textbf{831} & \textbf{53.4} & \textbf{2335.7} & \textbf{82.3} & \textbf{82.0} & \textbf{82.9} & \textbf{97.5\%} \\
    \rowcolor{budgetrow}\multicolumn{11}{c}{\textit{128 Tokens on Average} ($\downarrow$90.1\%)} \\
    VisionZip{\footnotesize\color{venuec}\texttt{(CVPR25)}} & 66.4 & 63.4 & 76.3 & 592 & 44.1 & 2184.1 & 77.8 & 77.9 & 71.8 & 84.1\% \\
    DivPrune{\footnotesize\color{venuec}\texttt{(CVPR25)}} & 65.9 & 51.4 & 72.7 & 535 & 43.2 & 2044.6 & 77.3 & 76.7 & 67.9 & 79.7\% \\
    CDPruner{\footnotesize\color{venuec}\texttt{(NeurIPS25)}} & 72.5 & 46.9 & 77.3 & 581 & 47.0 & 2181.6 & 79.2 & 77.6 & 71.0 & 83.4\% \\
    HiPrune{\footnotesize\color{venuec}\texttt{(ACL26)}} & 68.6 & 66.9 & 77.1 & 629 & 47.1 & 2192.7 & 79.1 & 78.5 & 73.7 & 86.6\% \\
    AnchorPrune{\footnotesize\color{venuec}\texttt{(ECCV26)}} & 75.9 & 71.1 & \underline{78.2} & 692 & 48.2 & 2230.9 & \underline{79.8} & \underline{80.3} & 76.8 & 90.2\% \\
    EADP{\footnotesize\color{venuec}\texttt{(ECCV26)}} & 60.1 & 54.5 & 73.2 & 597 & 42.8 & 2152.9 & 78.8 & 77.3 & 69.3 & 81.1\% \\
    FastV{\footnotesize\color{venuec}\texttt{(ECCV24)}} & 39.3 & 24.1 & 67.7 & 310 & 37.8 & 1395.8 & 41.3 & 40.7 & 44.0 & 52.3\% \\
    SparseVLM{\footnotesize\color{venuec}\texttt{(ICML25)}} & 60.4 & 28.0 & 72.0 & 363 & 41.9 & 1969.7 & 72.0 & 71.2 & 60.0 & 70.6\% \\
    DART{\footnotesize\color{venuec}\texttt{(EMNLP25)}} & 56.5 & 47.0 & 69.8 & 512 & 40.5 & 2046.3 & 77.2 & 76.0 & 65.1 & 76.2\% \\
    HAWK{\footnotesize\color{venuec}\texttt{(CVPR26)}} & \textbf{79.9} & \underline{71.3} & 77.5 & \underline{705} & 48.8 & 2225.6 & 79.0 & 78.6 & \underline{77.1} & \underline{90.6\%} \\
    VScan{\footnotesize\color{venuec}\texttt{(TMLR26)}} & 74.4 & 65.6 & 72.9 & 605 & 46.3 & 2138.3 & 75.5 & 75.9 & 72.3 & 84.8\% \\
    N\"uwa{\footnotesize\color{venuec}\texttt{(ICLR26)}} & 71.3 & 70.6 & 77.4 & 611 & \textbf{49.2} & \underline{2247.8} & 78.2 & 79.4 & 74.9 & 88.0\% \\
    \rowcolor{carrow}\textbf{CAR} (Ours) & \underline{78.8} & \textbf{76.1} & \textbf{79.4} & \textbf{731} & \underline{49.0} & \textbf{2314.3} & \textbf{81.1} & \textbf{81.3} & \textbf{79.3} & \textbf{93.0\%} \\
    \bottomrule
    \end{tabular}
  }
\end{table}

\begin{table}[t]
  \centering
  \caption{VG performance comparison on Qwen2.5-VL-7B. \textbf{Bold}: Best. \underline{Underline}: Second best.}
  \label{tab:grounding-main}
  \scriptsize
  \setlength{\tabcolsep}{8.5pt}  %
  \setlength{\aboverulesep}{0.2ex}
  \setlength{\belowrulesep}{0.2ex}
  \renewcommand{\arraystretch}{0.88}  %
    \begin{tabular}{l|ccc|ccc|cc|cc}
      \toprule
      \multirow{2}{*}{\textbf{Method}} & \multicolumn{3}{c|}{\textbf{RefCOCO}} & \multicolumn{3}{c|}{\textbf{RefCOCO+}}
      & \multicolumn{2}{c|}{\textbf{RefCOCOg}} & \multirow{2}{*}{\textbf{Avg.}} & \multirow{2}{*}{\textbf{Rel.}} \\
      & testA & testB & val & testA & testB & val & test & val & & \\
      \midrule
      \rowcolor{budgetrow}\multicolumn{11}{c}{\textit{Upper Bound, All Visual Tokens} (100\%)} \\
      \rowcolor{refrow} Qwen2.5-VL-7B & 92.2 & 84.8 & 89.4 & 87.6 & 74.1 & 82.8 & 86.7 & 86.7 & 85.5 & 100.0\% \\
      \rowcolor{budgetrow}\multicolumn{11}{c}{\textit{Retain 25\% of Visual Tokens}} \\
      Random & 63.9 & 52.1 & 58.3 & 58.1 & 44.5 & 52.5 & 52.7 & 52.7 & 54.4 & 63.4\% \\
      Pooling & 64.1 & 49.1 & 57.5 & 58.8 & 42.0 & 51.6 & 50.7 & 50.3 & 53.0 & 61.8\% \\
      VisionZip{\tiny\color{venuec}\texttt{(CVPR25)}} & 71.0 & 59.0 & 66.7 & 65.0 & 50.5 & 59.1 & 61.2 & 61.2 & 61.7 & 72.0\% \\
      DivPrune{\tiny\color{venuec}\texttt{(CVPR25)}} & 74.1 & 59.8 & 67.5 & 68.6 & 51.9 & 60.9 & 63.3 & 63.7 & 63.7 & 74.3\% \\
      AnchorPrune{\tiny\color{venuec}\texttt{(ECCV26)}} & 75.6 & 62.4 & 68.3 & 68.4 & 53.2 & 61.5 & 64.7 & 65.5 & 64.9 & 75.8\% \\
      EADP{\tiny\color{venuec}\texttt{(ECCV26)}} & 71.9 & 53.1 & 63.3 & 68.0 & 44.4 & 57.2 & 56.9 & 56.3 & 58.9 & 68.6\% \\
      FastV{\tiny\color{venuec}\texttt{(ECCV24)}} & 23.7 & 30.7 & 28.4 & 21.6 & 25.6 & 24.1 & 29.1 & 29.2 & 26.6 & 31.2\% \\
      SparseVLM{\tiny\color{venuec}\texttt{(ICML25)}} & 30.9 & 31.7 & 31.3 & 26.4 & 25.3 & 24.9 & 29.3 & 29.9 & 28.7 & 33.6\% \\
      HAWK{\tiny\color{venuec}\texttt{(CVPR26)}} & 40.4 & 45.7 & 42.7 & 36.0 & 40.0 & 38.0 & 44.9 & 45.7 & 41.7 & 48.9\% \\
      VScan{\tiny\color{venuec}\texttt{(TMLR26)}} & \underline{78.7} & 66.9 & 71.7 & \underline{72.3} & 55.5 & 64.0 & 67.9 & 68.2 & 68.2 & 79.5\% \\
      N\"uwa{\tiny\color{venuec}\texttt{(ICLR26)}} & 77.5 & \underline{72.5} & \underline{75.6} & 69.7 & \underline{60.3} & \underline{65.8} & \underline{72.8} & \underline{72.3} & \underline{70.8} & \underline{82.7\%} \\
      \rowcolor{carrow}\textbf{CAR} (Ours) & \textbf{87.0} & \textbf{80.1} & \textbf{84.3} & \textbf{80.2} & \textbf{68.3} & \textbf{75.9} & \textbf{79.4} & \textbf{79.6} & \textbf{79.4} & \textbf{92.7\%} \\
      \rowcolor{budgetrow}\multicolumn{11}{c}{\textit{Retain 9.9\% of Visual Tokens}} \\
      Random & 23.1 & 17.2 & 21.5 & 20.3 & 14.5 & 18.6 & 17.3 & 18.1 & 18.8 & 21.9\% \\
      Pooling & 20.0 & 15.6 & 17.5 & 18.2 & 13.0 & 15.5 & 14.8 & 14.2 & 16.1 & 18.8\% \\
      VisionZip{\tiny\color{venuec}\texttt{(CVPR25)}} & 23.3 & 19.2 & 20.5 & 18.8 & 16.0 & 17.7 & 18.4 & 19.5 & 19.2 & 22.4\% \\
      DivPrune{\tiny\color{venuec}\texttt{(CVPR25)}} & \underline{27.8} & 19.6 & 24.2 & \underline{24.7} & 15.8 & \underline{20.9} & 21.3 & 21.5 & 22.0 & 25.6\% \\
      AnchorPrune{\tiny\color{venuec}\texttt{(ECCV26)}} & 24.4 & 18.0 & 22.0 & 21.4 & 16.2 & 19.4 & 19.7 & 20.2 & 20.2 & 23.5\% \\
      EADP{\tiny\color{venuec}\texttt{(ECCV26)}} & 24.1 & 16.5 & 21.2 & 23.4 & 14.6 & 18.6 & 17.4 & 18.2 & 19.2 & 22.4\% \\
      FastV{\tiny\color{venuec}\texttt{(ECCV24)}} & 3.2 & 2.7 & 3.2 & 3.3 & 2.9 & 3.5 & 3.6 & 3.3 & 3.2 & 3.8\% \\
      SparseVLM{\tiny\color{venuec}\texttt{(ICML25)}} & 5.7 & 5.7 & 5.8 & 3.8 & 4.2 & 3.8 & 5.4 & 5.3 & 5.0 & 5.8\% \\
      HAWK{\tiny\color{venuec}\texttt{(CVPR26)}} & 8.1 & 8.6 & 7.5 & 6.9 & 7.0 & 7.0 & 8.5 & 7.5 & 7.6 & 8.9\% \\
      VScan{\tiny\color{venuec}\texttt{(TMLR26)}} & 23.8 & 19.8 & 21.4 & 20.4 & 15.5 & 18.0 & 18.5 & 20.5 & 19.7 & 23.0\% \\
      N\"uwa{\tiny\color{venuec}\texttt{(ICLR26)}} & 27.5 & \underline{23.7} & \underline{26.2} & 21.8 & \underline{17.4} & 20.5 & \underline{23.0} & \underline{22.0} & \underline{22.8} & \underline{26.5\%} \\
      \rowcolor{carrow}\textbf{CAR} (Ours) & \textbf{75.4} & \textbf{66.7} & \textbf{72.2} & \textbf{67.2} & \textbf{55.9} & \textbf{62.3} & \textbf{67.8} & \textbf{66.5} & \textbf{66.8} & \textbf{77.9\%} \\
      \bottomrule
    \end{tabular}
\end{table}

\paragraph{Performance on VQA tasks.}
Table~\ref{tab:qwen2.5-vl} compares CAR with representative baselines on Qwen2.5-VL-7B. CAR achieves
the best average accuracy at both retention ratios and leads on most benchmarks. With $80\%$ and
$90\%$ of the visual tokens pruned, it retains $97.5\%$ and $93.0\%$ of the unpruned performance,
against $96.2\%$ and $90.6\%$ for the strongest baseline, HAWK. The best-performing baseline varies across benchmarks (HAWK on TextVQA, AnchorPrune on AI2D and N\"uwa on HallusionBench), whereas CAR ranks first or second on every benchmark at both ratios. 

\textbf{Performance on visual grounding tasks.}
  Visual grounding relies on fine-grained spatial information and suffers from severe degradation
  under aggressive pruning. As shown in Table~\ref{tab:grounding-main}, with $75\%$ of the visual
  tokens pruned, CAR retains $92.7\%$ of the unpruned performance, against $82.7\%$ for the strongest
  baseline, N\"uwa. With $90\%$ pruned, all baselines collapse to at most $26.5\%$, close to random
  sampling ($21.9\%$). At both ratios the decoder-side methods (FastV, SparseVLM and HAWK) even fall
  below random sampling: they keep the top-$k$ tokens ranked by the attention from the text, so the
  retained tokens cluster on a few regions and leave much of the image uncovered, which breaks the
  spatial structure that grounding relies on. Our CAR significantly outperforms other methods,
  retaining $77.9\%$ of the original model's performance. This lead does not come from the bias
  alone: with the same bias added to every method, CAR still leads the strongest one by $17.7$ points
  (Table~\ref{tab:grounding-budget}).

\begin{table}[t]
  \centering
  \caption{Average performance comparison across backbones and tasks.}
  \label{tab:cross-backbone}
  \setlength{\tabcolsep}{6.5pt}
  \setlength{\aboverulesep}{0.2ex}
  \setlength{\belowrulesep}{0.2ex}
  \renewcommand{\arraystretch}{0.90}
  \resizebox{\linewidth}{!}{
    \begin{tabular}{l|cc|cc|cc|cc|c|cc}
      \toprule
      & \multicolumn{4}{c|}{\textbf{VQA}} & \multicolumn{5}{c|}{\textbf{Visual Grounding}} & \multicolumn{2}{c}{\textbf{Video}} \\
      & \multicolumn{2}{c|}{Qwen3-VL-8B} & \multicolumn{2}{c|}{InternVL3.5-8B} & \multicolumn{2}{c|}{Qwen3-VL-8B}
      & \multicolumn{2}{c|}{InternVL3.5-8B} & \multicolumn{1}{c|}{LLaVA-1.5-7B} & \multicolumn{2}{c}{Qwen3-VL-8B} \\
      \textbf{Method} & 22.2\% & 11.1\% & 22.2\% & 11.1\% & 25\% & 9.9\% & 25\% & 9.9\% & 11.1\% & 25\% & 9.9\% \\
      \midrule
      \rowcolor{refrow} Unpruned & 85.9 & 85.9 & 83.0 & 83.0 & 85.0 & 85.0 & 89.4 & 89.4 & 51.3 & 65.5 & 65.5 \\
      VisionZip{\footnotesize\color{venuec}\texttt{(CVPR25)}} & 75.1 & 59.9 & 66.1 & 62.8 & 64.0 & 25.5 & 78.5 & 70.8 & 8.0 & 63.2 & 59.5 \\
      DivPrune{\footnotesize\color{venuec}\texttt{(CVPR25)}} & 72.5 & 62.0 & 73.3 & 69.1 & 58.7 & 25.4 & 80.1 & 73.3 & 17.3 & 61.7 & 58.1 \\
      AnchorPrune{\footnotesize\color{venuec}\texttt{(ECCV26)}} & \underline{76.5} & 62.2 & 71.0 & 66.5 & \underline{65.2} & 26.0 & 78.9 & 71.9 & 15.1 & 63.4 & 60.4 \\
      EADP{\footnotesize\color{venuec}\texttt{(ECCV26)}} & 75.1 & \underline{66.6} & \underline{73.6} & \underline{69.5} & 62.0 & 22.4 & \underline{82.0} & \underline{75.0} & 20.0 & 63.3 & 59.8 \\
      GSTEP{\footnotesize\color{venuec}\texttt{(MM26)}} & -- & -- & -- & -- & -- & -- & -- & -- & -- & 62.7 & 60.9 \\
      N\"uwa{\footnotesize\color{venuec}\texttt{(ICLR26)}} & 75.9 & 64.7 & 63.4 & 58.2 & 64.5 & \underline{28.9} & 72.9 & 62.7 & \underline{28.2} & \underline{63.5} & \underline{61.6} \\
      \rowcolor{carrow}\textbf{CAR} (Ours) & \textbf{79.7} & \textbf{70.6} & \textbf{79.1} & \textbf{74.4} & \textbf{72.2} & \textbf{46.6} & \textbf{83.9} & \textbf{76.2} & \textbf{30.4} & \textbf{64.0} & \textbf{62.0} \\
      \bottomrule
    \end{tabular}
  }
\end{table}

\paragraph{Generalization across backbones and tasks.}
To verify the generality and effectiveness of CAR, we evaluate it on three additional MLLMs,
Qwen3-VL-8B with DeepStack, InternVL3.5-8B with dynamic tiling, and LLaVA-1.5-7B, across VQA, VG,
and video understanding (Table~\ref{tab:cross-backbone}). CAR performs best across varying backbones,
tasks and retention ratios. On VQA it leads the strongest baseline at both retention ratios on Qwen3-VL and InternVL3.5. On grounding its advantage is largest at the aggressive ratio on Qwen3-VL, the
setting where pruning costs the most. On video it also leads every baseline, including GSTEP, which
is designed for video. Per-benchmark results are in Appendix~\ref{sec:appendix:results}.

\paragraph{Efficiency analysis.}
\label{sec:exp:efficiency}

Table~\ref{tab:efficiency} evaluates efficiency on TextVQA with Qwen2.5-VL-7B, keeping $128$ of the
$1{,}296$ visual tokens, with batch size $1$. Compared with the unpruned model, CAR reduces the
prefill time by $1.96\times$ and the end-to-end latency by $1.78\times$, and shrinks the KV cache by
$7.8\times$. CAR reaches the highest accuracy at a cost comparable to the other pruning methods.

\begin{table}[t]
  \centering
  \caption{Efficiency on Qwen2.5-VL-7B at $128$ of $1{,}296$ visual tokens.}
  \label{tab:efficiency}
  \setlength{\tabcolsep}{10pt}  %
  \setlength{\aboverulesep}{0.2ex}
  \setlength{\belowrulesep}{0.2ex}
  \renewcommand{\arraystretch}{0.90}
  \resizebox{\linewidth}{!}{
    \begin{tabular}{l|ccc|ccc}
    \toprule
    \textbf{Method} & \textbf{Prefill (ms)} & \textbf{Decode (ms/tok)} & \textbf{Latency (ms)} & \textbf{FLOPs (T)} & \textbf{KV (MB)} & \textbf{Mem. (GB)} \\
    \midrule
    \rowcolor{refrow} Qwen2.5-VL-7B & 484.0 & 27.1 & 543.3 & 10.28 & 73.2 & 16.02 \\
    \rowcolor{budgetrow}\multicolumn{7}{c}{\textit{128 Tokens on Average} ($\downarrow$90.1\%)} \\
    AnchorPrune{\small\color{venuec}\texttt{(ECCV26)}} & 262.0 {\scriptsize\color{black!50}(1.85$\times$)} & 27.3 & 319.8 {\scriptsize\color{black!50}(1.70$\times$)} & 3.81 {\scriptsize\color{black!50}(2.70$\times$)} & 9.4 & 15.78 \\
    EADP{\small\color{venuec}\texttt{(ECCV26)}} & 254.8 {\scriptsize\color{black!50}(1.90$\times$)} & 27.1 & 312.1 {\scriptsize\color{black!50}(1.74$\times$)} & 3.81 {\scriptsize\color{black!50}(2.70$\times$)} & 9.4 & 16.34 \\
    FastV{\small\color{venuec}\texttt{(ECCV24)}} & 222.6 {\scriptsize\color{black!50}(2.17$\times$)} & 27.1 & 280.1 {\scriptsize\color{black!50}(1.94$\times$)} & 3.83 {\scriptsize\color{black!50}(2.69$\times$)} & 9.3 & 15.73 \\
    SparseVLM{\small\color{venuec}\texttt{(ICML25)}} & 241.5 {\scriptsize\color{black!50}(2.00$\times$)} & 27.1 & 300.0 {\scriptsize\color{black!50}(1.81$\times$)} & 3.85 {\scriptsize\color{black!50}(2.67$\times$)} & 9.6 & 16.44 \\
    HAWK{\small\color{venuec}\texttt{(CVPR26)}} & 231.8 {\scriptsize\color{black!50}(2.09$\times$)} & 27.1 & 288.1 {\scriptsize\color{black!50}(1.89$\times$)} & 3.81 {\scriptsize\color{black!50}(2.70$\times$)} & 9.4 & 16.34 \\
    VScan{\small\color{venuec}\texttt{(TMLR26)}} & 284.5 {\scriptsize\color{black!50}(1.70$\times$)} & 27.3 & 358.0 {\scriptsize\color{black!50}(1.52$\times$)} & 3.80 {\scriptsize\color{black!50}(2.70$\times$)} & 9.3 & 19.81 \\
    N\"uwa{\small\color{venuec}\texttt{(ICLR26)}} & 232.1 {\scriptsize\color{black!50}(2.09$\times$)} & 27.0 & 287.6 {\scriptsize\color{black!50}(1.89$\times$)} & 3.81 {\scriptsize\color{black!50}(2.70$\times$)} & 9.4 & 16.36 \\
    \rowcolor{carrow}\textbf{CAR} (Ours) & 246.6 {\scriptsize\color{black!50}(1.96$\times$)} & 27.3 & 304.7 {\scriptsize\color{black!50}(1.78$\times$)} & 3.81 {\scriptsize\color{black!50}(2.70$\times$)} & 9.4 & 15.82 \\
    \bottomrule
    \end{tabular}
  }
\end{table}

\subsection{Ablation}
\label{sec:exp:ablation}

\begin{table}[t]
  \setlength{\tabcolsep}{3pt}
  \renewcommand{\arraystretch}{1.05}
  \begin{minipage}[t]{0.345\linewidth}
    \centering
    \caption{Component ablation.}
    \label{tab:abl-comp}
    \scriptsize
    \begin{tabular}{l|cc}
      \toprule
      Configuration & VG & VQA \\
      \midrule
      Unpruned & 85.5 & 84.9 \\
      \midrule
      Random & 18.8 & 64.2 \\
      \ + bias & 38.6 & 64.5 \\
      \midrule
      Coverage, $w_i{=}1$ & 20.1 & 72.2 \\
      \ + bias & 50.0 & 73.7 \\
      \ + saliency $w$ (head avg.) & 54.8 & 78.5 \\
      \ + head $h^{*}$ & 62.5 & \cellcolor{carrow}\textbf{79.3} \\
      \ + spatial kernel & \cellcolor{carrow}\textbf{66.8} & 79.1 \\
      \bottomrule
    \end{tabular}
  \end{minipage}\hfill
  \begin{minipage}[t]{0.635\linewidth}
    \centering
    \caption{Selection-rule ablation.}
    \label{tab:abl-select}
    \scriptsize
    \setlength{\aboverulesep}{1.25pt}\setlength{\belowrulesep}{1.25pt}
    \renewcommand{\arraystretch}{0.95}
    \begin{tabular}{cll|cc}
      \toprule
      & Score or objective & Representative & VG & VQA \\
      \midrule
      \rnum{1} & $w_i$ & FastV & 58.6 & 78.1 \\
      \rnum{2} & $\min_{j\in S}(1-s_{ij})$, $w$ unused & DivPrune & 44.4 & 67.8 \\
      \rnum{3} & $\lambda w_i-(1-\lambda)\max_{j\in S}s_{ij}$ & IDPruner & 59.1 & 78.4 \\
      \rnum{4} & $w_i^{\alpha}\min_{j\in S}(1-s_{ij})$ & AnchorPrune & 59.6 & 78.6 \\
      \rnum{5} & $\log\det K_S,\ K_{ij}=w_iw_js_{ij}$ & CDPruner & 59.5 & 78.1 \\
      \rnum{6} & $(1-\alpha)w_i+\alpha\,\Delta_i\log\det(X_SX_S^{\top}+\lambda I)$ & CoIn & 52.5 & 74.3 \\
      \rnum{7} & $w_i$ subject to $\max_{j\in S}s_{ij}<\tau$ & AgilePruner & 60.5 & 76.1 \\
      \rnum{8} & Pivots by $w_i$, then graph suppression & D\textsuperscript{2}Pruner & 61.4 & 77.1 \\
      \rnum{9} & $w_i\sum_{l\in V}[s_{il}-\max_{j\in S}s_{jl}]_+$ & SCOPE & 62.1 & 78.4 \\
      \midrule
      \rnum{10} & $\sum_{i\in V}w_i\max_{j\in S}s_{ij}$ (Ours) & --- & \cellcolor{carrow}\textbf{66.4} & \cellcolor{carrow}\textbf{78.7} \\
      \bottomrule
    \end{tabular}
  \end{minipage}
\end{table}

\paragraph{Ablation on components.}
To assess the contribution of each component, Table~\ref{tab:abl-comp} starts from the unweighted
coverage selection ($w_i=1$) and adds one component at a time, with random selection as a
reference. On grounding the bias is decisive, raising accuracy by $29.9$ points, while on VQA it
is worth $1.5$. Selection behaves the other way: unweighted coverage alone gains $8.0$ points
compared with random selection on VQA, and weighting it by the encoder saliency averaged over the heads adds $4.8$.
Selecting the head with the probe improves both tasks, and the spatial kernel adds $4.3$ points on
grounding but costs $0.2$ on VQA, so we apply it to grounding only. The two tasks therefore rely
on different components, one for each error term of Eq.~\ref{eq:decomp}. VG averages the eight
RefCOCO splits and VQA the eight benchmarks; the per-split and per-benchmark results are in
Appendix~\ref{sec:appendix:results-components}.

\paragraph{Ablation on selection rule.}
Table~\ref{tab:abl-select} compares ten representative selection rules that share the saliency, retention ratio, bias and features of CAR and differ only in how tokens are selected (Appendix~\ref{sec:appendix:taxonomy}); VG is averaged over three of the eight RefCOCO splits and
VQA over the eight benchmarks. Averaged over the two tasks, the coverage-based rules, CAR \rnum{10} and SCOPE \rnum{9}, outperform the other rules, and by also accounting for spatial distance in the similarity, CAR performs best on grounding. Combining the saliency with the diversity is better than either alone: diversity alone \rnum{2} drops task-relevant tokens, and saliency alone \rnum{1} keeps redundant ones.
The per-split and per-benchmark
results are given in Appendix~\ref{sec:appendix:selection-full}.

\vspace{1mm}
\noindent
\begin{minipage}[t]{0.38\linewidth}
  \vspace{0pt}

\captionof{table}{Bias-form ablation.}
\label{tab:abl-bias}
\centering
\scriptsize
\setlength{\tabcolsep}{3pt}
\renewcommand{\arraystretch}{1.62}  %
\begin{tabular}{l|cc}
  \toprule
  Bias $b_j$ & VG & VQA \\
  \midrule
  None & 26.8 & 78.7 \\
  $\log(N/k)$ & \cellcolor{carrow}\textbf{66.8} & 78.3 \\
  $\log(1{+}|C_j|)$ & 65.5 & 79.2 \\
  $\min\{\log(1{+}|C_j|),\log(N/k)\}$ & 66.3 & \cellcolor{carrow}\textbf{79.3} \\
  \bottomrule
\end{tabular}

\end{minipage}\hfill
\begin{minipage}[t]{0.58\linewidth}
  \vspace{0pt}

\captionof{table}{Comparison with RESTORE.}
\label{tab:restore-main}
\centering
\scriptsize
\setlength{\tabcolsep}{3pt}
\renewcommand{\arraystretch}{1.05}
\begin{tabular}{l|cc|cc}
  \toprule
  \multirow{2}{*}{Method}
  & \multicolumn{2}{c|}{DivPrune} & \multicolumn{2}{c}{WCS} \\
  & VQA & VG & VQA & VG \\
  \midrule
  Unpruned & 84.9 & 85.5 & 84.9 & 85.5 \\
  \midrule
  No calibration & 67.9 & 22.0 & 78.7 & 26.8 \\
  \ + RESTORE merging & 69.8 & 13.1 & 75.3 & 14.3 \\
  \ + RESTORE merging and bias & \textbf{70.0} & 32.6 & 75.2 & 44.9 \\
  \ + our bias & 69.3 & \textbf{44.4} & \textbf{79.3} & \textbf{66.8} \\
  \bottomrule
\end{tabular}

\end{minipage}
\vspace{1mm}

\paragraph{Ablation on bias form.}
Table~\ref{tab:abl-bias} fixes the retained tokens and compares the three bias forms of
Section~\ref{sec:method:bias}. On grounding, all forms lift accuracy from $26.8$ to above $65$,
and the uniform bias, which preserves the attention across image regions, performs best. On VQA,
the per-cell and capped forms perform better. This matches the location-sensitive nature of
grounding and the content-centric nature of VQA, so we use the uniform bias for grounding and the
capped bias for VQA. The per-split and per-benchmark results are in
Appendix~\ref{sec:appendix:bias-forms}.

\begingroup

\paragraph{Comparison with RESTORE.}
\label{sec:exp:restore}
RESTORE~\citep{cho2026restore} is the closest work to our attention reallocation: it also compensates
for the attention of the removed tokens, but merges them into representatives and calibrates the
attention to these representatives. Table~\ref{tab:restore-main} compares RESTORE with our attention reallocation on two selectors, DivPrune and our weighted coverage selection (WCS), on Qwen2.5-VL-7B at $9.9\%$ retention. RESTORE's merging improves DivPrune on VQA but reduces WCS
accuracy and degrades grounding with both selectors, since it replaces part of the selected tokens
with merged tokens whose features mix distant positions. Our bias leaves the selected tokens and
their features unchanged and only restores their attention share, improving both selectors on both
tasks, by $22.4$ and $40.0$ points on grounding.
Detailed results are provided in Appendix~\ref{sec:appendix:restore}.
\par
\endgroup

\begin{figure}[!t]
  \centering
  \includegraphics[width=\linewidth]{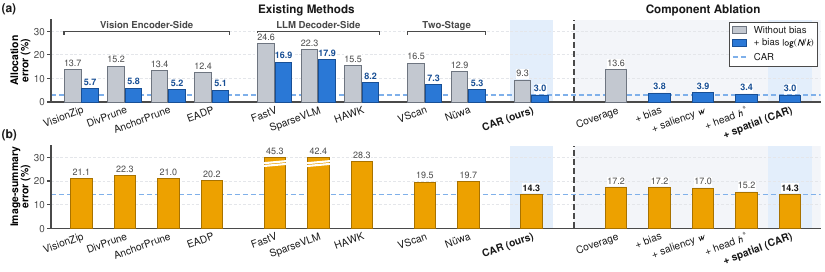}
  \caption{\textbf{The two errors of Eq.~\ref{eq:decomp} on grounding} for existing methods and the
  ablation of Table~\ref{tab:abl-comp} (Qwen2.5-VL-7B, $9.9\%$). Each bar is
  $\sum\lVert\mathcal{E}\rVert^{2}/\sum\lVert O\rVert^{2}$ (\%) over layers $13$--$19$; lower is
  better.}
  \label{fig:error-ablation}
\end{figure}

\paragraph{Error decomposition.}
Figure~\ref{fig:error-ablation} measures the two errors of Eq.~\ref{eq:decomp} on grounding; the
  measurement protocol is described in Appendix~\ref{sec:appendix:error-decomp}. Compared with
  existing methods, CAR has both the lowest image summary error and the lowest allocation error. The
  bias lowers the allocation error of each method by restoring the visual attention share. In the
  component ablation, the bias lowers the allocation error from $13.6$ to $3.8$ and leaves the summary
  error unchanged, whereas the selection components lower the summary error from $17.2$ to $14.3$
  with little change in the allocation error, so each component acts on the error term it targets.

\begingroup

\section{Related Work}
\label{sec:related}

\paragraph{Visual Token Pruning.}
Training-free visual token pruning can be categorized by its operational stage.
Encoder-side methods select tokens on the output of the vision encoder, before the tokens enter the
language model~\citep{yang2025visionzip,alvar2025divprune,liu2026hiprune,zhangcdpruner,oh2026anchorprune,wang2026eadp,tan2026idpruner,du2026coin,baek2026agilepruner,deng2025scope,chen2026spider}.
Decoder-side methods prune inside the language model, typically guided by the attention from the
text tokens~\citep{chen2024fastv,zhang2024sparsevlm,xing2024pdrop,wen2025dart,zhu2026hawk,zhang2026d2pruner}.
Two-stage methods combine both, selecting tokens on the encoder side and pruning them further inside
the language model~\citep{zhang2025vscan,huang2026nuwa,wang2026mico,li2026sinkpruner}.

\paragraph{Attention Calibration.}
Compensating for the attention of removed tokens originates in token merging.
ToMe~\citep{bolya2023tome} adds $\log s$ to the logit of a merged token,
where $s$ is the number of original tokens it represents.
RESTORE~\citep{cho2026restore} retains selected tokens and merges
the remaining tokens into representatives, then calibrates attention
using group size and relative distance.
ERA~\citep{wang2026era} merges discarded tokens into selected anchors
and adds a logit bias based on saliency-weighted group sizes.
CAR leaves retained features unchanged, with biases determined by
the retention ratio or token counts from coverage assignments.
A detailed review is provided in Appendix~\ref{sec:appendix:related}.
\par
\endgroup

\section{Conclusion}
\label{sec:conclusion}

Under aggressive pruning, existing visual token pruning methods degrade substantially, especially
on the grounding task. We analyze pruning through the shift it causes in the decoder output, and decompose this
shift exactly into an image summary error and an attention allocation error, which correspond to
the information loss and the attention distortion caused by pruning. Based on this decomposition,
we propose CAR, which reduces the image summary error with weighted coverage selection guided by a
head probe, and the attention allocation error with a logit bias on the retained visual tokens that
restores the attention share the image held before pruning. Since the bias does not change which tokens are retained, it can also be applied directly
to existing methods. Across four backbones and VQA, grounding, and video understanding, CAR
consistently achieves state-of-the-art performance.

\subsection*{Reproducibility statement}

The proofs and their assumptions are given in Appendix~\ref{sec:appendix:theory}, the packages,
weights and settings shared by all experiments in Appendix~\ref{sec:appendix:impl-common}, the
protocol of the head probe in Appendix~\ref{sec:appendix:probe}, and the configuration of each
backbone in Appendices~\ref{sec:appendix:impl-backbones}
to~\ref{sec:appendix:results-llava}.

\subsection*{AI use statement}

In this work, we used generative AI tools to assist in proving the mathematical claims in
Appendix. Additionally, we used generative AI tools to aid and polish the
  writing of the paper, and for retrieval and discovery, such as finding related work. We reviewed all
  AI-assisted content: every proof was checked by the authors, the text was reviewed and edited by the
  authors, and every retrieved reference was verified against its original publication. We take
  responsibility for the final content of this work, including text, claims and artifacts produced
  with the aid of generative AI.

\bibliography{references}
\bibliographystyle{references}

\clearpage
\appendix
\section*{Appendix}

\section{Related Work}
\label{sec:appendix:related}

\subsection{Visual Token Pruning}
\label{sec:appendix:related-pruning}

Visual token pruning shortens the visual sequence of a multimodal model, and existing approaches can
be classified along several dimensions. Based on their training requirements, token reduction methods
fall into \emph{training-based} and \emph{training-free} approaches. Training-based methods, such as
TOP-RL~\citep{wang2026toprl}, IF-Prune~\citep{sun2026ifprune} and
QuietPrune~\citep{gao2026quietprune}, adapt the model to the shortened sequence through gradient
updates. Training-free methods operate at inference time without modifying model parameters, which
makes them directly applicable to off-the-shelf models; our comparison focuses on training-free
methods. Regarding the stage at which pruning is applied (or at which the pruning signal is computed), they
are \emph{encoder-side}, such as
VisionZip~\citep{yang2025visionzip}, DivPrune~\citep{alvar2025divprune},
CDPruner~\citep{zhangcdpruner}, HiPrune~\citep{liu2026hiprune}, AnchorPrune~\citep{oh2026anchorprune},
EADP~\citep{wang2026eadp}, SCOPE~\citep{deng2025scope}, CoIn~\citep{du2026coin},
SCoRe~\citep{xu2026score}, SHARP~\citep{ma2026sharp}{ and SPIDER~\citep{chen2026spider}}; \emph{decoder-side}, such as
FastV~\citep{chen2024fastv}, SparseVLM~\citep{zhang2024sparsevlm}, PDrop~\citep{xing2024pdrop},
HAWK~\citep{zhu2026hawk}, PriorTR~\citep{chen2026accelerating}, PLPHP~\citep{meng2025plphp},
HSSP~\citep{lo2026hssp} and D\textsuperscript{2}Pruner~\citep{zhang2026d2pruner}; or \emph{two-stage}, such as VScan~\citep{zhang2025vscan},
N\"uwa~\citep{huang2026nuwa}, CLASP~\citep{dang2026clasp}{,
MiCo~\citep{wang2026mico} and SinkPruner~\citep{li2026sinkpruner}}. Each method further involves two
decisions: (1) \emph{identifying which tokens to retain}, based on their \emph{importance}
(FastV, SparseVLM, VisionZip, HAWK), their \emph{mutual similarity} (DivPrune, DART~\citep{wen2025dart}),
their \emph{spatial position} (N\"uwa, IVC-Prune~\citep{sun2026ivc}), or a combination of these signals
(CDPruner, AnchorPrune, D\textsuperscript{2}Pruner,
IDPruner~\citep{tan2026idpruner}, EADP, SCOPE, CoIn, SCoRe); and (2) \emph{handling the discarded tokens},
whose useful features can be aggregated into the retained ones, as in ToMe~\citep{bolya2023tome},
PruMerge+~\citep{shang2025prumerge+}, RESTORE~\citep{cho2026restore} and ERA~\citep{wang2026era}, or
which can be reused at a later stage, as in Reroute~\citep{yang2026reroute}.

\paragraph{Encoder-side pruning.} These methods select tokens before the language model. DivPrune
selects a maximally diverse subset based on the similarity between visual features. VisionZip and
HiPrune use the visual attention from the vision encoder as the importance signal, VisionZip keeping
the dominant tokens and merging the remaining ones, and HiPrune organizing the attention
hierarchically across encoder stages.
{SPIDER~\citep{chen2026spider} combines middle- and deep-layer visual features for
token selection with adaptive decoder sub-layer skipping.}
CDPruner, AnchorPrune and EADP score tokens by their relevance to the instruction and balance it
against diversity: CDPruner maximizes conditional diversity through a determinantal point process,
AnchorPrune protects a relevance anchor and expands it with complementary context, and EADP denoises
the relevance score with statistical entropy. The methods that rely on the encoder attention,
however, largely overlook the varying importance of the attention heads, assuming that all heads
contribute equally to interpreting visual information.

\paragraph{Decoder-side pruning.} FastV pioneered the approach of performing token pruning within
language models based on text--visual attention scores, providing guidance for subsequent studies
such as SparseVLM and PyramidDrop. These methods progressively remove less informative tokens
according to the attention scores. Since the text--visual attention is biased, PriorTR and
D\textsuperscript{2}Pruner focus on rectifying the attention signal, PriorTR by separating the
task-conditioned attention from a model-induced prior and D\textsuperscript{2}Pruner by debiasing
the positional bias of the attention scores. DART relies purely on feature similarity: it selects a
small set of pivot tokens and then removes the tokens that duplicate them. HAWK weights the
text-guided attention by head importance estimated offline, by ablating each head and measuring the
resulting accuracy drop on benchmark datasets.

\paragraph{Two-stage pruning.} These methods combine a selection in the vision encoder with a second
pruning step inside the language model. VScan merges tokens during visual encoding through
complementary global and local scans and then prunes at intermediate decoder layers. N\"uwa
partitions the token grid and aggregates the discarded tokens into their spatial neighbours, so that
the spatial reference frame on which grounding relies is preserved, and refines the selection with a
text-guided step inside the decoder.
{MiCo~\citep{wang2026mico} uses weighted coverage for encoder-side candidate
selection and decoder-side task-aware refinement.}
{SinkPruner~\citep{li2026sinkpruner} aggregates high-norm outliers and selects
informative visual tokens before text-guided decoder pruning.}

\subsection{Attention Calibration}
\label{sec:appendix:related-calibration}

Compensating for the attention of removed tokens originates in token merging.
ToMe~\citep{bolya2023tome} merges similar tokens inside a vision transformer and averages their
features; it then adds $\log s$ to the logit of the merged token, where $s$ is the number of original
tokens it represents, so that the merged token receives attention in proportion to them. RESTORE and
ERA transfer this idea to multimodal models. RESTORE~\citep{cho2026restore} is applied on top of an
existing pruning method and splits its budget in half: one half is the tokens that method selects,
and the other half is formed by merging the tokens it discards. The logit of each merged token is
raised by the logarithm of the number of tokens it has absorbed, scaled by a distance factor that
counteracts the decay of attention induced by RoPE. ERA~\citep{wang2026era} selects its own anchors, recycles the pruned tokens into them by moving each
anchor towards the centroid of its cluster, and raises its logit by the logarithm of the
saliency-weighted size of that cluster. In all three the bias is attached to the tokens that
absorbed the discarded ones, and its magnitude follows the number of tokens absorbed.

These calibrations are developed within a merging framework, so the features of the retained tokens
are modified. Our bias instead only restores the visual attention share and does not change the
features of the retained tokens. Our uniform form depends only on the retention
ratio, so it can be applied on top of any pruning method. Our per-cell form recovers the same
$\log s$ correction, with the assignment from the discarded tokens to the retained ones following
naturally from our coverage selection.

Adjusting the attention that text pays to the image has also been studied without pruning.
  PAI~\citep{liu2024paying} amplifies the attention weights on image tokens at inference to counter
  text inertia, and VAR~\citep{kang2025see} redistributes the attention absorbed by visual sink
  tokens to the remaining visual tokens in image-centric heads. Unlike PAI and VAR, our bias has a
  defined target, the share the image held before pruning, and its magnitude follows from the
  retention ratio rather than a tuned factor.

\section{Derivations and Proofs}
\label{sec:appendix:theory}

\begin{table}[t]
  \centering
  \scriptsize
  \setlength{\tabcolsep}{3pt}
  \caption{Summary of notation used in the derivations of Section~\ref{sec:method}.}
  \label{tab:notation}
  \begin{tabularx}{\linewidth}{@{}l>{\raggedright\arraybackslash}X@{\hspace{3mm}}l>{\raggedright\arraybackslash}X@{}}
    \toprule
    \textbf{Symbol} & \textbf{Description} & \textbf{Symbol} & \textbf{Description} \\
    \midrule
    $T,V,S$ & Text, visual and retained tokens
      & $w_i$ & Saliency of token $i$ \\
    $N,k$ & Number of visual tokens; number retained
      & $s_{ij}$ & Similarity between tokens $i$ and $j$ \\
    $z_u,\,e^{z_u}$ & Attention logit of token $u$ and its unnormalized mass
      & $C_j$ & Tokens assigned to the retained token $j$ \\
    $p_u,p'_u$ & Attention weight on $u$, before and after pruning
      & $b_j,\,m_j$ & Logit bias on $j$; its mass factor $m_j=e^{b_j}$ \\
    $\sigma,\sigma'$ & Share of attention on the image, before and after pruning
      & $Z_T,Z_V$ & Total attention mass of the text and of the image \\
    $\bar v_{\mathrm{img}},\bar v_S,\bar v_{\mathrm{txt}}$ & Summaries of $V$, of $S$ and of the text
      & $\rho$ & Fraction of the visual mass on the retained tokens \\
    \bottomrule
  \end{tabularx}
  \vspace{-2mm}
\end{table}

\subsection{Notation and the Two-Group Form of the Attention Output}
\label{sec:appendix:notation}

This section derives Eq.~\ref{eq:blend}. The attention output $O$ is an
attention-weighted average of the value vectors of all visual and text tokens. We rewrite
this average in two stages. The first stage averages the values inside each modality, under
the attention renormalized within that modality, and returns one summary vector per
modality. The second stage mixes the two summaries, weighting each by the attention mass its
modality holds.

\paragraph{Notation.}
Throughout we fix one decoder layer, one attention head and one text query, and suppress
the corresponding indices. Let $T$ index the text tokens that precede the query and $V$ the $N$
visual tokens,
with attention logits $z_u$ and values $v_u$, and let $S\subseteq V$ with $|S|=k$ be the
visual tokens retained by pruning. Write
\begin{equation}
Z_T=\sum_{t\in T}e^{z_t},
\qquad
Z_V=\sum_{i\in V}e^{z_i},
\qquad
Z_S=\sum_{j\in S}e^{z_j},
\qquad
\rho=\frac{Z_S}{Z_V},
\end{equation}
so that $\rho\in(0,1]$ is the fraction of the visual softmax mass carried by $S$.

\paragraph{Step 1 (split the sum at the modality boundary).}
Writing $Z=Z_T+Z_V$ for the full softmax denominator, Eq.~\ref{eq:attn} splits as
\begin{equation}
O=\sum_{u\in T\cup V}\frac{e^{z_u}}{Z}\,v_u
 =\sum_{i\in V}\frac{e^{z_i}}{Z}\,v_i \;+\; \sum_{t\in T}\frac{e^{z_t}}{Z}\,v_t .
\end{equation}

\paragraph{Step 2 (write each weight as a group share times a within-group weight).}
A token's weight in the whole sequence factors into the share held by its group and its
weight inside that group:
\begin{equation}
\frac{e^{z_i}}{Z}=\frac{Z_V}{Z}\cdot\frac{e^{z_i}}{Z_V},
\qquad
\frac{e^{z_t}}{Z}=\frac{Z_T}{Z}\cdot\frac{e^{z_t}}{Z_T}.
\end{equation}
The group mass does not depend on the summation index, so it leaves the sum:
\begin{equation}
O=\frac{Z_V}{Z}\underbrace{\sum_{i\in V}\frac{e^{z_i}}{Z_V}\,v_i}_{\textstyle\bar v_{\mathrm{img}}}
 \;+\;\frac{Z_T}{Z}\underbrace{\sum_{t\in T}\frac{e^{z_t}}{Z_T}\,v_t}_{\textstyle\bar v_{\mathrm{txt}}} .
\end{equation}
Each brace is a convex combination of that group's values: its weights are non-negative and
sum to one by construction ($\sum_{i\in V}e^{z_i}/Z_V=1$, and likewise for $T$), so
$\bar v_{\mathrm{img}}$ and $\bar v_{\mathrm{txt}}$ are exactly the attention-weighted mean
values of the two groups.

\paragraph{Step 3 (identify the two prefactors).}
Since $Z=Z_T+Z_V$,
\begin{equation}
\frac{Z_V}{Z}=\sum_{i\in V}\frac{e^{z_i}}{Z}=\sum_{u\in V}p_u=\sigma,
\qquad
\frac{Z_T}{Z}=\frac{Z-Z_V}{Z}=1-\sigma,
\end{equation}
and substituting into Step 2 gives the first line of Eq.~\ref{eq:blend},
\begin{equation}
O=\sigma\,\bar v_{\mathrm{img}}+(1-\sigma)\,\bar v_{\mathrm{txt}} .
\end{equation}

\paragraph{The same rewriting after pruning.}
Pruning restricts the sum to $T\cup S$ and adds the bias $b_j$ to the visual logits, with
$b_t=0$ on the text side. The visual part of the denominator becomes
$Z_S^{b}=\sum_{j\in S}e^{z_j+b_j}$, and the text part remains $Z_T$. The three steps go
through unchanged and give the second line of Eq.~\ref{eq:blend},
\begin{equation}
O'=\sigma'\,\bar v_S+(1-\sigma')\,\bar v_{\mathrm{txt}},
\qquad
\sigma'=\frac{Z_S^{b}}{Z_T+Z_S^{b}},
\qquad
\bar v_S=\frac{1}{Z_S^{b}}\sum_{j\in S}e^{z_j+b_j}v_j .
\end{equation}
The text summary $\bar v_{\mathrm{txt}}$ is the same vector as in the first line, since its
weights are $e^{z_t}/Z_T$ and pruning changes neither $Z_T$ nor the text logits.

\subsection{The Coverage Bound}
\label{sec:appendix:proof}

This section proves the bound quoted in Section~\ref{sec:method:select}: the image summary
error is controlled by how well the retained set covers the whole token set, so that
maximizing coverage minimizes an upper bound on the error.

\paragraph{Setup.}
Write $a_i=e^{z_i}/Z_V$ for the attention on visual token $i$ renormalized within the image,
as in Appendix~\ref{sec:appendix:notation}, so that $\sum_{i\in V}a_i=1$ and
$\sum_{j\in S}a_j=\rho$. Let
$\pi(i)=\arg\max_{j\in S}\cos(v_i,v_j)$ for $i\notin S$ and $\pi(j)=j$ for $j\in S$, let
$A_j=\sum_{i:\pi(i)=j}a_i$ be the attention mass of the tokens that $j$ represents, so that
$\sum_{j\in S}A_j=1$, and write $c_i=\cos(v_i,v_{\pi(i)})$. The objective
\begin{equation}
F_v(S)=\sum_{i\in V} a_i \max_{j\in S}\cos(v_i,v_j)
\label{eq:fv}
\end{equation}
is a facility-location objective: the retained tokens are the facilities, all $N$ visual
tokens are the clients, and each client is weighted by the attention it receives.

We restate Proposition~\ref{prop:bound}: if the value vectors have equal norm $R$, and if each
retained token carries the attention mass of the tokens it represents, that is,
$\bar v_S=\sum_{j\in S}A_jv_j$, then
\begin{equation}
\big\lVert\bar v_S-\bar v_{\mathrm{img}}\big\rVert
\;\le\;\sqrt{2}\,R\,\sqrt{1-\rho}\,\sqrt{1-F_v(S)} .
\label{eq:bound-sharp}
\end{equation}

\paragraph{Step 1 (replacement view).}
Splitting $\bar v_{\mathrm{img}}$ into its retained and discarded parts and using the
assumption on $\bar v_S$,
\begin{align}
\bar v_S-\bar v_{\mathrm{img}}
&=\sum_{j\in S}A_jv_j-\sum_{j\in S}a_jv_j-\sum_{i\notin S}a_iv_i\\
&=\sum_{j\in S}\sum_{i\notin S:\,\pi(i)=j}a_iv_j-\sum_{i\notin S}a_iv_i
 \;=\;\sum_{i\notin S}a_i\big(v_{\pi(i)}-v_i\big),
\label{eq:step1}
\end{align}
so the error is the weighted sum of the replacements, one per discarded token.

\paragraph{Step 2 (triangle inequality).}
Since the weights are non-negative,
\begin{equation}
\big\lVert\bar v_S-\bar v_{\mathrm{img}}\big\rVert\le\sum_{i\notin S}a_i\lVert v_{\pi(i)}-v_i\rVert .
\label{eq:step2}
\end{equation}

\paragraph{Step 3 (equal norms turn distance into cosine).}
With $\lVert v_i\rVert=\lVert v_{\pi(i)}\rVert=R$,
\begin{equation}
\lVert v_i-v_{\pi(i)}\rVert^{2}=2R^{2}-2\langle v_i,v_{\pi(i)}\rangle=2R^{2}(1-c_i),
\qquad\text{hence}\qquad
\lVert v_i-v_{\pi(i)}\rVert=\sqrt{2}\,R\,\sqrt{1-c_i}.
\label{eq:step3}
\end{equation}

\paragraph{Step 4 (Cauchy--Schwarz).}
Splitting each weight as $a_i=\sqrt{a_i}\cdot\sqrt{a_i}$ and using
$\sum_{i\notin S}a_i=1-\rho$,
\begin{equation}
\sum_{i\notin S}a_i\sqrt{1-c_i}
=\sum_{i\notin S}\sqrt{a_i}\cdot\sqrt{a_i(1-c_i)}
\le\sqrt{1-\rho}\;\Big(\sum_{i\notin S}a_i(1-c_i)\Big)^{1/2}.
\label{eq:step4}
\end{equation}

\paragraph{Step 5 (identify the coverage objective).}
A retained token covers itself with $\cos=1$, and a discarded token is covered by
$\pi(i)$, which is the $\arg\max$ of the same cosine that defines $F_v$, so
\begin{equation}
F_v(S)=\rho+\sum_{i\notin S}a_ic_i ,
\qquad\text{and therefore}\qquad
\sum_{i\notin S}a_i(1-c_i)=(1-\rho)-\big(F_v(S)-\rho\big)=1-F_v(S).
\label{eq:step5}
\end{equation}
Combining Steps 1 to 5 gives Eq.~\ref{eq:bound-sharp}.

\paragraph{Tightness.}
The bound is loose: Step 2 is an equality only when the replacement vectors all point the
same way, and Step 4 only when every discarded token is represented equally well. It
therefore separates a good retained set from a bad one, but does not rank two sets that are
close.

\paragraph{Scope.}
  Since $\rho\ge0$, the right-hand side is at most $\sqrt{2}R\sqrt{1-F_v(S)}$; the form
  quoted in Section~\ref{sec:method:select} keeps the factor $\sqrt{1-\rho}$, which shows that a set keeping more
  attention mass tightens the bound as well. Equal norms enter only in Step 3. Without
  them, Step 3 gains the term $(\lVert v_i\rVert-\lVert v_{\pi(i)}\rVert)^2$, and the same steps give
  the bound with $R$ replaced by $R_{\max}=\max_u\lVert v_u\rVert$ plus an additive term
  $(1-\rho)(R_{\max}-R_{\min})$, which vanishes when the norms are equal. In practice,
  the selector computes this objective from encoder features and encoder saliency, since the
  decoder values and attention shares are not available before the tokens enter the decoder. The
  assumption on $\bar v_S$ is exactly what the per-cell bias realizes; with the uniform bias, it holds only approximately.

\subsection{Submodularity of the Selection Objective}
\label{sec:appendix:submod}

This section proves that the objective of Eq.~\ref{eq:fl-ours} is monotone and submodular, so
that the greedy selection of Eq.~\ref{eq:greedy} comes with the classical $1-1/e$ guarantee.

\paragraph{Setup.}
Let $\phi(i,j)\ge0$ be any non-negative similarity between tokens, let $w_i\ge0$ be any
weights, and set
\begin{equation}
F(S)=\sum_{i=1}^{N}w_i\,m_i(S),
\qquad
m_i(S)=\max_{j\in S}\phi(i,j),
\qquad
m_i(\emptyset)=0 .
\label{eq:generic-fl}
\end{equation}

\paragraph{Monotonicity.}
A set function is monotone if $F(A)\le F(B)$ whenever $A\subseteq B$. For any such $A$ and
$B$, the set $\{\phi(i,j):j\in A\}$ is contained in $\{\phi(i,j):j\in B\}$, and a maximum
over a larger set is no smaller, so $m_i(A)\le m_i(B)$ for every $i$. Summing with
non-negative weights preserves the inequality, hence $F(A)\le F(B)$.

\paragraph{Submodularity.}
A set function is submodular if it has diminishing returns, that is,
$F(A\cup\{u\})-F(A)\ge F(B\cup\{u\})-F(B)$ for every $A\subseteq B$ and every $u\notin B$.
The gain of a single client is
\begin{equation}
m_i(S\cup\{u\})-m_i(S)=\big(\phi(i,u)-m_i(S)\big)_+ .
\label{eq:gain}
\end{equation}
Monotonicity gives $m_i(A)\le m_i(B)$, and $(\cdot)_+$ is non-decreasing, so
$\big(\phi(i,u)-m_i(A)\big)_+\ge\big(\phi(i,u)-m_i(B)\big)_+$: the more a client is already
covered, the less a new facility can add. Summing over $i$ with non-negative weights gives
the claim.

\paragraph{Our objective.}
  Both similarities we use are non-negative: $[\cos(x_i,x_j)]_+$ of Eq.~\ref{eq:fl-ours}, and its
  product with the positive Gaussian kernel of Eq.~\ref{eq:fl-kernel}. With $w_i$ the encoder
  saliency, both objectives are instances of Eq.~\ref{eq:generic-fl}.

\paragraph{Approximation guarantee.}
Maximizing a monotone submodular function under a cardinality constraint is NP-hard, but
greedy selection is near-optimal: writing $S_{\mathrm{greedy}}$ for the set it returns and
$S_{\mathrm{opt}}$ for the optimum of the same size,
\begin{equation}
F(S_{\mathrm{greedy}})\;\ge\;\Big(1-\tfrac1e\Big)F(S_{\mathrm{opt}})\;\approx\;0.632\,
F(S_{\mathrm{opt}})
\label{eq:greedy-guarantee}
\end{equation}
\citep{nemhauser1978analysis}. The step of Eq.~\ref{eq:greedy} adds the token with the
largest $\sum_i w_i\big(\phi(i,j)-m_i(S)\big)_+$, which by Eq.~\ref{eq:gain} is exactly the
marginal gain $F(S\cup\{j\})-F(S)$ the theorem requires, so the guarantee applies to the
selection we run.

\subsection{Properties of the Uniform Bias}
\label{sec:appendix:derivation}

This section establishes the properties of the uniform bias of
Section~\ref{sec:method:bias}.

\paragraph{A uniform bias does not move the visual summary.}
By Appendix~\ref{sec:appendix:notation}, the image summary after pruning $\bar v_S$ is the
average of the retained values under the weights renormalized inside the image,
\begin{equation}
\bar v_S=\sum_{j\in S}\frac{e^{z_j+b_j}}{\sum_{j'\in S}e^{z_{j'}+b_{j'}}}\,v_j .
\end{equation}
With the same $b_j=b$ on every $j\in S$, the factor $e^{b}$ cancels between the numerator and
the denominator,
\begin{equation}
\frac{e^{z_j+b}}{\sum_{j'\in S}e^{z_{j'}+b}}
=\frac{e^{b}e^{z_j}}{e^{b}Z_S}
=\frac{e^{z_j}}{Z_S},
\end{equation}
so every weight, and therefore $\bar v_S$, is independent of $b$. A uniform bias changes
$\sigma'$ alone and acts on $\mathcal{E}_{\mathrm{alloc}}$ alone.

\paragraph{Remark on the substitution $\mathbb{E}[\rho]=k/N$.}
  For a given query the exact constant is $b^{\star}=-\log\rho$, and $\rho$ varies from query to
  query. We cannot evaluate it, so we substitute $\mathbb{E}[\rho]=k/N$ and use $b=\log(N/k)$ for
  every query. The restored mass is $e^{b}\rho=\rho N/k$, which equals the exact value $1$
  in expectation. The share, however, is a concave function of this mass,
  \begin{equation}
  \sigma'=\frac{x}{Z_T/Z_V+x},\qquad x=e^{b}\rho ,
  \end{equation}
  so by Jensen's inequality $\mathbb{E}[\sigma']\le\sigma'(\mathbb{E}[x])=\sigma$: the bias restores
  the share less than the exact solution does on average. The gap is zero when $\rho$ is the same
  for every query and grows with its variance.

\paragraph{Remark on the identity $\mathbb{E}[\rho]=k/N$.}
The identity assumes that the visual tokens are exchangeable, which the selector does not
respect, since it retains the tokens that carry the most saliency. Where the retained set
holds more than $k/N$ of the visual attention, the exact constant $-\log\rho$ is smaller than
the $\log(N/k)$ we apply, so the bias restores more share than pruning removed. The convexity
of $-\log$ and the failure of exchangeability act in opposite directions, and which one
dominates changes with the layer. The share measured in
Appendix~\ref{sec:appendix:mechanism} exceeds the unpruned level in the early layers and falls
below it in the later ones.

\subsection{The Per-Cell Bias}
\label{sec:appendix:percell}

This section derives the per-cell bias of Section~\ref{sec:method:bias}.

\paragraph{Setup.}
We keep the notation of Appendix~\ref{sec:appendix:proof}: $a_i=e^{z_i}/Z_V$,
$\pi(i)=\arg\max_{j\in S}\cos(v_i,v_j)$ and $A_j=\sum_{i:\pi(i)=j}a_i$. The selector assigns
each discarded token to the retained token most similar to it, and those assigned to $j$ form
its cell $C_j$. The bound defines $\pi$ by the cosine between value vectors, while the
selector assigns by the cosine between encoder features, and we use $C_j$ as an approximation
of $\pi^{-1}(j)$ throughout.

\paragraph{The per-cell constraint.}
The mass condition of Eq.~\ref{eq:mass-condition} is one equation in $k$ unknowns. We impose
the $k$ constraints that give each retained token the mass of the tokens it represents,
\begin{equation}
e^{z_j}m_j \;=\; e^{z_j}+\sum_{i\in C_j}e^{z_i} ,
\qquad j\in S ,
\label{eq:percell-constraint}
\end{equation}
where the left-hand side is the mass of $j$ after the bias and the right-hand side is the mass
of $j$ together with that of its cell. The $k$ constraints determine every unknown,
$m_j=1+\big(\sum_{i\in C_j}e^{z_i}\big)/e^{z_j}$.

\paragraph{The constraint solves the mass condition.}
Summing Eq.~\ref{eq:percell-constraint} over $j\in S$ gives
\begin{equation}
\sum_{j\in S}e^{z_j}m_j=\sum_{j\in S}e^{z_j}+\sum_{j\in S}\sum_{i\in C_j}e^{z_i}=Z_V ,
\end{equation}
where the first sum is the mass of the retained tokens and the second is the mass of the
discarded ones, since the cells partition them. Together they are the mass of all $N$ visual
tokens, so Eq.~\ref{eq:mass-condition} holds.

\paragraph{The weights are the ones the bound assumes.}
Proposition~\ref{prop:bound} holds under the hypothesis that each retained token carries the
attention of the tokens it represents, that is, $\bar v_S=\sum_{j\in S}A_jv_j$. The per-cell
constraint produces exactly these weights. The biased masses sum to $Z_V$, so the visual
weight of a retained token is
\begin{equation}
\frac{e^{z_j}m_j}{\sum_{j'\in S}e^{z_{j'}}m_{j'}}
=\frac{e^{z_j}+\sum_{i\in C_j}e^{z_i}}{Z_V}
=a_j+\sum_{i\in C_j}a_i
=A_j .
\end{equation}
The per-cell bias therefore makes the hypothesis of the bound hold.

\paragraph{Counting inside a cell.}
The sum $\sum_{i\in C_j}e^{z_i}$ on the right-hand side of Eq.~\ref{eq:percell-constraint} is
not available at run time. Setting $e^{z_i}\approx e^{z_j}$ for $i\in C_j$ turns the ratio
into a count,
\begin{equation}
m_j=1+\frac{\sum_{i\in C_j}e^{z_i}}{e^{z_j}}\;\approx\;1+|C_j| .
\end{equation}
The substitution assumes exchangeability inside a cell, whose members the selector grouped by
similarity to their representative.

\paragraph{The capped bias.}
The count overstates the mass a cell carried. The representative is a token the selector
retained for its saliency and the members are tokens it discarded, so $e^{z_i}<e^{z_j}$ is the
common case and a large cell is the one that overstates the most. We therefore cap the count
at its mean, which is $N/k$ since $\sum_{j\in S}(1+|C_j|)=N$,
\begin{equation}
m_j \;=\; \min\Big\{1+|C_j|,\ \frac{N}{k}\Big\} ,
\qquad
b_j \;=\; \min\Big\{\log\big(1+|C_j|\big),\ \log\frac{N}{k}\Big\} .
\end{equation}
The cap leaves a below-average cell at its per-cell value and returns an above-average one to
the mean. The biased masses then sum to less than $Z_V$, so the capped form restores less
share than pruning removed. The capped $m_j$ are flatter than those of the uncapped form, so the cap perturbs $\bar v_S$
less, which is the safer choice when the count is unreliable.

\section{Experimental Details and Results}
\label{sec:appendix:impl}
\label{sec:appendix:results}

\subsection{Algorithm}
\label{sec:appendix:algorithm}

To more clearly illustrate the proposed method, particularly the greedy coverage selection and the
form of the bias it produces, we provide the pseudocode of CAR in Algorithm~\ref{alg:main}.

\begin{algorithm}[h]
\caption{CAR with Weighted Coverage Selection and Attention Reallocation}
\label{alg:main}
\small
\begin{algorithmic}[1]
\Require
\Statex \quad Encoder features $X\in\mathbb{R}^{N\times d}$ (visual tokens entering the language model)
\Statex \quad Saliency $w\in[0,1]^{N}$ from the probed head $h^{*}$
\Statex \quad Grid positions $g_1,\dots,g_N$ of the visual tokens
\Statex \quad Target number of retained tokens $k$
\Statex \quad Task family (grounding or question answering)
\Ensure
\Statex \quad Retained indices $S\subset\{1,\dots,N\}$, $|S|=k$ (positions of the selected tokens)
\Statex \quad Logit bias $b\in\mathbb{R}^{k}$ (one value per retained token)
\Statex \textit{Similarity between visual tokens}
\State $X \gets \mathrm{rownormalise}(X)$ \Comment{Unit-norm features}
\State $M \gets [\,XX^{\top}]_+$ \Comment{Clipped cosine similarity, Eq.~\ref{eq:fl-ours}}
\If{grounding}
  \State $K_{ij} \gets \exp(-\lVert g_i-g_j\rVert^{2}/2(N/k))$ \Comment{Eq.~\ref{eq:fl-kernel}}
  \State $M \gets M \odot K$ \Comment{Coverage becomes local}
\EndIf
\Statex \textit{Greedy coverage selection}
\State $S \gets \emptyset$, \ $c \gets \mathbf{0}_N$ \Comment{$c_i$: how well token $i$ is covered}
\For{$t=1,\dots,k$}
  \State $j^{*} \gets \arg\max_{j\notin S} w^{\top}[\,M_{:,j}-c\,]_+$ \Comment{Largest gain, Eq.~\ref{eq:greedy}}
  \State $S \gets S\cup\{j^{*}\}$, \ $c \gets \max(c,\,M_{:,j^{*}})$ \Comment{Update coverage}
\EndFor
\Statex \textit{Cell of each retained token}
\State $C_j \gets \{\,i\notin S:\ j=\arg\max_{j'\in S} M_{ij'}\,\}$ \Comment{Closest survivor of each discarded token}
\Statex \textit{Attention reallocation}
\If{grounding}
  \State $b_j \gets \log(N/k)$ \Comment{Uniform bias, Eq.~\ref{eq:bias-uniform}}
\Else
  \State $b_j \gets \min\{\log(1+|C_j|),\log(N/k)\}$ \Comment{Capped per-cell bias, Eq.~\ref{eq:bias-capped}}
\EndIf
\State Add $b_j$ to the logits of $j\in S$ in every decoder layer during prefill
        \Comment{Restores the visual attention share, Eq.~\ref{eq:pruned}}
\State \Return $S$, $b$
\end{algorithmic}
\end{algorithm}

\subsection{Common Setup}
\label{sec:appendix:impl-common}

\paragraph{Environment.}
\label{sec:appendix:impl-env}

To ensure the reproducibility of the results presented in this paper, Table~\ref{tab:env} lists the main packages. Model weights are
\texttt{Qwen/Qwen2.5-VL-7B-Instruct}, \texttt{Qwen/Qwen3-VL-8B-Instruct},
\texttt{OpenGVLab/InternVL3\_5-8B-HF} and \texttt{liuhaotian/llava-v1.5-7b} from Hugging Face. 

\begin{table}[h]
  \centering
  \small
  \caption{Main packages of the three environments.}
  \label{tab:env}
  \begin{tabular}{lccc}
    \toprule
    Package & Qwen2.5-VL-7B & Qwen3-VL-8B / InternVL3.5-8B & LLaVA-1.5-7B \\
    \midrule
    python        & 3.10.4        & 3.10.4        & 3.10.4 \\
    torch         & 2.3.0+cu118   & 2.8.0+cu126   & 2.6.0+cu124 \\
    transformers  & 4.51.3        & 4.57.6        & 4.54.0 \\
    flash-attn    & 2.5.8         & 2.8.3         & -- \\
    accelerate    & 1.12.0        & 1.12.0        & 1.8.1 \\
    qwen-vl-utils & 0.0.14        & 0.0.14        & -- \\
    llava         & --            & --            & 1.2.2.post1 \\
    VLMEvalKit    & 0.1.0         & 0.1.0         & -- \\
    lmms-eval     & --            & 0.5.0 (video) & 0.3.0 \\
    \bottomrule
  \end{tabular}
\end{table}

\paragraph{Benchmarks.}
\label{sec:appendix:impl-benchmarks}

Grounding uses the eight splits of RefCOCO, RefCOCO+~\citep{yu2016modeling} and
RefCOCOg~\citep{mao2016generation}; on LLaVA-1.5-7B it uses the seven splits evaluated by
N\"uwa~\citep{huang2026nuwa} through lmms-eval (Appendix~\ref{sec:appendix:results-llava}). Question answering uses
TextVQA~\citep{singh2019towards}, ChartQA~\citep{masry2022chartqa},
AI2D~\citep{kembhavi2016diagram}, OCRBench~\citep{liu2024ocrbench},
HallusionBench~\citep{guan2024hallusionbench}, MME~\citep{fu2023mme} and
MMBench-EN/CN~\citep{liu2024mmbench}. Video understanding uses
Video-MME~\citep{fu2025video}, LongVideoBench~\citep{wu2024longvideobench} and
MVBench~\citep{li2024mvbench} on Qwen3-VL-8B at $32$ frames. In the question answering tables, Acc.\ is the mean of the eight benchmarks, with OCRBench divided by $10$ and the MME total score by $20$, and Rel.\ is the mean over benchmarks of the ratio to the unpruned score.

\paragraph{Implementation Details.}
\label{sec:appendix:impl-baselines}

For a fair comparison, we reproduce all baselines from their released code and evaluate them in
the same pipeline as our method. We align the token budget of all methods by the average number of
visual tokens per decoder layer,
\begin{equation}
\bar{k} = \frac{1}{L}\sum_{l=1}^{L} k_l ,
\label{eq:avg-budget}
\end{equation}
where $L$ is the number of decoder layers and $k_l$ is the number of visual tokens entering layer $l$.

\subsection{Saliency Source and Head Probe}
\label{sec:appendix:probe}

\paragraph{Probe Protocol.}
\label{sec:appendix:probe-protocol}

We detail below how the probe of Eq.~\ref{eq:head-probe} is run. Table~\ref{tab:probe-heads} lists
the source it selects for each backbone and task family.

\textit{Head probe on VQA.} We take the VQA task of Qwen2.5-VL-7B as an example, where $128$ visual
tokens are retained. The probe set consists of $500$ samples from each of ChartQA, OCRBench, TextVQA
and MMBench-EN, and the unpruned model is run over it once to give the
reference answers. The candidates are the $16$ heads of the final encoder block and their average.
We then run CAR once per candidate, with $w$ read from it, and score it by the exact-match rate of
its answers against the references. Since these rates are not on a common scale across benchmarks,
we rank the candidates within each benchmark and keep the one with the best mean rank as $h^{*}$. The VQA and video probes on Qwen3-VL-8B proceed in the
same way.

\textit{Head probe on grounding.} We take the grounding task of Qwen2.5-VL-7B as an example, where
$9.9\%$ of the visual tokens are retained. The probe set consists of $500$ images of RefCOCO val, over
which the unpruned model gives the reference boxes. We then run CAR once per candidate, again the
$16$ heads of the final encoder block and their average, and score it by the mean IoU of its boxes
against the references; the highest-scoring candidate is kept as $h^{*}$. The grounding probe on Qwen3-VL-8B
proceeds in the same way.

\textit{Layer and head probe on InternVL3.5-8B and LLaVA-1.5-7B.} We take the VQA task of InternVL3.5-8B as an example,
where $11.1\%$ of the visual tokens are retained. The probe set again consists of $500$ samples from
each of ChartQA, OCRBench, TextVQA and MMBench-EN, over which the unpruned model gives the reference
answers. The late layers of this encoder exhibit attention
sinks~\citep{darcet2024vision,luo2026sink}, where a few tokens absorb most of the attention, so the
final layer is not the best source and the probe covers the layers as well, in two steps. It first runs CAR once per encoder layer, with $w$ read
from that layer and averaged over its heads, and keeps the layer with the best mean rank; it then
repeats this over the heads of that layer and their average, which gives $h^{*}$. The grounding probe on
InternVL3.5-8B proceeds in the same way. On LLaVA-1.5-7B, which we evaluate only on grounding, the
probe likewise selects the layer and then the head, with the probe set and score of the grounding
probe above.

The probe is run once for each backbone and task family and needs one evaluation of the probe set
per candidate, together with one run of the unpruned model. On Qwen2.5-VL-7B this
takes about {$0.65$ hours on grounding and $1.0$ hour on VQA, measured on six RTX 3090 GPUs.} The source it selects is then fixed for all later
experiments.

\begin{table}[h]
  \centering
  \small
  \caption{Saliency source selected by the probe on each backbone and task family. The candidates are
  the $16$ heads of the final encoder block and their average; on InternVL3.5-8B and LLaVA-1.5-7B the probe selects the
  encoder layer as well.}
  \label{tab:probe-heads}
  \begin{tabular}{llcc}
    \toprule
    Backbone & Task & Encoder layer & Head \\
    \midrule
    \multirow{2}{*}{Qwen2.5-VL-7B}
      & VQA       & final & Head $2$ \\
      & Grounding & final & Head $13$ \\
    \midrule
    \multirow{3}{*}{Qwen3-VL-8B}
      & VQA       & final & Head $13$ \\
      & Grounding & final & Head $8$ \\
      & Video     & final & Head avg. \\
    \midrule
    \multirow{2}{*}{InternVL3.5-8B}
      & VQA       & Layer $10$ & Head $10$ \\
      & Grounding & Layer $12$ & Head avg. \\
    \midrule
    LLaVA-1.5-7B & Grounding & Layer $9$ & Head $15$ \\
    \bottomrule
  \end{tabular}
\end{table}

\paragraph{Attention Concentration in the InternVL3.5-8B Vision Encoder.}
\label{sec:appendix:internvl-attn}

Table~\ref{tab:internvl-attn} quantifies the attention sinks of the late layers of the
InternVL3.5-8B vision encoder. Per image tile, we average the attention each token receives over the
heads, sum it over the queries, and report the share taken by the class token together with the
share of patch attention taken by the top $1\%$ of patches.  
Up to layer $11$ both quantities stay at about $7\%$ or below; from layer $12$ on they grow steadily, and over the last eight layers the top $1\%$ of patches take up to a quarter of the patch attention.
This is exactly where the accuracy in
Figure~\ref{fig:phenomena}(f) begins to decline, since a few tokens absorb most of the attention and the
tokens the task needs are then hard to cover.

\begin{table}[h]
  \centering
  \small
  \caption{Attention concentration across layers of the InternVL3.5-8B vision encoder.}
  \label{tab:internvl-attn}
  \begin{tabular}{lcc}
    \toprule
    Layers & Class-token share (\%) & Top-$1\%$ patch share (\%) \\
    \midrule
    $0$--$11$  & $3.3$--$7.1$   & $2.4$--$6.3$ \\
    $12$       & $13.8$         & $6.7$ \\
    $13$--$15$ & $17.4$--$22.8$ & $12.2$--$13.5$ \\
    $16$--$23$ & $31.2$--$38.8$ & $17.9$--$26.8$ \\
    \bottomrule
  \end{tabular}
\end{table}

\paragraph{Reliability of the Probe.}
\label{sec:appendix:probe-reliability}

\begin{figure}[h]
  \centering
  \includegraphics[width=0.8\linewidth]{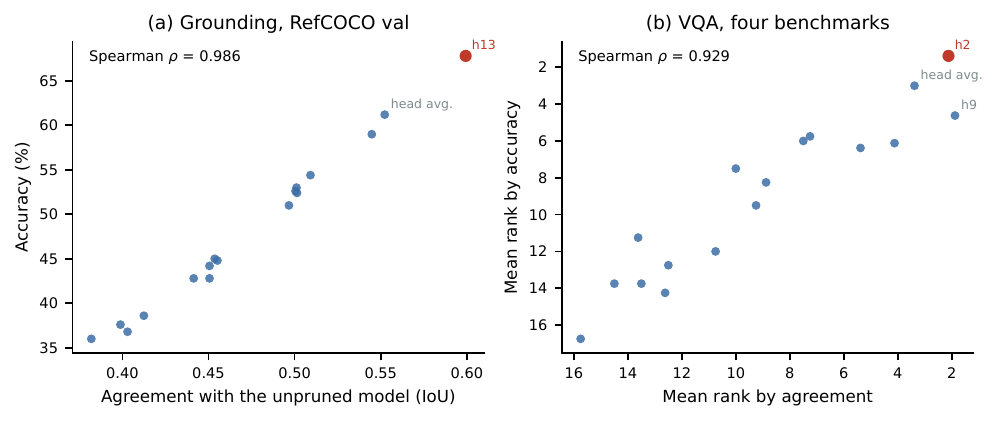}
  \caption{\textbf{Agreement with the unpruned model tracks accuracy across the candidate sources}
  (Qwen2.5-VL-7B, $16$ heads and their average; the selected source is in red). (a) Grounding at
  $9.9\%$, on RefCOCO val. (b) VQA at $128$ tokens, mean ranks over the four probe benchmarks.}
  \label{fig:probe}
\end{figure}

The probe selects the source whose outputs agree most with the unpruned model. To check its
reliability, Figure~\ref{fig:probe} compares this ranking with the ranking by labeled accuracy. On
grounding the two agree almost perfectly (Spearman $0.986$), and the probe's first choice, head
$13$, is also the most accurate. On VQA they correlate at $0.929$, and the two candidates the probe
ties at the top, heads $2$ and $9$, differ by $0.15$ points in accuracy. On both tasks the
label-free criterion therefore identifies the important head.

\paragraph{Probe on Held-Out Data.}
\label{sec:appendix:probe-heldout}
To test whether the selected source depends on the probe data, we run the probe on Qwen2.5-VL-7B
again on data disjoint from every evaluation set. For VQA, the probe set takes $500$ samples from each of DocVQA val,
InfographicVQA val, MMStar and RealWorldQA, none of which is among the eight VQA benchmarks. For
grounding, it takes $500$ referring expressions from the RefCOCO train split, one per image,
excluding every image used by the evaluation splits of RefCOCO, RefCOCO+ and RefCOCOg. All
other settings follow the probe protocol above (Appendix~\ref{sec:appendix:probe-protocol}). On both tasks the
held-out probe selects the same source as before, head $2$ on VQA and head
$13$ on grounding, and its ranking of the $17$ candidates agrees with that of the
original probe (Spearman $0.931$ and $0.956$). The source therefore
transfers across probe data.

\subsection{Qwen2.5-VL-7B}
\label{sec:appendix:impl-backbones}

\paragraph{Setup.}
We describe how our method is implemented on Qwen2.5-VL-7B. The saliency of visual token $i$ is
\begin{equation}
w_i=\frac{1}{4}\sum_{p\in\mathcal{P}_i}\sum_{q}A^{(h^{*})}_{qp},
\label{eq:saliency-qwen25}
\end{equation}
rescaled to $[0,1]$ within each image, where $A^{(h^{*})}$ is the attention map of head $h^{*}$ in
the last vision-encoder block, which uses full attention, and $\mathcal{P}_i$ is the set of four
patches merged into token $i$. The features $x_i$ are the merger outputs, which
are the vectors passed to the language model. For VQA, we pad each image to a square and resize it
to $1008{\times}1008$, giving $1{,}296$ visual tokens. For grounding, images keep Qwen's native
dynamic resolution, and budgets are given as fractions of the visual tokens. Retained tokens keep
their original M-RoPE positions.

\paragraph{VQA under an Extreme Budget.}
\label{sec:appendix:results-qwen25}
When about $90\%$ of the visual tokens are pruned, existing methods degrade severely on grounding
but largely preserve their performance on VQA. To examine VQA under more aggressive pruning,
Table~\ref{tab:qwen2.5-vl-32} compares CAR with the three strongest baselines when only $32$ tokens
are retained, a $97.5\%$ reduction. Even under this budget, the best baseline retains $70.9\%$ of the
unpruned performance, which indicates that grounding is more sensitive to pruning than VQA. CAR
retains $76.9\%$, leading the best baseline by $6.0$ percentage points.

\begin{table}[t]
  \centering
  \caption{\textbf{VQA performance comparison on Qwen2.5-VL-7B with $32$ retained tokens.}}
  \label{tab:qwen2.5-vl-32}
  \resizebox{\linewidth}{!}{
    \begin{tabular}{l|cccccccc|cc}
    \toprule
    \textbf{Method} & \textbf{TextVQA} & \textbf{ChartQA} & \textbf{AI2D} & \textbf{OCRBench} & \textbf{HallBench} & \textbf{MME} & \textbf{MMB-EN} & \textbf{MMB-CN} & \textbf{Acc.} & \textbf{Rel.} \\
    \midrule
    \rowcolor{lightgray!75}\multicolumn{11}{c}{\textit{Upper Bound, All 1{,}296 Tokens} (100\%)} \\
    \rowcolor{lightgray!25} Qwen2.5-VL-7B & 84.3 & 87.1 & 81.6 & 867 & 56.2 & 2342.3 & 83.1 & 82.9 & 84.9 & 100.0\% \\
    \rowcolor{lightgray!75}\multicolumn{11}{c}{\textit{Retain 32 Tokens} ($\downarrow$97.5\%)} \\
    
     AnchorPrune{\small\texttt{(ECCV26)}} & 53.5 & 35.4 & 69.8 & \underline{435} & 39.0 & \underline{1879.5} & \underline{73.2} & \underline{72.1} & \underline{60.0} & 70.6\% \\
    HAWK{\small\texttt{(CVPR26)}} & \textbf{66.1} & \textbf{38.8} & \underline{70.8} & 431 & \underline{43.1} & 1791.2 & 64.2 & 63.9 & 59.9 & \underline{70.9\%} \\
    N\"uwa{\small\texttt{(ICLR26)}} & 48.7 & \underline{38.1} & 69.2 & 381 & 38.8 & 1820.6 & 70.4 & 70.0 & 58.0 & 68.3\% \\
   
    \textbf{CAR} (Ours) & \underline{60.6} & 37.5 & \textbf{73.1} & \textbf{523} & \textbf{44.0} & \textbf{2120.6} & \textbf{76.5} & \textbf{74.5} & \textbf{65.6} & \textbf{76.9\%} \\
    \bottomrule
    \end{tabular}
  }
\end{table}

\subsection{Qwen3-VL-8B}
\label{sec:appendix:results-qwen3}

\paragraph{Setup.}
We describe how our method is implemented on Qwen3-VL-8B. The saliency is computed as in
Eq.~\ref{eq:saliency-qwen25} from head $h^{*}$ of the last vision-encoder block, and the features
$x_i$ are the merger outputs. Images are processed at native dynamic resolution, so budgets are
given as fractions of the visual tokens. Qwen3-VL adopts DeepStack, which injects features from
three intermediate vision-encoder blocks into the first three decoder layers. We keep DeepStack
enabled for all methods and prune these intermediate features with the same indices retained by
token pruning. Retained tokens keep their original M-RoPE positions.

\paragraph{VQA and Visual Grounding Results on Qwen3-VL-8B.}
Tables~\ref{tab:qwen3-vl-vqa} and~\ref{tab:qwen3-vl-grounding} compare CAR with five competitive
baselines on Qwen3-VL-8B. On VQA, CAR retains $92.5\%$ of the unpruned performance at a retention
ratio of $22.2\%$ and $81.7\%$ at $11.1\%$, leading the best baseline by $3.6$ and $4.8$ percentage points. On
grounding, it retains $84.9\%$ at a retention ratio of $25\%$ and $54.7\%$ at $9.9\%$, leading the
best baseline by $8.3$ and $20.7$ percentage points. CAR consistently demonstrates superior performance across
different retention ratios for both tasks, leading in the vast majority of benchmarks. Its advantage
becomes more pronounced as pruning becomes more aggressive on both tasks, and is also more prominent
on visual grounding, which is more sensitive to pruning.

\begin{table}[t]
  \centering
  \caption{\textbf{VQA performance comparison on Qwen3-VL-8B.}}
  \label{tab:qwen3-vl-vqa}
  \resizebox{\linewidth}{!}{
    \begin{tabular}{l|cccccccc|cc}
    \toprule
    \textbf{Method} & \textbf{TextVQA} & \textbf{ChartQA} & \textbf{AI2D} & \textbf{OCRBench} & \textbf{HallBench} & \textbf{MME} & \textbf{MMB-EN} & \textbf{MMB-CN} & \textbf{Acc.} & \textbf{Rel.} \\
    \rowcolor{lightgray!75}\multicolumn{11}{c}{\textit{Upper Bound, All Visual Tokens} ($\mathbf{100\%}$)} \\
    \rowcolor{lightgray!25} Qwen3-VL-8B & 83.5 & 83.5 &  84.1 & 874 &  57.9 &  2417.2 & 85.4 &  84.9 &  85.9 & 100.0\% \\
    \rowcolor{lightgray!75}\multicolumn{11}{c}{\textit{Retain} $\mathbf{22.2\%}$ \textit{of Visual Tokens}} \\
    VisionZip{\small\texttt{(CVPR25)}} & 73.9 & \underline{65.9} & 79.0 & 610 & 52.0 & 2162.9 & 80.7 & \underline{79.9} & 75.1 & 87.4\% \\
    DivPrune{\small\texttt{(CVPR25)}} & 72.9 & 52.9 & 77.5 & 630 & 48.4 & 2150.0 & 80.2 & 77.5 & 72.5 & 84.1\% \\
    AnchorPrune{\small\texttt{(ECCV26)}} & \underline{77.0} & 64.1 & \textbf{80.5} & 651 & 51.9 & 2250.9 & 81.4 & 79.8 & \underline{76.5} & \underline{88.9\%} \\
    EADP{\small\texttt{(ECCV26)}} & 75.3 & 58.6 & 78.7 & \underline{675} & 49.5 & 2245.0 & 80.3 & 78.9 & 75.1 & 87.1\% \\
    N\"uwa{\small\texttt{(ICLR26)}} & 76.0 & 60.9 & 79.0 & 628 & \textbf{53.9} & \underline{2260.0} & \underline{81.6} & \underline{79.9} & 75.9 & 88.2\% \\
    \textbf{CAR} (Ours) & \textbf{79.5} & \textbf{70.6} & \underline{80.4} & \textbf{739} & \underline{53.0} & \textbf{2340.0} & \textbf{82.1} & \textbf{81.2} & \textbf{79.7} & \textbf{92.5\%} \\
    \rowcolor{lightgray!75}\multicolumn{11}{c}{\textit{Retain} $\mathbf{11.1\%}$ \textit{of Visual Tokens}} \\
    VisionZip{\small\texttt{(CVPR25)}} & 55.6 & 43.0 & 71.5 & 347 & 41.9 & 1806.6 & 71.1 & 71.1 & 59.9 & 69.6\% \\
    DivPrune{\small\texttt{(CVPR25)}} & 61.6 & 34.2 & 70.6 & 477 & 43.4 & 1897.3 & 72.1 & 71.7 & 62.0 & 72.0\% \\
    AnchorPrune{\small\texttt{(ECCV26)}} & 61.8 & 40.5 & \underline{73.4} & 405 & 44.3 & 1884.8 & 72.0 & 71.1 & 62.2 & 72.3\% \\
    EADP{\small\texttt{(ECCV26)}} & \underline{67.8} & 40.0 & \underline{73.4} & \underline{539} & 42.7 & \underline{2074.2} & \underline{75.8} & \underline{75.3} & \underline{66.6} & \underline{76.9\%} \\
    N\"uwa{\small\texttt{(ICLR26)}} & 64.4 & \underline{45.8} & 72.9 & 403 & \underline{44.9} & 2004.1 & 74.5 & 74.7 & 64.7 & 75.1\% \\
    \textbf{CAR} (Ours)  & \textbf{72.2} & \textbf{46.8} & \textbf{75.7} & \textbf{564} & \textbf{48.1} & \textbf{2183.7} & \textbf{78.2} & \textbf{77.9} & \textbf{70.6} & \textbf{81.7\%} \\
    \bottomrule
    \end{tabular}
  }
\end{table}

\begin{table}[t]
  \centering
  \caption{\textbf{Visual grounding performance comparison on Qwen3-VL-8B.}}
  \label{tab:qwen3-vl-grounding}
  \resizebox{\linewidth}{!}{
    \begin{tabular}{l|ccc|ccc|cc|cc}
    \toprule
    \multirow{2}{*}{\textbf{Method}}
    & \multicolumn{3}{c|}{\textbf{RefCOCO}} & \multicolumn{3}{c|}{\textbf{RefCOCO+}}
    & \multicolumn{2}{c|}{\textbf{RefCOCOg}} & \multirow{2}{*}{\textbf{Avg.}} & \multirow{2}{*}{\textbf{Rel.}} \\
    & testA & testB & val & testA & testB & val & test & val & & \\
    \rowcolor{lightgray!75}\multicolumn{11}{c}{\textit{Upper Bound, All Visual Tokens} ($\mathbf{100\%}$)} \\
    \rowcolor{lightgray!25} Qwen3-VL-8B & 90.4 & 84.7 & 88.5 & 85.8 & 75.5 & 81.4 & 87.1 & 86.4 & 85.0 & 100.0\% \\
    \rowcolor{lightgray!75}\multicolumn{11}{c}{\textit{Moderate Budget, Retain} $\mathbf{25\%}$ \textit{of Visual Tokens}} \\
    VisionZip{\small\texttt{(CVPR25)}} & 70.4 & 65.4 & 67.8 & 61.6 & 55.8 & 59.5 & 66.1 & 65.6 & 64.0 & 75.3\% \\
    DivPrune{\small\texttt{(CVPR25)}} & 66.3 & 56.1 & 62.4 & 60.6 & 47.5 & 55.1 & 61.7 & 59.6 & 58.7 & 68.9\% \\
    AnchorPrune{\small\texttt{(ECCV26)}} & \underline{72.6} & 61.9 & \underline{69.0} & \underline{66.0} & 53.9 & \underline{61.4} & \underline{68.4} & \underline{68.4} & \underline{65.2} & \underline{76.6\%} \\
    EADP{\small\texttt{(ECCV26)}} & 66.7 & 63.3 & 65.7 & 58.3 & 54.9 & 57.5 & 65.0 & 64.4 & 62.0 & 72.9\% \\
    N\"uwa{\small\texttt{(ICLR26)}} & 67.7 & \underline{68.3} & 67.7 & 60.3 & \underline{58.1} & 58.6 & 67.3 & 67.6 & 64.5 & 75.8\% \\
    \textbf{CAR} (Ours) & \textbf{78.3} & \textbf{73.8} & \textbf{76.8} & \textbf{69.6} & \textbf{63.1} & \textbf{66.9} & \textbf{74.5} & \textbf{74.7} & \textbf{72.2} & \textbf{84.9\%} \\
    \rowcolor{lightgray!75}\multicolumn{11}{c}{\textit{Aggressive Budget, Retain} $\mathbf{9.9\%}$ \textit{of Visual Tokens}} \\
    VisionZip{\small\texttt{(CVPR25)}} & 30.4 & 26.4 & 28.2 & 24.2 & 21.8 & 23.9 & 24.6 & 24.4 & 25.5 & 29.9\% \\
    
    DivPrune{\small\texttt{(CVPR25)}} & 31.9 & 25.2 & 27.9 & 26.3 & 19.0 & 22.8 & 24.7 & 25.4 & 25.4 & 29.8\% \\
    
    AnchorPrune{\small\texttt{(ECCV26)}} & 31.3 & 25.3 & 28.3 & 25.1 & 21.4 & 23.7 & 26.4 & 26.5 & 26.0 & 30.5\% \\
   
    EADP{\small\texttt{(ECCV26)}} & 26.5 & 23.8 & 24.7 & 20.9 & 19.0 & 20.6 & 22.3 & 21.6 & 22.4 & 26.3\% \\

    N\"uwa{\small\texttt{(ICLR26)}} & \underline{32.3} & \underline{32.5} & \underline{30.4} & \underline{26.8} & \underline{24.4} & \underline{25.2} & \underline{30.2} & \underline{29.4} & \underline{28.9} & \underline{34.0\%} \\
    
    \textbf{CAR} (Ours) & \textbf{53.7} & \textbf{48.7} & \textbf{51.4} & \textbf{44.3} & \textbf{38.9} & \textbf{41.9} & \textbf{47.0} & \textbf{46.8} & \textbf{46.6} & \textbf{54.7\%} \\
    \bottomrule
    \end{tabular}
  }
  \vspace{-4mm}
\end{table}

\subsection{Video Understanding on Qwen3-VL-8B}
\label{sec:appendix:results-video}

\paragraph{Setup.}
We describe how our method is implemented for video understanding on Qwen3-VL-8B. Each video is
sampled to $32$ frames, and the visual tokens of all frames are selected jointly. The saliency is
computed as in Eq.~\ref{eq:saliency-qwen25} from the last vision-encoder block, with the attention
averaged over all heads (Appendix~\ref{sec:appendix:probe}), and the features $x_i$ are the merger
outputs. We use the capped bias of Eq.~\ref{eq:bias-capped} without the spatial kernel. We keep DeepStack enabled
for all methods and prune its intermediate features with the same indices retained by token
pruning. Retained tokens keep their original M-RoPE positions.

\paragraph{Video Understanding Results on Qwen3-VL-8B.}
Table~\ref{tab:qwen3-vl-video} compares CAR with six competitive baselines, including the
video-specific GSTEP~\citep{zhang2026gstep}. CAR retains $97.9\%$ of the unpruned performance at a
retention ratio of $25\%$ and $94.7\%$ at $9.9\%$, leading the best baseline by $0.9$ and $0.5$
percentage points. It performs best on Video-MME and LongVideoBench at both retention ratios. Videos in these
benchmarks are longer and repeat the same content across many frames, and the coverage objective
removes this redundancy and spreads the retained tokens over the distinct content of the video. On
MVBench, CAR trails the best baseline by about $1$ point, mainly on the five subtasks built on
CLEVRER~\citep{yi2020clevrer}, which ask about the motion of small objects: such an object looks
nearly identical across frames, so CAR keeps few of its copies and loses the positions that reveal
its motion.

\begin{table}[t]
  \centering
  \caption{\textbf{Video understanding performance comparison on Qwen3-VL-8B.}}
  \label{tab:qwen3-vl-video}
  \resizebox{\linewidth}{!}{
    \begin{tabular}{l|cccc|c|c|cc}
    \toprule
    \multirow{2}{*}{\textbf{Method}} & \multicolumn{4}{c|}{\textbf{Video-MME}}
    & \multirow{2}{*}{\shortstack{\textbf{Long}\\\textbf{VideoBench}}}
    & \multirow{2}{*}{\textbf{MVBench}} & \multirow{2}{*}{\textbf{Avg.}} & \multirow{2}{*}{\textbf{Rel.}} \\
    & Short & Medium & Long & Overall & & & & \\
    \rowcolor{lightgray!75}\multicolumn{9}{c}{\textit{Upper Bound, All Visual Tokens} ($\mathbf{100\%}$)} \\
    \rowcolor{lightgray!25} Qwen3-VL-8B & 78.1 & 63.4 & 55.7 & 65.7 & 60.2 & 69.5 & 65.5 & 100.0\% \\
    \rowcolor{lightgray!75}\multicolumn{9}{c}{\textit{Retain} $\mathbf{25\%}$ \textit{of Visual Tokens} } \\
    VisionZip{\small\texttt{(CVPR25)}} & 74.4 & 60.9 & 54.2 & 63.2 & 57.6 & \textbf{68.9} & 63.2 & 96.6\% \\
    DivPrune{\small\texttt{(CVPR25)}} & 72.9 & 59.1 & 53.8 & 61.9 & 57.1 & 65.6 & 61.7 & 94.4\% \\
    AnchorPrune{\small\texttt{(ECCV26)}} & 75.3 & 61.1 & \underline{54.3} & 63.6 & 57.8 & 68.1 & 63.4 & 96.8\% \\
    EADP{\small\texttt{(ECCV26)}} & 74.1 & \underline{61.9} & \underline{54.3} & 63.4 & 57.5 & 68.7 & 63.3 & 96.8\% \\
    GSTEP{\small\texttt{(MM26)}} & 74.9 & 58.2 & 53.7 & 62.3 & \underline{58.6} & \underline{68.8} & 62.7 & 95.8\% \\
    N\"uwa{\small\texttt{(ICLR26)}} & \underline{75.9} & 61.0 & 54.1 & \underline{63.7} & 57.8 & 68.5 & \underline{63.5} & \underline{97.0\%} \\

     \textbf{CAR} (Ours) & 
    \textbf{76.1} & \textbf{62.4} & \textbf{54.8} & \textbf{64.4} & \textbf{58.9} & 67.7 & \textbf{64.0} & \textbf{97.9\%} \\

    \rowcolor{lightgray!75}\multicolumn{9}{c}{\textit{Retain} $\mathbf{9.9\%}$ \textit{of Visual Tokens}} \\
    VisionZip{\small\texttt{(CVPR25)}} & 69.9 & 57.2 & 53.1 & 60.1 & 55.0 & 61.5 & 59.5 & 91.1\% \\
    DivPrune{\small\texttt{(CVPR25)}} & 68.7 & 56.7 & 50.3 & 58.6 & 54.4 & 60.1 & 58.1 & 88.9\% \\
    AnchorPrune{\small\texttt{(ECCV26)}} & 71.4 & 57.7 & 53.1 & 60.7 & 55.3 & 64.1 & 60.4 & 92.4\% \\
    EADP{\small\texttt{(ECCV26)}} & 70.4 & 57.6 & \underline{53.8} & 60.6 & 53.9 & 62.8 & 59.8 & 91.6\% \\
    GSTEP{\small\texttt{(MM26)}} & 71.9 & 56.8 & 52.6 & 60.4 & \underline{57.4} & \textbf{66.1} & 60.9 & 93.0\% \\
    N\"uwa{\small\texttt{(ICLR26)}} & \underline{72.9} & \textbf{59.6} & 53.6 & \underline{62.0} & 55.7 & \underline{66.0} & \underline{61.6} & \underline{94.2\%} \\
    
     \textbf{CAR} (Ours) & \textbf{73.7} & \underline{59.1} & \textbf{53.9} & \textbf{62.2} & \textbf{57.8} & 65.0 & \textbf{62.0} & \textbf{94.7\%} \\
    
    \bottomrule
    \end{tabular}
  }
  \vspace{-4mm}
\end{table}

\subsection{InternVL3.5-8B}
\label{sec:appendix:results-internvl}

\paragraph{Setup.}
We describe how our method is implemented on InternVL3.5-8B. Following the InternVL evaluation
setting in VLMEvalKit, each image is split into at most $12$ tiles of $448{\times}448$ pixels for
ChartQA, $24$ for OCRBench and $6$ for the other benchmarks, plus a thumbnail of the whole image
when more than one tile is used. We find that retaining thumbnail tokens is important, since the
thumbnail is the only view of the whole image. The thumbnail has $256$ tokens. For all methods,
we therefore reserve $128$ of the $k$ retained tokens for the thumbnail, or all $k$ when $k<128$,
and assign the rest to the other tiles. The tokens of the thumbnail and of the other tiles are
selected separately. The saliency is
computed per tile as in Eq.~\ref{eq:saliency-qwen25}, from the vision-encoder layer and head
selected in Appendix~\ref{sec:appendix:probe}, and the features $x_i$ are the projector outputs. The
spatial kernel on InternVL3.5-8B is
\begin{equation}
s_{ij}=\big[\cos(x_i,x_j)\big]_+\cdot\exp\!\Big(\!-\frac{\lVert g_i-g_j\rVert^{2}}{2\gamma}\Big),
\label{eq:internvl-coords}
\end{equation}
with the grid position $g_i$ and the scale $\gamma$ given by
\begin{equation*}
\begin{aligned}
&\text{tiles:} && g_i=(16r+u,\;16c+v), && \gamma=256\,n_{\mathrm{tile}}/k,\\
&\text{thumbnail:} && g_i=(u,\,v), && \gamma=256/k_{\mathrm{th}},
\end{aligned}
\end{equation*}
where $(u,v)$ is the row and column of the token in the $16{\times}16$ token grid of its tile or of
the thumbnail, $(r,c)$ is the row and column of the tile in the tile layout, $n_{\mathrm{tile}}$ is
the number of tiles, and $k_{\mathrm{th}}$ is the number of tokens reserved for the thumbnail. We disable the bias on this backbone, where it does not improve accuracy (Table~\ref{tab:bias-backbones}). Retained tokens keep their original positions.

\paragraph{VQA and Visual Grounding Results on InternVL3.5-8B.}
Tables~\ref{tab:internvl35-vqa} and~\ref{tab:internvl35-grounding} compare CAR with five competitive
baselines on InternVL3.5-8B. On VQA, CAR retains $94.9\%$ of the unpruned performance at a retention
ratio of $22.2\%$ and $89.3\%$ at $11.1\%$, leading the best baseline by $6.6$ and $5.9$ percentage points. On
grounding, it retains $93.8\%$ at a retention ratio of $25\%$ and $85.2\%$ at $9.9\%$, leading the
best baseline by $2.1$ and $1.4$ percentage points. This is attributed to the saliency-weighted coverage: since the late layers of the InternVL3.5
vision encoder suffer from attention sinks (Appendix~\ref{sec:appendix:internvl-attn}), the probe
selects a suitable intermediate layer to extract the saliency, which helps retain important tokens.

\begin{table}[t]
\vspace{2mm}
  \centering
  \caption{\textbf{VQA performance comparison on InternVL3.5-8B.}}
  \label{tab:internvl35-vqa}
  \resizebox{\linewidth}{!}{
    \begin{tabular}{l|cccccccc|cc}
    \toprule
    \textbf{Method} & \textbf{TextVQA} & \textbf{ChartQA} & \textbf{AI2D} & \textbf{OCRBench} & \textbf{HallBench} & \textbf{MME} & \textbf{MMB-EN} & \textbf{MMB-CN} & \textbf{Acc.} & \textbf{Rel.} \\
    \rowcolor{lightgray!75}\multicolumn{11}{c}{\textit{Upper Bound, All Visual Tokens} ($\mathbf{100\%}$)} \\
    \rowcolor{lightgray!25} InternVL3.5-8B & 76.5 & 86.3 & 83.4 & 839 & 52.7 & 2333.1 & 83.0 & 81.3 & 83.0 & 100.0\% \\
    \rowcolor{lightgray!75}\multicolumn{11}{c}{\textit{Retain} $\mathbf{22.2\%}$ \textit{of Visual Tokens}} \\
    VisionZip{\small\texttt{(CVPR25)}} & 56.6 & 54.5 & 78.5 & 369 & 42.0 & 2143.8 & 77.2 & 75.7 & 66.1 & 79.1\% \\
    DivPrune{\small\texttt{(CVPR25)}} & 67.7 & \underline{68.6} & 78.8 & 591 & 44.4 & \underline{2244.2} & \underline{79.1} & 76.8 & 73.3 & 87.9\% \\
    AnchorPrune{\small\texttt{(ECCV26)}} & 64.0 & 64.4 & \underline{79.0} & 508 & \underline{44.6} & 2159.4 & \textbf{79.7} & \underline{77.7} & 71.0 & 85.3\% \\
    EADP{\small\texttt{(ECCV26)}} & \underline{71.0} & 65.4 & 77.5 & \underline{653} & 43.9 & 2220.0 & 77.6 & 77.1 & \underline{73.6} & \underline{88.3\%} \\
    
    N\"uwa{\small\texttt{(ICLR26)}} & 56.7 & 45.8 & 76.1 & 296 & 42.9 & 2055.0 & 77.0 & 76.2 & 63.4 & 76.2\% \\
    \textbf{CAR} (Ours) & \textbf{72.7} & \textbf{80.9} & \textbf{81.3} & \textbf{759} & \textbf{47.7} & \textbf{2336.7} & \textbf{79.7} & \textbf{77.9} & \textbf{79.1} & \textbf{94.9\%} \\

    \rowcolor{lightgray!75}\multicolumn{11}{c}{\textit{Retain} $\mathbf{11.1\%}$ \textit{of Visual Tokens}} \\
    VisionZip{\small\texttt{(CVPR25)}} & 52.2 & 52.8 & 76.4 & 343 & 38.3 & 2066.2 & 73.2 & 71.9 & 62.8 & 75.0\% \\
    DivPrune{\small\texttt{(CVPR25)}} & 61.4 & \underline{60.9} & 76.3 & 505 & 42.8 & \underline{2186.2} & \underline{76.5} & \underline{74.8} & 69.1 & 82.7\% \\
    AnchorPrune{\small\texttt{(ECCV26)}} & 57.0 & 58.0 & \underline{77.2} & 435 & 42.2 & 2096.3 & 75.7 & 73.5 & 66.5 & 79.7\% \\
    EADP{\small\texttt{(ECCV26)}} & \underline{64.0} & 56.3 & 75.2 & \underline{576} & \underline{44.2} & 2173.8 & 75.7 & 74.4 & \underline{69.5} & \underline{83.4\%} \\
    N\"uwa{\small\texttt{(ICLR26)}} & 48.1 & 37.6 & 73.6 & 252 & 39.4 & 1991.3 & 71.9 & 70.2 & 58.2 & 69.7\% \\
    \textbf{CAR} (Ours) & \textbf{66.1} & \textbf{68.7} & \textbf{79.5} & \textbf{666} & \textbf{47.5} & \textbf{2280.4} & \textbf{77.7} & \textbf{75.0} & \textbf{74.4} & \textbf{89.3\%} \\

    \bottomrule
    \end{tabular}
  }
  \vspace{-4mm}
\end{table}

\begin{table}[t]
  \centering
  \caption{\textbf{Visual grounding performance comparison on InternVL3.5-8B.}}
  \label{tab:internvl35-grounding}
  \resizebox{\linewidth}{!}{
    \begin{tabular}{l|ccc|ccc|cc|cc}
    \toprule
    \multirow{2}{*}{\textbf{Method}}
    & \multicolumn{3}{c|}{\textbf{RefCOCO}} & \multicolumn{3}{c|}{\textbf{RefCOCO+}}
    & \multicolumn{2}{c|}{\textbf{RefCOCOg}} & \multirow{2}{*}{\textbf{Avg.}} & \multirow{2}{*}{\textbf{Rel.}} \\
    & testA & testB & val & testA & testB & val & test & val & & \\
    \rowcolor{lightgray!75}\multicolumn{11}{c}{\textit{Upper Bound, All Visual Tokens} ($\mathbf{100\%}$)} \\
    \rowcolor{lightgray!25} InternVL3.5-8B & 94.3 & 88.9 & 92.0 & 92.2 & 82.5 & 87.7 & 89.2 & 88.6 & 89.4 & 100.0\% \\

    \rowcolor{lightgray!75}\multicolumn{11}{c}{\textit{Moderate Budget, Retain} $\mathbf{25\%}$ \textit{of Visual Tokens}} \\
    VisionZip{\small\texttt{(CVPR25)}} & 83.6 & 79.0 & 81.1 & 79.5 & 70.5 & 75.3 & 79.5 & 79.6 & 78.5 & 87.8\% \\
    DivPrune{\small\texttt{(CVPR25)}} & 87.4 & 78.0 & 81.8 & 84.8 & 71.1 & 76.9 & 80.5 & 80.2 & 80.1 & 89.5\% \\
    AnchorPrune{\small\texttt{(ECCV26)}} & 85.3 & 78.3 & 80.6 & 81.7 & 70.8 & 75.3 & 80.1 & 79.2 & 78.9 & 88.2\% \\
    EADP{\small\texttt{(ECCV26)}} & \underline{88.8} & \underline{80.5} & \underline{84.6} & \underline{85.9} & \underline{73.5} & \underline{79.4} & \underline{81.7} & \underline{81.7} & \underline{82.0} & \underline{91.7\%} \\
    N\"uwa{\small\texttt{(ICLR26)}} & 78.1 & 72.6 & 74.7 & 73.4 & 65.6 & 69.3 & 74.7 & 75.0 & 72.9 & 81.5\% \\
    
     \textbf{CAR} (Ours) & \textbf{89.9} & \textbf{82.5} & \textbf{86.5} & \textbf{86.8} & \textbf{76.2} & \textbf{81.3} & \textbf{84.3} & \textbf{83.5} & \textbf{83.9} & \textbf{93.8\%} \\
    \rowcolor{lightgray!75}\multicolumn{11}{c}{\textit{Aggressive Budget, Retain} $\mathbf{9.9\%}$ \textit{of Visual Tokens}} \\
    VisionZip{\small\texttt{(CVPR25)}} & 76.9 & 71.2 & 72.3 & 71.3 & 63.7 & 66.3 & 72.4 & 72.6 & 70.8 & 79.2\% \\
    
    DivPrune{\small\texttt{(CVPR25)}} & 80.9 & 70.9 & 76.1 & 77.7 & 63.7 & 70.1 & 73.2 & 73.4 & 73.3 & 81.8\% \\
    
    AnchorPrune{\small\texttt{(ECCV26)}} & 77.9 & 71.0 & 73.8 & 74.3 & 63.8 & 67.8 & 73.0 & 73.3 & 71.9 & 80.3\% \\
    
    EADP{\small\texttt{(ECCV26)}} & \underline{82.0} & \underline{73.8} & \underline{77.2} & \underline{78.9} & \underline{67.0} & \underline{71.8} & \underline{74.3} & \underline{75.0} & \underline{75.0} & \underline{83.8\%} \\

    N\"uwa{\small\texttt{(ICLR26)}} & 67.8 & 64.3 & 64.4 & 61.6 & 56.2 & 57.6 & 64.0 & 65.7 & 62.7 & 70.1\% \\

     \textbf{CAR} (Ours) & \textbf{82.7} & \textbf{75.5} & \textbf{78.7} & \textbf{79.3} & \textbf{67.8} & \textbf{73.7} & \textbf{76.0} & \textbf{76.1} & \textbf{76.2} & \textbf{85.2\%} \\
    
    \bottomrule
    \end{tabular}
  }
  \vspace{-4mm}
\end{table}

\paragraph{Why the Bias Is Disabled on InternVL3.5-8B.}
\label{sec:appendix:internvl-bias}
InternVL3.5-8B splits an image into tiles and appends a thumbnail of the whole image, and it relies on
the thumbnail much more than on the other tiles: in the unpruned model the thumbnail holds $16\%$ of
the visual tokens but receives $36\%$ of the attention the text pays to the image. Prior work
likewise treats the thumbnail as the global view of high-resolution LVLMs and uses it to guide the
compression of the tiles~\citep{liu2026globalcom,sun2026ivc,liang2025ptp}. Following
IVC-Prune~\citep{sun2026ivc}, we keep half of the thumbnail tokens and assign the rest of the budget
to the other tiles.

Table~\ref{tab:internvl-pool-share} gives the visual attention share under this protocol before and
after pruning, separately for the thumbnail and the other tiles. Pruning lowers the visual share
from $25.8\%$ to $15.8\%$, and the drop comes entirely from the other tiles: their share falls to
$0.25$ of its unpruned value, while that of the thumbnail does not drop and even rises to $1.26$, as
it now competes with far fewer tile tokens. The global view the model relies on most is thus not
weakened, and the missing attention lies on the high-resolution tiles, whose content the thumbnail
already represents at a coarser scale. {A single bias $\log(N/k)$ for all visual tokens
ignores this imbalance: the thumbnail loses far fewer tokens but receives the same bias as the tiles
and is over-compensated: its share rises to $2.44$ times its unpruned value, whereas that of the tiles
reaches only $0.59$ (Table~\ref{tab:internvl-pool-share}). This lowers the accuracy from $76.2$ to $66.4$ on grounding ($9.9\%$
retention) and from $74.4$ to $71.5$ on VQA ($11.1\%$ retention). We therefore compute the bias
separately for the thumbnail and the other tiles, replacing $N/k$ with the ratio of total to retained
tokens within each pool, so that almost all of the compensation goes to the tiles. This restores their
share to $0.81$ but does not improve accuracy ($75.8$ on grounding and $73.4$ on VQA; Table~\ref{tab:bias-backbones}), so the restored
tile attention adds little information.} We therefore disable the bias on this backbone.

\begin{table}[h]
  \centering
  \small
  \caption{\textbf{Visual attention share on InternVL3.5-8B} ($11.1\%$ of the visual tokens
  retained, half of the thumbnail kept). Shares (\%) are fractions of the attention the text pays to all
  tokens, averaged over layers and $32$ images; parentheses give the ratio to the unpruned value. {The single bias adds $\log(N/k)$ to all retained visual tokens;} the per-pool bias is computed separately for the thumbnail and the other tiles. }
  \label{tab:internvl-pool-share}
  \begin{tabular}{l|ccc}
  \toprule
  & \textbf{Thumbnail} & \textbf{Other tiles} & \textbf{All visual} \\
  \midrule
  Unpruned                   & $9.2$                & $16.6$               & $25.8$ \\
  Pruned, no bias            & $11.6$ ($1.26\times$)      & $4.1$ ($0.25\times$)       & $15.8$ ($0.61\times$) \\
  {Pruned, single bias} & {$22.4$ ($2.44\times$)} & {$9.8$ ($0.59\times$)} & {$32.3$ ($1.25\times$)} \\
  Pruned, per-pool bias      & $11.1$ ($1.20\times$)      & $13.4$ ($0.81\times$)      & $24.5$ ($0.95\times$) \\
  \bottomrule
  \end{tabular}
\end{table}

\subsection{LLaVA-1.5-7B}
\label{sec:appendix:results-llava}

\paragraph{Setup.}
We describe how our method is implemented on LLaVA-1.5-7B. Each image is padded to a square and
resized to $336{\times}336$, giving $576$ visual tokens on a $24{\times}24$ grid. The saliency is
computed as in Eq.~\ref{eq:saliency-qwen25} from the layer and head selected by the probe
(Appendix~\ref{sec:appendix:probe}). The features $x_i$ are the projector outputs, and $g_i$ in Eq.~\ref{eq:fl-kernel} is the grid
position of token $i$. Following N\"uwa~\citep{huang2026nuwa}, we evaluate grounding with lmms-eval
on seven splits of RefCOCO, RefCOCO+ and RefCOCOg, retaining $64$ visual tokens on average. Since
grounding depends on where the retained tokens lie, retained tokens keep their original positions
for all methods.

\paragraph{Visual Grounding Results on LLaVA-1.5-7B.}
Table~\ref{tab:llava15-grounding} compares CAR with five competitive baselines on LLaVA-1.5-7B.
Retaining $64$ visual tokens on average, CAR retains $59.0\%$ of the unpruned performance,
performing best on every split and leading the best baseline, N\"uwa, by $4.4$ percentage points.
N\"uwa and CAR both preserve the spatial structure that grounding relies on, N\"uwa by partitioning
the token grid and aggregating the discarded tokens into their spatial neighbours, and CAR by
weighted coverage selection with a spatial kernel. Two factors contribute
to the further lead of CAR. The late layers of the CLIP encoder exhibit attention sinks, so the
probe selects the intermediate layer $9$ as the saliency source, which helps retain important
tokens. The bias restores the attention share lost to pruning and further improves accuracy on
average (Table~\ref{tab:bias-backbones}).

\begin{table}[t]
  \centering
  \caption{\textbf{Visual grounding performance comparison on LLaVA-1.5-7B.}}
  \label{tab:llava15-grounding}
  \resizebox{\linewidth}{!}{
    \begin{tabular}{l|cc|ccc|cc|cc}
    \toprule
    \multirow{2}{*}{\textbf{Method}}
    & \multicolumn{2}{c|}{\textbf{RefCOCO}} & \multicolumn{3}{c|}{\textbf{RefCOCO+}}
    & \multicolumn{2}{c|}{\textbf{RefCOCOg}} & \multirow{2}{*}{\textbf{Avg.}} & \multirow{2}{*}{\textbf{Rel.}} \\
    & test & val & testA & testB & val & test & val & & \\
    \rowcolor{lightgray!75}\multicolumn{10}{c}{\textit{Upper Bound, All Visual Tokens} ($\mathbf{576}$)} \\
    \rowcolor{lightgray!25} LLaVA-1.5-7B & 57.8 & 56.2 & 59.3 & 38.9 & 49.9 & 48.4 & 48.7 & 51.3 & 100.0\% \\

    \rowcolor{lightgray!75}\multicolumn{10}{c}{\textit{Average Tokens} $\mathbf{64}$ ($\mathbf{11.1\%}$ \textit{of Visual Tokens})} \\
    VisionZip{\small\texttt{(CVPR25)}} & 9.4  & 9.4  & 7.4  & 8.1  & 7.3  & 7.2   & 7.2 & 8.0 & 15.8\% \\
    DivPrune{\small\texttt{(CVPR25)}} & 20.3 & 19.6 & 19.2 & 13.8 & 16.7 & 15.6  & 16.1 & 17.3 & 33.8\% \\
    AnchorPrune{\small\texttt{(ECCV26)}} & 17.3  & 16.1 & 15.5 & 12.6 & 14.3  & 15.2  & 15.0 & 15.1 & 29.7\% \\
    EADP{\small\texttt{(ECCV26)}} & 23.1  & 22.2 & 22.6  & 15.0 & 19.3  & 19.2  & 18.8 & 20.0 & 39.0\% \\
    N\"uwa{\small\texttt{(ICLR26)}} & \underline{33.2} & \underline{32.7} & \underline{34.0} & \underline{20.6} & \underline{26.5} & \underline{24.5} & \underline{25.6} & \underline{28.2} & \underline{54.6\%} \\

    \textbf{CAR} (Ours) & \textbf{36.4} & \textbf{35.5} & \textbf{35.2} & \textbf{22.3} & \textbf{29.0} & \textbf{27.2} & \textbf{27.3} & \textbf{30.4} & \textbf{59.0\%} \\

    \bottomrule
    \end{tabular}
  }
  \vspace{-4mm}
\end{table}

\subsection{Full Results of the Component Ablation}
\label{sec:appendix:results-components}

Tables~\ref{tab:ablation-grounding} and~\ref{tab:ablation-vqa} give the per-split and per-benchmark
results of the component ablation on Qwen2.5-VL-7B, of which Table~\ref{tab:abl-comp} reports the
averages. To separate the contribution of the selection from that of the attention reallocation, the
tables first list two references without any selection signal: random selection, and random
selection with our bias alone. The remaining rows add the components cumulatively: (i) the
unweighted facility-location objective; (ii) the logit bias; (iii) the saliency $w$ weighting each
token in that objective, with $w$ averaged over the heads; (iv) the probed head $h^{*}$ in place of
the head average; and (v) the spatial kernel on the similarity, which completes the method.

The two tasks depend on different components. On grounding the attention reallocation dominates:
adding the bias raises the average accuracy from $18.8$ to $38.6$ under random selection and from
$20.1$ to $50.0$ under coverage selection, whereas without the bias, replacing random selection by
coverage selection is worth only $1.3$ points. Restoring the visual attention share is therefore a
precondition at this budget, and the selection pays off only once the share is restored. On top of
it, the saliency weighting and the probed head add $4.8$ and $7.7$ points and the spatial kernel a
further $4.3$, consistently over the eight splits. On VQA the components rank the other way round:
the bias is worth $0.3$ and $1.5$ points under the two selections, while coverage selection over
random selection is worth $8.0$ points, and the saliency weighting and the probed head add $4.8$ and
$0.8$. The spatial kernel costs $0.2$ points there, so we enable it on grounding only.

\begin{table}[t]
\vspace{2mm}
    \centering
    \caption{\textbf{Component ablation on visual grounding with Qwen2.5-VL-7B}
    ($9.9\%$ of the visual tokens retained). $\Delta$ is the gain over the row above.}
    \label{tab:ablation-grounding}
    \resizebox{\linewidth}{!}{
      \begin{tabular}{ccccc|ccc|ccc|cc|cc}
      \toprule
      \multicolumn{5}{c|}{\textbf{Component}}
      & \multicolumn{3}{c|}{\textbf{RefCOCO}} & \multicolumn{3}{c|}{\textbf{RefCOCO+}}
      & \multicolumn{2}{c|}{\textbf{RefCOCOg}}
      & \multirow{2}{*}{\textbf{Avg.}} & \multirow{2}{*}{\textbf{$\Delta$}} \\
      \cmidrule(lr){1-5}\cmidrule(lr){6-8}\cmidrule(lr){9-11}\cmidrule(lr){12-13}
      Facility loc. & Bias & Saliency $w$ & Head sel. & Spatial
      & testA & testB & val & testA & testB & val & test & val & & \\
      \midrule
      \rowcolor{lightgray!75}\multicolumn{15}{c}{\textit{Upper Bound, All Visual Tokens} ($\mathbf{100\%}$)} \\
      \rowcolor{lightgray!25}
      \multicolumn{5}{c|}{--}
      & 92.2 & 84.8 & 89.4 & 87.6 & 74.1 & 82.8 & 86.7 & 86.7 & 85.5 & -- \\
      \midrule
      \rowcolor{lightgray!75}\multicolumn{15}{c}{\textit{No Selection Signal (Random Selection)}} \\
      $\times$ & $\times$ & -- & -- & -- & 23.1 & 17.2 & 21.5 & 20.3 & 14.5 & 18.6 & 17.3 & 18.1  & 18.8 & -- \\
      $\times$ & \checkmark & -- & -- & -- & 44.0
      & 37.6 & 42.9 & 38.3 & 31.7 & 37.0 & 39.0 & 38.4 & 38.6 & +19.8 \\
      \midrule
      \checkmark & $\times$ & -- & -- & $\times$
      & 25.3 & 17.6 & 21.5 & 22.9 & 16.0 & 19.2 & 18.8 & 19.4 & 20.1 & -- \\
      \checkmark & \checkmark & -- & -- & $\times$
      & 57.9 & 46.7 & 55.0 & 53.2 & 38.6 & 48.7 & 49.8 & 50.1 & 50.0 & +29.9 \\
     \checkmark & \checkmark & \checkmark & $\times$ & $\times$
      & 62.9 & 53.4 & 58.6 & 56.5 & 44.6 & 50.8 & 55.6 & 56.3 & 54.8 & +4.8 \\
      \checkmark & \checkmark & \checkmark & \checkmark & $\times$
      & 71.8 & 60.6 & 67.7 & 64.2 & 50.8 & 59.5 & 63.6 & 61.7 & 62.5 & $+7.7$ \\
      \rowcolor{lightgray!25}
      \checkmark & \checkmark & \checkmark & \checkmark & \checkmark
      & \textbf{75.4} & \textbf{66.7} & \textbf{72.2} & \textbf{67.2} & \textbf{55.9} & \textbf{62.3}
      & \textbf{67.8} & \textbf{66.5} & \textbf{66.8} & $+4.3$ \\
      \bottomrule
      \end{tabular}
    }
  \end{table}
  
\begin{table}[t]
  \centering
  \caption{\textbf{Component ablation on VQA with Qwen2.5-VL-7B}
  ($128$ of $1296$ visual tokens retained). The shaded row is the deployed configuration.
  $\Delta$ is the gain over the row above.}
  \label{tab:ablation-vqa}
  \resizebox{\linewidth}{!}{
  \begin{tabular}{ccccc|cccccccc|cc}
  \toprule
  \multicolumn{5}{c|}{\textbf{Component}}
  & \multirow{2}{*}{\textsc{TextVQA}} & \multirow{2}{*}{\textsc{ChartQA}}
  & \multirow{2}{*}{\textsc{AI2D}} & \multirow{2}{*}{\textsc{OCRBench}}
  & \multirow{2}{*}{\textsc{HallB}} & \multirow{2}{*}{\textsc{MME}}
  & \multirow{2}{*}{\textsc{MMB}\textsubscript{EN}}
  & \multirow{2}{*}{\textsc{MMB}\textsubscript{CN}}
  & \multirow{2}{*}{\textbf{Acc.}} & \multirow{2}{*}{\textbf{$\Delta$}} \\
  \cmidrule(lr){1-5}
  Facility loc. & Bias & Saliency $w$ & Head sel. & Spatial & & & & & & & & & & \\
  \midrule
  \rowcolor{lightgray!75}\multicolumn{15}{c}{\textit{Upper Bound, All Visual Tokens} ($\mathbf{100\%}$)} \\
  \rowcolor{lightgray!25}
  \multicolumn{5}{c|}{--} & 84.3 & 87.1 & 81.6 & 867 & 56.2 & 2342.3 & 83.1 & 82.9 & 84.9 & -- \\
   \midrule
  \rowcolor{lightgray!75}\multicolumn{15}{c}{\textit{No Selection Signal (Random Selection)}} \\
  $\times$ & $\times$ & -- & -- & -- & 56.4 & 38.0 & 72.5 & 497 & 41.2 & 2066.8 & 76.2 & 76.4 & 64.2 & -- \\
  $\times$ & \checkmark & -- & -- & -- & 55.0 &  37.0 & 73.9 & 489 & 42.5 & 2100.4 & 75.8 & 77.6 & 64.5 & +0.3 \\ %
  
  \midrule
  \checkmark & $\times$ & -- & -- & $\times$
    & 69.7 & 54.5 &  76.7 & 630 & 45.1 & 2201.6  & 79.8 & 79.0 & 72.2 & -- \\
  \checkmark & \checkmark & -- & -- & $\times$
    & 70.1 & 58.1 & 77.2  &  645 & 48.3 & 2217.6 & 80.4 & 80.1 & 73.7 & +1.5 \\
  \checkmark & \checkmark & \checkmark & $\times$ & $\times$
    & 78.3 & 71.8 & 79.3 & 708 & 49.5 & 2330.7 & 80.9 & 80.5 & 78.5 & +4.8 \\
  \rowcolor{lightgray!25}
  \checkmark & \checkmark & \checkmark & \checkmark & $\times$
    & 78.8 & 76.1 & 79.4 & 731 & 49.0
    & 2314.3 & 81.1 & 81.3 & 79.3 & $+0.8$ \\
  \checkmark & \checkmark & \checkmark & \checkmark & \checkmark
    & 77.6  & 76.5 & 79.5 & 735 & 49.3 & 2302.5 & 80.3 &  80.8 & 79.1 & -0.2 \\
  \bottomrule
  \end{tabular}
  }
\end{table}

{
\paragraph{Component ablation on Qwen3-VL-8B.}
We follow the same cumulative protocol on Qwen3-VL-8B, adding facility-location selection, AR,
saliency weighting, the probe-selected head and the spatial kernel in turn
(Tables~\ref{tab:ablation-grounding-qwen3} and~\ref{tab:ablation-vqa-qwen3}). Both components remain
effective on Qwen3-VL: without the bias, facility-location selection improves over random selection
by $5.2$ points on grounding and $8.6$ on VQA, and AR on top of it adds a further $2.1$ and $0.8$.
Saliency weighting and the probe-selected head add $1.3$ and $1.8$ on grounding. On VQA, saliency
averaged over heads does not help ($-0.5$), whereas the probe-selected head adds $3.6$. As on
Qwen2.5-VL, the spatial kernel improves grounding ($+4.8$) and leaves VQA unchanged ($0.0$),
consistent with the task-specific configuration of the paper.
}

{
\begin{table}[t]
  \centering
  
  \caption{\textbf{Component ablation on visual grounding with Qwen3-VL-8B}
  ($9.9\%$ of the visual tokens retained). The shaded row is the deployed configuration.
  $\Delta$ is the gain over the row above.}
  \label{tab:ablation-grounding-qwen3}
  \resizebox{\linewidth}{!}{
  \begin{tabular}{ccccc|ccc|ccc|cc|cc}
  \toprule
  \multicolumn{5}{c|}{\textbf{Component}}
  & \multicolumn{3}{c|}{\textbf{RefCOCO}} & \multicolumn{3}{c|}{\textbf{RefCOCO+}}
  & \multicolumn{2}{c|}{\textbf{RefCOCOg}}
  & \multirow{2}{*}{\textbf{Avg.}} & \multirow{2}{*}{\textbf{$\Delta$}} \\
  \cmidrule(lr){1-5}\cmidrule(lr){6-8}\cmidrule(lr){9-11}\cmidrule(lr){12-13}
  Facility loc. & Bias & Saliency $w$ & Head sel. & Spatial
  & testA & testB & val & testA & testB & val & test & val & & \\
  \midrule
  \rowcolor{lightgray!75}\multicolumn{15}{c}{\textit{Upper Bound, All Visual Tokens} ($\mathbf{100\%}$)} \\
  \rowcolor{lightgray!25}
  \multicolumn{5}{c|}{--} & 90.4 & 84.7 & 88.5 & 85.8 & 75.5 & 81.4 & 87.1 & 86.4 & 85.0 & -- \\
  \midrule
  \rowcolor{lightgray!75}\multicolumn{15}{c}{\textit{No Selection Signal (Random Selection)}} \\
  $\times$ & $\times$ & -- & -- & --
    & 39.4 & 31.5 & 34.1 & 33.5 & 25.7 & 28.3 & 29.3 & 28.6 & 31.3 & -- \\
  $\times$ & \checkmark & -- & -- & --
    & 35.7 & 31.3 & 33.4 & 28.4 & 25.4 & 26.4 & 28.3 & 28.5 & 29.7 & $-1.6$ \\
  \midrule
  \checkmark & $\times$ & -- & -- & $\times$
    & 45.5 & 34.8 & 39.0 & 39.4 & 28.2 & 33.5 & 36.2 & 35.8 & 36.5 & -- \\
  \checkmark & \checkmark & -- & -- & $\times$
    & 46.3 & 39.5 & 41.5 & 38.4 & 30.2 & 34.1 & 39.3 & 39.6 & 38.6 & $+2.1$ \\
  \checkmark & \checkmark & \checkmark & $\times$ & $\times$
    & 46.0 & 38.7 & 43.8 & 38.7 & 32.5 & 36.4 & 41.7 & 41.9 & 40.0 & $+1.3$ \\
  \checkmark & \checkmark & \checkmark & \checkmark & $\times$
    & 48.2 & 41.8 & 45.6 & 39.6 & 33.7 & 37.8 & 43.6 & 44.0 & 41.8 & $+1.8$ \\
  \rowcolor{lightgray!25}
  \checkmark & \checkmark & \checkmark & \checkmark & \checkmark
    & 53.7 & 48.7 & 51.4 & 44.3 & 38.9 & 41.9 & 47.0 & 46.8 & 46.6 & $+4.8$ \\
  \bottomrule
  \end{tabular}
  }
\end{table}
}

{
\begin{table}[t]
  \centering
  
  \caption{\textbf{Component ablation on VQA with Qwen3-VL-8B}
  ($11.1\%$ of the visual tokens retained). The shaded row is the deployed configuration.
  $\Delta$ is the gain over the row above.}
  \label{tab:ablation-vqa-qwen3}
  \resizebox{\linewidth}{!}{
  \begin{tabular}{ccccc|cccccccc|cc}
  \toprule
  \multicolumn{5}{c|}{\textbf{Component}}
  & \multirow{2}{*}{\textsc{TextVQA}} & \multirow{2}{*}{\textsc{ChartQA}}
  & \multirow{2}{*}{\textsc{AI2D}} & \multirow{2}{*}{\textsc{OCRBench}}
  & \multirow{2}{*}{\textsc{HallB}} & \multirow{2}{*}{\textsc{MME}}
  & \multirow{2}{*}{\textsc{MMB}\textsubscript{EN}}
  & \multirow{2}{*}{\textsc{MMB}\textsubscript{CN}}
  & \multirow{2}{*}{\textbf{Acc.}} & \multirow{2}{*}{\textbf{$\Delta$}} \\
  \cmidrule(lr){1-5}
  Facility loc. & Bias & Saliency $w$ & Head sel. & Spatial & & & & & & & & & & \\
  \midrule
  \rowcolor{lightgray!75}\multicolumn{15}{c}{\textit{Upper Bound, All Visual Tokens} ($\mathbf{100\%}$)} \\
  \rowcolor{lightgray!25}
  \multicolumn{5}{c|}{--} & 83.5 & 83.5 & 84.1 & 874 & 57.9 & 2417.2 & 85.4 & 84.9 & 85.9 & -- \\
  \midrule
  \rowcolor{lightgray!75}\multicolumn{15}{c}{\textit{No Selection Signal (Random Selection)}} \\
  $\times$ & $\times$ & -- & -- & --
    & 48.7 & 32.5 & 69.9 & 395 & 37.9 & 1863.0 & 71.3 & 72.8 & 58.2 & -- \\
  $\times$ & \checkmark & -- & -- & --
    & 48.9 & 30.2 & 69.9 & 402 & 40.7 & 1922.8 & 73.5 & 73.3 & 59.1 & $+0.9$ \\
  \midrule
  \checkmark & $\times$ & -- & -- & $\times$
    & 67.9 & 38.1 & 74.7 & 516 & 45.3 & 2115.4 & 75.5 & 75.2 & 66.8 & -- \\
  \checkmark & \checkmark & -- & -- & $\times$
    & 67.4 & 38.8 & 74.4 & 525 & 44.7 & 2137.0 & 78.1 & 77.3 & 67.5 & $+0.8$ \\
  \checkmark & \checkmark & \checkmark & $\times$ & $\times$
    & 68.5 & 40.7 & 73.5 & 490 & 45.5 & 2097.6 & 77.5 & 76.3 & 67.0 & $-0.5$ \\
  \rowcolor{lightgray!25}
  \checkmark & \checkmark & \checkmark & \checkmark & $\times$
    & 72.2 & 46.8 & 75.7 & 564 & 48.1 & 2183.7 & 78.2 & 77.9 & 70.6 & $+3.6$ \\
  \checkmark & \checkmark & \checkmark & \checkmark & \checkmark
    & 70.0 & 50.4 & 76.3 & 555 & 49.9 & 2136.7 & 78.2 & 77.4 & 70.6 & $0.0$ \\
  \bottomrule
  \end{tabular}
  }
\end{table}
}

\subsection{Attention Bias}
\label{sec:appendix:bias}

\paragraph{Full Results of the Bias-Form Ablation.}
\label{sec:appendix:bias-forms}

Tables~\ref{tab:ablation-biasing-grounding} and~\ref{tab:ablation-biasing-vqa} give the per-split
and per-benchmark results of the bias-form ablation, of which Table~\ref{tab:abl-bias} reports
the averages. All rows share the same selector and therefore the same retained tokens, and differ
only in the bias added to the retained visual logits: no bias, the uniform bias $\log(N/k)$ of
Eq.~\ref{eq:bias-uniform}, the per-cell bias $\log(1+|C_j|)$ of Eq.~\ref{eq:bias-accounting} and the
capped bias of Eq.~\ref{eq:bias-capped}. The tables separate the two factors that distinguish these
forms: \emph{Mass} is the total $\sum_{j\in S}e^{b_j}$ that the bias hands to the survivors, which
the condition $\sigma'=\sigma$ fixes at $N$, and \emph{Distribution} is how that total is spread
over them, which the condition leaves free. Rows two and three restore the same mass with
different distributions, rows three and four share a distribution and differ in mass, so each factor
is read off an adjacent pair.

Restoring the attention share helps on both tasks, but the two factors do not act alike. On
grounding all three biases raise the average accuracy from $26.8$ to between $65.5$ and $66.8$, a
spread of at most $1.3$ points, so the gain comes from the mass that is restored; at equal mass, the
constant distribution is $1.3$ points above the per-cell one. On VQA the bias matters far less and
the order reverses: the uniform bias is $0.4$ points below no bias, while the per-cell and capped
forms add $0.5$ and $0.6$. At equal distribution, the capped bias is no worse than the per-cell one
on either task ($66.3$ against $65.5$, $79.3$ against $79.2$). We therefore use the uniform bias for
grounding and the capped bias for VQA.

\begin{table}[t]
\vspace{2mm}
    \centering
    \caption{\textbf{Bias form ablation on visual grounding with Qwen2.5-VL-7B}
    ($9.9\%$ of the visual tokens retained).}
    \label{tab:ablation-biasing-grounding}
    \resizebox{\linewidth}{!}{
    \begin{tabular}{llc|ccc|ccc|cc|c}
    \toprule
    \multirow{2}{*}{\textbf{Bias } $b_j$} & \multirow{2}{*}{\textbf{Distribution}}
    & \multirow{2}{*}{\textbf{Mass}}
    & \multicolumn{3}{c|}{\textbf{RefCOCO}} & \multicolumn{3}{c|}{\textbf{RefCOCO+}}
    & \multicolumn{2}{c|}{\textbf{RefCOCOg}} & \multirow{2}{*}{\textbf{Avg.}} \\
    \cmidrule(lr){4-6}\cmidrule(lr){7-9}\cmidrule(lr){10-11}
    & & & testA & testB & val & testA & testB & val & test & val & \\
    \midrule
    None
      & --- & $k$ & 33.9 & 24.5 & 29.2 & 30.2 & 20.9 & 25.4 & 25.2 & 25.0 & 26.8 \\
    \midrule
    $\log(N/k)$
      & constant & $N$ & 75.4 & 66.7 & 72.2 & 67.2 & 55.9 & 62.3 & 67.8 & 66.5 & \textbf{66.8} \\
    $\log(1+|C_j|)$
      & per-group & $N$ & 74.5 & 66.2 & 70.7 & 66.1 & 54.7 & 61.0 & 65.9 & 65.3 & 65.5 \\
    $\min\!\big(\log(1+|C_j|),\log(N/k)\big)$
      & per-group & $<N$ & 74.5 & 66.8 & 71.1 & 66.6 & 55.9 & 61.5 & 67.2 & 66.6 & 66.3 \\
    \bottomrule
    \end{tabular}
    }
\end{table}

\begin{table}[t]
    \centering
    \caption{\textbf{Bias form ablation on VQA with Qwen2.5-VL-7B}
    ($128$ of $1296$ visual tokens retained).}
    \label{tab:ablation-biasing-vqa}
    \resizebox{\linewidth}{!}{
    \begin{tabular}{llc|cccccccc|c}
    \toprule
    \textbf{Bias } $b_j$ & \textbf{Distribution} & \textbf{Mass}
    & \textsc{TextVQA} & \textsc{ChartQA} & \textsc{AI2D} & \textsc{OCRBench}
    & \textsc{HallB} & \textsc{MME} & \textsc{MMB}\textsubscript{EN}
    & \textsc{MMB}\textsubscript{CN} & \textbf{Acc.} \\
    \midrule
    None
      & --- & $k$ & 78.5 & 74.6 & 79.1 & 732 &  48.8  & 2270.0 & 80.7 & 81.0 & 78.7 \\
    \midrule
    $\log(N/k)$
      & constant & $N$ & 77.8 & 74.2 & 78.3 & 727 & 47.8 & 2298.5 & 80.6 & 80.0 & 78.3 \\
    $\log(1+|C_j|)$
      & per-group & $N$ & 79.0 & 76.3 & 79.3 & 732 & 48.8 & 2302.3 &  80.5 &  81.5 & 79.2 \\
    $\min\!\big(\log(1+|C_j|),\log(N/k)\big)$
      & per-group & $<N$ & 78.8 & 76.1 & 79.4 & 731 & 49.0 & 2314.3 & 81.1 & 81.3 & \textbf{79.3} \\
    \bottomrule
    \end{tabular}
    }
\end{table}

\paragraph{Visual Attention Share after Pruning.}
\label{sec:appendix:mechanism}

\begin{figure}[h]
  \centering
  \includegraphics[width=0.92\linewidth]{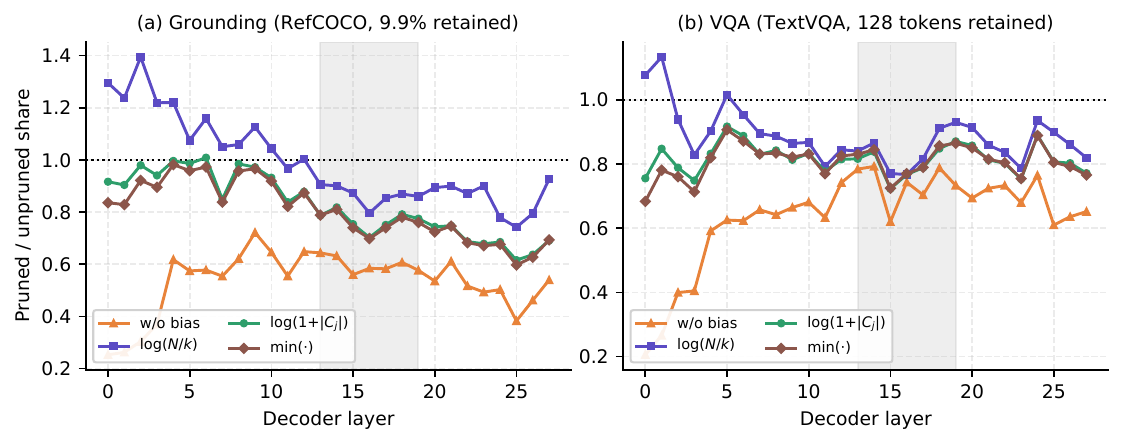}
  \caption{\textbf{Visual attention share restored by each bias, across decoder layers}
  (Qwen2.5-VL-7B, $100$ images per task). The curves share the selector of the deployed
  configuration and differ only in the bias; the dotted line marks the unpruned level and the band
  marks layers $13$ to $19$.}
  \label{fig:bias-share}
\end{figure}

We check how much of the visual attention share each bias restores across the decoder layers. At
every layer we compute the softmax mass that a text query assigns to the visual tokens, divide it by
the value the unpruned model gives on the same image, and average over images and heads; a ratio of
$1$ means the retained tokens receive as much attention as the whole image did. Figure~\ref{fig:bias-share}
shows that without a bias the ratio stays below $1$ at every layer, reaching $0.60$ on grounding and
$0.74$ on VQA over layers $13$ to $19$, where text attends to the image most. The three biases raise
it to $0.87$, $0.77$ and $0.76$ on grounding and to $0.84$, $0.81$ and $0.81$ on VQA. The uniform bias restores
the most and overshoots in the early layers, where it exceeds $1$ on grounding, as expected from a
value that satisfies $\sigma'=\sigma$ only in expectation. How much is restored does not order the
accuracies, however: on VQA the uniform bias restores the most share and is the least accurate of
the three (Table~\ref{tab:ablation-biasing-vqa}). Restoring the share is therefore necessary but
not sufficient, and accuracy also depends on how the restored share is spread over the retained
tokens.

\paragraph{Where the Bias Is Applied.}
\label{sec:appendix:bias-where}

The bias enters the prefill only: the queries of the prefill attend to the retained tokens through
the biased logits, while each decoding step runs the unmodified kernel on the cached keys and values,
so decoding costs the same as in the pruned model (Table~\ref{tab:bias-overhead}).
Table~\ref{tab:bias-where} compares adding the bias at the prefill only with adding it at every
decoding step as well. The two settings share the selector and the retained tokens and differ only
in where the bias is added. They are equally accurate on VQA and within 0.2 points on VG, so adding the
bias at the decoding steps brings no further gain. The correction still reaches the decoding steps, through the prompt tokens. At the
prefill, the bias raises the attention that every prompt token pays to the retained visual tokens,
so their hidden states, and the keys and values computed from them and stored in the cache, carry
the corrected image information. Each generated token attends to these cached prompt tokens, and so
reads the corrected information without a bias of its own. Reading the image through the prompt tokens is how MLLMs normally
work. In the middle layers the question tokens gather the visual information relevant to the
question, which is then passed to the last prompt position that predicts the answer
\citep{zhang2025crossmodalflow}, and a model can still describe an image from the question tokens
alone when the generated tokens are not allowed to attend to the image tokens
\citep{kaduri2025whatsintheimage}.

\begin{table}[h]
  \centering
  \small
  \caption{\textbf{Bias applied at the prefill only or at the prefill and every decoding step}
  (Qwen2.5-VL-7B). VG: average over RefCOCO\textsubscript{testA}, RefCOCO+\textsubscript{testB}
  and RefCOCOg\textsubscript{test} at $9.9\%$ of the visual tokens. VQA: average over eight
  benchmarks at $128$ of $1{,}296$ tokens.}
  \label{tab:bias-where}
  \begin{tabular}{l|cc}
  \toprule
  \textbf{Bias applied at} & \textbf{VG} & \textbf{VQA} \\
  \midrule
  None (no bias)                    & 26.7 & 78.7 \\
  Prefill only (ours)               & 66.4   & 79.3 \\
  Prefill + Decode                  & 66.2 & 79.3 \\
  \bottomrule
  \end{tabular}
  \vspace{-2mm}
\end{table}

\paragraph{Effect of the Bias across Backbones.}
\label{sec:appendix:bias-backbones}
Table~\ref{tab:bias-backbones} compares CAR with and without the bias on the four backbones, at the
lowest retention ratio of each task; the two settings differ only in the bias. On Qwen2.5-VL-7B,
Qwen3-VL-8B and LLaVA-1.5-7B, the bias improves accuracy on every task, and the gain is larger on grounding than on
VQA. It is smaller on Qwen3-VL-8B than on Qwen2.5-VL-7B because the pruned
Qwen3-VL-8B degrades less without the bias ($43.5$ against $26.8$ on grounding at $9.9\%$), which leaves
less for the bias to recover; this may relate to DeepStack, through which Qwen3-VL-8B injects visual
features from several encoder layers into the language model
(Appendix~\ref{sec:appendix:results-qwen3}). On InternVL3.5-8B the bias slightly lowers
the accuracy ($-0.4$ on grounding and $-1.0$ on VQA). This backbone splits an image into tiles and appends a thumbnail, and the
attention that pruning withdraws falls on the tiles, whose content the thumbnail already provides
(Appendix~\ref{sec:appendix:internvl-bias}); we therefore disable the bias on this backbone.
On LLaVA-1.5-7B the bias adds $7.7$ points on grounding ($22.7$ to $30.4$). Like Qwen2.5-VL-7B, it has no
other visual pathway such as DeepStack or a thumbnail, so the pruned model degrades more without the
bias and leaves more for it to recover.

\begin{table}[h]
  \centering
  \small
  \caption{\textbf{CAR with and without the bias across backbones}, at the lowest retention ratio of
  each task. On InternVL3.5-8B, {w/ bias denotes the per-pool bias, computed} separately for the thumbnail and the other tiles {(Appendix~\ref{sec:appendix:internvl-bias})}.}
  \label{tab:bias-backbones}
  \begin{tabular}{ll c|cc}
  \toprule
  \textbf{Backbone} & \textbf{Task} & \textbf{Retained} & \textbf{w/o bias} & \textbf{w/ bias} \\
  \midrule
  \multirow{2}{*}{Qwen2.5-VL-7B}  & VG    & $9.9\%$  & $26.8$ & $66.8$ ($+40.0$) \\
                                  & VQA   & $9.9\%$  & $78.7$ & $79.3$ ($+0.6$) \\
  \midrule
  \multirow{3}{*}{Qwen3-VL-8B}    & VG    & $9.9\%$  & $43.5$ & $46.6$ ($+3.1$) \\
                                  & VQA   & $11.1\%$ & $69.8$ & $70.6$ ($+0.8$) \\
                                  & Video & $9.9\%$  & $60.7$ & $62.0$ ($+1.3$) \\
  \midrule
  \multirow{2}{*}{InternVL3.5-8B} & VG    & $9.9\%$  & $76.2$ & $75.8$ ($-0.4$) \\
                                  & VQA   & $11.1\%$ & $74.4$ & $73.4$ ($-1.0$) \\
  \midrule
  LLaVA-1.5-7B                    & VG    & $11.1\%$ & $22.7$ & $30.4$ ($+7.7$) \\
  \bottomrule
  \end{tabular}
\end{table}

\paragraph{Comparison with RESTORE.}
\label{sec:appendix:restore}

Here we compare with RESTORE, which also calibrates attention after pruning. RESTORE splits the
token budget in half: one half is selected by the underlying pruning method and the other half is
formed by merging the tokens it discards, each merged token receiving on its logit the logarithm of
the number of tokens it absorbs. Since RESTORE ties this merging to its calibration and applies to
different selectors, Tables~\ref{tab:restore} and~\ref{tab:restore-grounding} compare, on DivPrune
and on our selector, its merging alone, its merging together with its bias, and our bias (on VQA the
capped per-cell bias, whose magnitude follows the number of discarded tokens each retained token
stands for under the similarity, an assignment used only for counting; on grounding the uniform
bias).

From Tables~\ref{tab:restore} and~\ref{tab:restore-grounding} we observe the following.
(1) On question answering, the merging of RESTORE gains $1.9$ points on DivPrune and costs $3.4$
points on our selector. The effect of the merging depends on the selector: DivPrune selects by
diversity alone and does not consider importance, so the discarded tokens still hold information
worth recycling; our selection already weights tokens by saliency, and what can be recycled does not
repay the half of the budget that is given up. On grounding the merging costs accuracy on both
selectors. The merging groups tokens by feature similarity, so the members of a merged token may lie far
apart on the grid, which breaks the spatial structure that grounding depends on.
(2) The bias of RESTORE behaves very differently on the two tasks. On question answering it has
almost no effect, whereas on grounding it gains $19.5$ and $30.6$ points. Grounding tasks are more
sensitive to pruning, and restoring the visual attention share helps the model locate answers
within the image.
(3) Our bias gains on both tasks and on both selectors, and on three of the four combinations it is
clearly better than the full form of RESTORE. On question answering with DivPrune, our bias comes close
to the gain that their merging brings.

\begin{table}[t]
  \centering
  \caption{\textbf{Comparison with RESTORE~\citep{cho2026restore} on VQA with Qwen2.5-VL-7B.}
  Each calibration, RESTORE's and ours, is applied on top of both selectors.}
  \label{tab:restore}
  \resizebox{\linewidth}{!}{
    \begin{tabular}{l|cccccccc|cc}
    \toprule
    \textbf{Method} & \textbf{TextVQA} & \textbf{ChartQA} & \textbf{AI2D} & \textbf{OCRBench} & \textbf{HallBench} & \textbf{MME} & \textbf{MMB-EN} & \textbf{MMB-CN} & \textbf{Acc.} & \textbf{Rel.} \\
    \midrule
    \rowcolor{lightgray!75}\multicolumn{11}{c}{\textit{Upper Bound, All 1{,}296 Tokens} (100\%)} \\
    \rowcolor{lightgray!25} Qwen2.5-VL-7B & 84.3 & 87.1 & 81.6 & 867 & 56.2 & 2342.3 & 83.1 & 82.9 & 84.9 & 100.0\% \\
    \rowcolor{lightgray!75}\multicolumn{11}{c}{\textit{Retain 128 Tokens} ($\downarrow$90.1\%)} \\
    DivPrune{\small\texttt{(CVPR25)}}, no calibration & 65.9 & 51.4 & 72.7 & 535 & 43.2 & 2044.6 & 77.3 & 76.7 & 67.9 & 79.7\% \\
    \ + RESTORE merging & 65.8 & 51.2 & 73.9 & 582 & 43.6 & 2156.4 & 79.1 & 78.6 & 69.8 & 81.8\% \\
    \ + RESTORE merging and bias & 65.3 & 51.2 & 73.8 & 578 & 43.3 & 2183.3 & 79.6 & 79.6 & 70.0 & 81.9\% \\
    \ + our bias & 66.2 & 53.2 & 73.7 & 549 & 44.2 & 2119.2 & 78.5 & 77.9 & 69.3 & 81.4\% \\
    \midrule
    Ours, no calibration & 78.5 & 74.6 & 79.1 & 732 & 48.8 & 2270.0 & 80.7 & 81.0 & 78.7 & 92.3\% \\
    \ + RESTORE merging & 74.0 & 64.6 & 78.2 & 665 & 47.3 & 2242.4 & 79.8 & 80.1 & 75.3 & 88.4\% \\
    \ + RESTORE merging and bias & 73.6 & 63.5 & 78.2 & 654 & 46.8 & 2255.4 & 80.8 & 80.3 & 75.2 & 88.1\% \\
    \ + our bias (\textbf{CAR}) & \textbf{78.8} & \textbf{76.1} & \textbf{79.4} & \textbf{731} & \textbf{49.0} & \textbf{2314.3} & \textbf{81.1} & \textbf{81.3} & \textbf{79.3} & \textbf{93.0\%} \\
    \bottomrule
    \end{tabular}
  }
\end{table}

\begin{table}[t]
  \centering
  \caption{\textbf{Comparison with RESTORE~\citep{cho2026restore} on visual grounding with
  Qwen2.5-VL-7B.} Each calibration, RESTORE's and ours, is applied on top of both selectors.}
  \label{tab:restore-grounding}
  \resizebox{\linewidth}{!}{
    \begin{tabular}{l|ccc|ccc|cc|cc}
    \toprule
    \multirow{2}{*}{\textbf{Method}}
    & \multicolumn{3}{c|}{\textbf{RefCOCO}} & \multicolumn{3}{c|}{\textbf{RefCOCO+}}
    & \multicolumn{2}{c|}{\textbf{RefCOCOg}}
    & \multirow{2}{*}{\textbf{Avg.}} & \multirow{2}{*}{\textbf{Rel.}} \\
    & testA & testB & val & testA & testB & val & test & val & & \\
    \midrule
    \rowcolor{lightgray!75}\multicolumn{11}{c}{\textit{Upper Bound, All Visual Tokens} (100\%)} \\
    \rowcolor{lightgray!25} Qwen2.5-VL-7B & 92.2 & 84.8 & 89.4 & 87.6 & 74.1 & 82.8 & 86.7 & 86.7 & 85.5 & 100.0\% \\
    \rowcolor{lightgray!75}\multicolumn{11}{c}{\textit{Retain 9.9\% of Visual Tokens}} \\
    DivPrune{\small\texttt{(CVPR25)}}, no calibration & 27.8 & 19.6 & 24.2 & 24.7 & 15.8 & 20.9 & 21.3 & 21.5 & 22.0 & 25.6\% \\
    \ + RESTORE merging & 15.7 & 13.4 & 13.9 & 13.9 & 11.4 & 12.1 & 12.3 & 12.4 & 13.1 & 15.3\% \\
    \ + RESTORE merging and bias & 37.1 & 31.8 & 35.7 & 33.4 & 26.7 & 30.6 & 33.4 & 31.8 & 32.6 & 38.0\% \\
    \ + our bias & 51.5 & 42.5 & 46.9 & 46.1 & 35.5 & 40.8 & 46.3 & 45.4 & 44.4 & 51.8\% \\
    \midrule
    Ours, no calibration & 33.9 & 24.5 & 29.2 & 30.2 & 20.9 & 25.4 & 25.2 & 25.0 & 26.8 & 31.2\% \\
    \ + RESTORE merging & 16.8 & 13.8 & 16.2 & 15.1 & 12.0 & 14.5 & 13.2 & 13.0 & 14.3 & 16.7\% \\
    \ + RESTORE merging and bias & 52.1 & 45.1 & 48.1 & 46.7 & 37.3 & 41.9 & 45.2 & 43.1 & 44.9 & 52.4\% \\
    \ + our bias (\textbf{CAR}) & \textbf{75.4} & \textbf{66.7} & \textbf{72.2} & \textbf{67.2} & \textbf{55.9} & \textbf{62.3} & \textbf{67.8} & \textbf{66.5} & \textbf{66.8} & \textbf{77.9\%} \\
    \bottomrule
    \end{tabular}
  }
\end{table}

\subsection{Selection Rule}
\label{sec:appendix:selection}

\paragraph{Selection Mechanisms.}
\label{sec:appendix:taxonomy}

To assess the weighted coverage selection, we compare it with other selection rules under the same
setting: every rule receives the same saliency $w$ from the probed head, the same budget, the same
bias and the same feature space, and every rule builds the retained set greedily, one token at a
time, so that they differ only in what each step selects on. The groups follow
Tables~\ref{tab:ablation-selection-full} and~\ref{tab:ablation-selection-vqa}.
\emph{No diversity signal}: \rnum{1} the $k$ tokens of largest saliency are kept, and the redundancy
among them is not considered (FastV, SparseVLM~\citep{chen2024fastv,zhang2024sparsevlm}).
\emph{Marginal penalty relative to the selected set}: these rules compute at each step the largest
similarity between a candidate and the selected set. \rnum{2} Taking the candidate whose largest similarity is
smallest maximizes the diversity among the retained tokens, and the saliency is not used
(DivPrune~\citep{alvar2025divprune}). \rnum{3} Subtracting that similarity from the saliency, with
$\lambda$ setting the weight of the two terms, takes the candidate that is salient and at the same
time unlike the tokens already selected (IDPruner~\citep{carbonell1998use,tan2026idpruner}).
\rnum{4} Multiplying the saliency by the
dissimilarity to the selected set keeps a token only when both are high
(AnchorPrune~\citep{oh2026anchorprune,xu2026score}).
\emph{Set-function maximization}: \rnum{5} a kernel is built from the saliency and the similarity, and
its log-determinant measures the volume the retained tokens span, so it scores importance and
redundancy at once. Each step takes the largest gain in that volume, which keeps the retained set
salient and non-redundant (CDPruner~\citep{zhangcdpruner}). \rnum{6} The saliency of the retained tokens is instead added to the log-determinant of their feature Gram matrix, with $\alpha$ weighting the two terms (CoIn~\citep{du2026coin}).
\emph{Hard suppression}: \rnum{7} tokens are scanned in order of saliency and, after each selection,
every candidate whose similarity to it exceeds a threshold is removed
(AgilePruner~\citep{baek2026agilepruner}); \rnum{8} a part of the budget first goes to pivots ranked by saliency.
A graph is then built from feature similarity fused with spatial adjacency, and every time a token
is selected its neighbours on the graph are removed, until $k$ tokens have been chosen
(D\textsuperscript{2}Pruner~\citep{zhang2026d2pruner}).
\emph{Weighted coverage selection}: these rules take at each step the largest gain in how well the retained set covers the whole image. \rnum{9} The gain counts every covered token equally and is multiplied by the saliency of the candidate (SCOPE~\citep{deng2025scope}); \rnum{10} ours weights each covered token by its saliency instead, so a candidate scores high when it covers many salient tokens, and multiplies the similarity by a spatial kernel (Eq.~\ref{eq:fl-kernel}) so that tokens far apart on the grid do not cover each other.

\paragraph{Full Results of the Selection-Rule Ablation.}
\label{sec:appendix:selection-full}

Tables~\ref{tab:ablation-selection-full} and~\ref{tab:ablation-selection-vqa} give the detailed
results of the selection-rule ablation on grounding and on VQA, of which
Table~\ref{tab:abl-select} reports the averages. Three observations follow.
(1) The coverage-based rules, \rnum{9} and \rnum{10}, give the best average over the two tasks. Their
objective is how well the retained set covers all the visual tokens, while the other rules examine
only the similarity among the retained tokens. Ours is the best on every grounding split, where the
spatial kernel adds to the coverage, and attains the highest VQA average.
(2) Using the saliency together with the diversity is in general better than using either alone. The
rule that uses the diversity alone, \rnum{2}, is clearly below the rest, and when the importance of a
token is not considered the tokens the task needs may not be retained. The rule that uses the
saliency alone, \rnum{1}, is not the best on either task either, since the tokens it selects are often
redundant and the retained set tends to concentrate on a few regions.
(3) The relative order of the rules is not the same on the two tasks. On grounding the rules that
suppress by a threshold, \rnum{7} and \rnum{8}, stand above \rnum{3}, \rnum{4} and \rnum{5}, as
grounding depends on the spatial structure of the image and removing similar tokens entirely helps
spread the retained tokens over different regions. On VQA the rules other than \rnum{2} and \rnum{6}
lie close together, and the suppression rules fall even below \rnum{1}, as text and charts consist
of many mutually similar tokens and a threshold may remove the salient ones among them as a group.

\begin{table}[t]
    \centering
    \caption{\textbf{Selection rule ablation on visual grounding with Qwen2.5-VL-7B}
    (three test splits, $9.9\%$ of the visual tokens retained). Every row uses the uniform bias
    $\log(N/k)$ of Eq.~\ref{eq:bias-uniform}. For our row, $s_{ij}$ is the
    similarity of Eq.~\ref{eq:fl-kernel}, with the spatial kernel, as deployed; among the other
    rules only D\textsuperscript{2}Pruner carries a geometric term.}
    \label{tab:ablation-selection-full}
    \resizebox{\linewidth}{!}{
    \begin{tabular}{cll|ccc}
    \toprule
    & \textbf{Score or objective} & \textbf{Representative}
    & RefCOCO\textsubscript{testA} & RefCOCO+\textsubscript{testB}
    & RefCOCOg\textsubscript{test} \\
    \midrule
    \rowcolor{lightgray!75}\multicolumn{6}{c}{\textit{No Diversity Signal}} \\
    \rnum{1} &$w_i$
      & FastV & 65.1 & 50.5 & 60.1 \\
    \midrule
    \rowcolor{lightgray!75}\multicolumn{6}{c}{\textit{Marginal Penalty Relative to the Selected Set}} \\
    \rnum{2} &$\min_{j\in S}(1-s_{ij})$, $w$ unused
      & DivPrune & 51.5 & 35.5 & 46.3 \\
    \rnum{3} &$\lambda w_i-(1-\lambda)\max_{j\in S}s_{ij}$
      & IDPruner & 65.5 & 51.1 & 60.7 \\
    \rnum{4} &$w_i^{\alpha}\min_{j\in S}(1-s_{ij})$
      & AnchorPrune & 66.9 & 50.0 & 61.8 \\
    \midrule
    \rowcolor{lightgray!75}\multicolumn{6}{c}{\textit{Set-Function Maximization}} \\
    \rnum{5} &$\log\det K_S$, $K_{ij}=w_iw_js_{ij}$
      & CDPruner & 65.8 & 51.0 & 61.8 \\
    \rnum{6} &$(1-\alpha)w_i+\alpha\,\Delta_i\log\det(X_SX_S^{\top}+\lambda I)$
      & CoIn & 58.3 & 44.0 & 55.2 \\
    \midrule
    \rowcolor{lightgray!75}\multicolumn{6}{c}{\textit{Hard Suppression}} \\
    \rnum{7} &$w_i$ subject to $\max_{j\in S}s_{ij}<\tau$
      & AgilePruner & 68.1 & 51.0 & 62.5 \\
    \rnum{8} &Pivots by $w_i$, then graph suppression
      & D\textsuperscript{2}Pruner & 68.9 & 53.3 & 62.1 \\
    \midrule
    \rowcolor{lightgray!75}\multicolumn{6}{c}{\textit{Weighted Coverage Selection}} \\
    \rnum{9} &$w_i\sum_{l\in V}\big[s_{il}-\max_{j\in S}s_{jl}\big]_+$
      & SCOPE & 70.7 & 52.3 & 63.2 \\
    \rowcolor{lightgray!25}
    \rnum{10} & \textbf{$\sum_{i\in V}w_i\max_{j\in S}s_{ij}$ (ours)}
      & --- & \textbf{75.4} & \textbf{55.9} & \textbf{67.8} \\
    \bottomrule
    \end{tabular}
    }
  \end{table}
\begin{table}[t]
  \centering
  \caption{\textbf{Selection rule ablation on VQA with Qwen2.5-VL-7B}
  ($128$ of $1296$ visual tokens retained). No row uses the bias. For our row, $s_{ij}$ is the similarity of
  Eq.~\ref{eq:fl-ours}, without the spatial kernel, as deployed.}
  \label{tab:ablation-selection-vqa}
  \resizebox{\linewidth}{!}{
  \begin{tabular}{cll|cccccccc|c}
  \toprule
  & \textbf{Score or objective} & \textbf{Representative}
  & TextVQA & ChartQA & AI2D & OCRBench & HallBench & MME & MMB-EN & MMB-CN
  & \textbf{Avg} \\
  \midrule
  \rowcolor{lightgray!75}\multicolumn{12}{c}{\textit{No Diversity Signal}} \\
  \rnum{1} &$w_i$
    & FastV & 76.8 & 73.0 & 78.6 & 724 & 49.9 & 2278.1 & 80.1  & 80.4 & 78.1  \\ 
  \midrule
  \rowcolor{lightgray!75}\multicolumn{12}{c}{\textit{Marginal Penalty Relative to the Selected Set}} \\
  \rnum{2} &$\min_{j\in S}(1-s_{ij})$, $w$ unused
    & DivPrune & 65.8 & 51.4 & 72.7 & 533 & 42.7 & 2041.2 & 77.4 & 76.7 & 67.8 \\
  \rnum{3} &$\lambda w_i-(1-\lambda)\max_{j\in S}s_{ij}$
    & IDPruner & 77.3 & 73.4 & 78.6 & 733 & 49.3 & 2278.1 & 80.7 & 80.4 & 78.4 \\
  \rnum{4} &$w_i^{\alpha}\min_{j\in S}(1-s_{ij})$
    & AnchorPrune & 77.3 & 75.4 & 79.1 & 736 & 48.2 & 2282.7 & 80.6 & 80.8 & 78.6 \\
  \midrule
  \rowcolor{lightgray!75}\multicolumn{12}{c}{\textit{Set-Function Maximization}} \\
  \rnum{5} &$\log\det K_S$, $K_{ij}=w_iw_js_{ij}$
    & CDPruner & 77.8 & 73.3 &  79.2 & 725 & 48.4 & 2250.6 & 80.3 & 80.7 & 78.1 \\
  \rnum{6} &$(1-\alpha)w_i+\alpha\,\Delta_i\log\det(X_SX_S^{\top}+\lambda I)$
    & CoIn & 76.0 & 64.6 & 77.0 & 650 & 46.7 & 2170.3 & 78.4 & 78.3 & 74.3 \\
  \midrule
  \rowcolor{lightgray!75}\multicolumn{12}{c}{\textit{Hard Suppression}} \\
  \rnum{7} &$w_i$ subject to $\max_{j\in S}s_{ij}<\tau$
    & AgilePruner & 76.2 & 69.7 & 77.7 & 692 & 46.4 & 2211.0 & 79.8 & 79.3 & 76.1 \\ 
  \rnum{8} &Pivots by $w_i$, then graph suppression
    & D\textsuperscript{2}Pruner & 75.6 & 69.5 & 78.9 & 685 & 49.4 & 2271.9 & 80.2 & 80.9 & 77.1 \\ 
  \midrule
  \rowcolor{lightgray!75}\multicolumn{12}{c}{\textit{Weighted Coverage Selection}} \\
  \rnum{9} &$w_i\sum_{l\in V}\big[s_{il}-\max_{j\in S}s_{jl}\big]_+$
    & SCOPE & 77.8 & 73.0 & 78.4 & 734 & 48.7 & 2276.6 & 80.7 & 81.4 & 78.4 \\
  \rowcolor{lightgray!25}
  \rnum{10} & \textbf{$\sum_{i\in V}w_i\max_{j\in S}s_{ij}$ (ours)}
    & --- & 78.5 & 74.6 & 79.1 & 732 & 48.8
  & 2270.0 & 80.7 & 81.0 & 78.7 \\
  \bottomrule
  \end{tabular}
  }
  \vspace{-4mm}
\end{table}

\subsection{Error Decomposition}
\label{sec:appendix:error-decomp}

This section describes how the two errors in Figure~\ref{fig:error-ablation} are measured. The
model itself is not pruned during the measurement. We run one full forward pass of the unpruned
model and, at every layer, compute two outputs from that layer's input: the output $O$ of the full
attention, and the output $O'$ obtained when only the tokens $S$ selected by the method are
retained and the bias $b$ is added. For methods that prune inside the decoder, $S$ and the bias
$\log(N/k_l)$ both follow the tokens the method actually retains at layer $l$. We then split
$O'-O$ into the two errors of Eq.~\ref{eq:decomp}. Since the input of every layer comes from
the unpruned model, the measurement captures the error introduced by pruning at that layer alone,
without the drift accumulated in earlier layers. Each error is aggregated over layers $13$ to $19$,
the intermediate layers marked by the band in Figure~\ref{fig:phenomena}, and reported as its
squared norm relative to that of $O$, in per cent. The samples are drawn from subsets of the
RefCOCO, RefCOCO+ and RefCOCOg validation sets, $100$ from each, $300$ in total.

\subsection{Details of the Motivating Observations}
\label{sec:appendix:motivation}

This section gives the setup behind the three groups of panels of
Figure~\ref{fig:phenomena}.

\paragraph{Grounding across retention ratios (panel a).}
We evaluate Qwen2.5-VL-7B~\citep{Bai2025qwen2.5-vl} on the eight splits of RefCOCO, RefCOCO+ and
RefCOCOg~\citep{yu2016modeling, mao2016generation}. The panel reports the aggressive retention ratio
of $9.9\%$, without and with the bias. The nine published methods are grouped by where the signal
that decides the retained set is read: encoder-side (VisionZip~\citep{yang2025visionzip},
DivPrune~\citep{alvar2025divprune}, AnchorPrune~\citep{oh2026anchorprune},
EADP~\citep{wang2026eadp}); decoder-side (FastV~\citep{chen2024fastv},
SparseVLM~\citep{zhang2024sparsevlm}, HAWK~\citep{zhu2026hawk}); and two-stage
(VScan~\citep{zhang2025vscan}, N\"uwa~\citep{huang2026nuwa}). We add two simple baselines
that carry no selection signal, uniform random sampling and average pooling, which merges
neighbouring visual tokens into $k$ averaged ones. The bias is the uniform $\log(N/k)$ of
Eq.~\ref{eq:bias-uniform}, added to the retained visual tokens only and leaving the set that the
selector retains unchanged.

Table~\ref{tab:grounding-budget} gives the per-split numbers at a moderate retention ratio of
$25\%$ and at $9.9\%$, each without and with the bias. At $25\%$ the encoder-side and two-stage
methods are clearly ahead of the two baselines, while the decoder-side ones stay below them at
either ratio; at $9.9\%$ the lead of the best method, N\"uwa, over random sampling falls from
$16.4$ points to $4.0$. The bias improves every method at both ratios; at $9.9\%$ it raises the published methods by up to $24.1$ points and the pooling baseline by $33.0$. With our bias, the simple pooling baseline reaches $49.1$ at $9.9\%$ and so outperforms
existing complex pruning designs, the best of them being DivPrune at $44.4$. 
CAR is the best at both ratios, $79.4$ and $66.8$, ahead of the
strongest biased baseline by $3.3$ and $17.7$ points.

\begin{table}[t]
  \centering
  \caption{\textbf{Grounding on Qwen2.5-VL-7B at a moderate and an aggressive retention ratio.}
  $\Delta$ is the gain of the uniform bias on the eight-split average.}
  \label{tab:grounding-budget}
  \setlength{\aboverulesep}{0.2ex}
  \setlength{\belowrulesep}{0.2ex}
  \renewcommand{\arraystretch}{0.90}
  \resizebox{\linewidth}{!}{
    \begin{tabular}{l|ccc|ccc|cc|cc}
    \toprule
    \multirow{2}{*}{\textbf{Method}}
    & \multicolumn{3}{c|}{\textbf{RefCOCO}} & \multicolumn{3}{c|}{\textbf{RefCOCO+}}
    & \multicolumn{2}{c|}{\textbf{RefCOCOg}} & \multirow{2}{*}{\textbf{Avg.}}
    & \multirow{2}{*}{\textbf{$\Delta$}} \\
    & testA & testB & val & testA & testB & val & test & val & & \\
    \rowcolor{lightgray!75}\multicolumn{11}{c}{\textit{Upper Bound, All Visual Tokens} ($\mathbf{100\%}$)} \\
    \rowcolor{lightgray!25} Qwen2.5-VL-7B & 92.2 & 84.8 & 89.4 & 87.6 & 74.1 & 82.8 & 86.7 & 86.7 & 85.5 & -- \\
    \rowcolor{lightgray!75}\multicolumn{11}{c}{\textit{Moderate Budget, Retain} $\mathbf{25\%}$ \textit{of Visual Tokens}} \\
    Random & 63.9 & 52.1 & 58.3 & 58.1 & 44.5 & 52.5 & 52.7 & 52.7 & 54.4 & -- \\
    +bias &  70.1 &  63.4 &  67.8 &  63.6 &  53.9 &  60.6 &  62.9 &  63.0 &       63.2 & $+8.8$ \\
    Pooling & 64.1 & 49.1 & 57.5 & 58.8 & 42.0 & 51.6 & 50.7 & 50.3 & 53.0 & -- \\
    +bias &  76.9 &  66.8 &  73.4 &  70.7 &  56.7 &  66.2 &  68.7 &  68.1 &       68.4 & $+15.4$ \\
    VisionZip{\small\texttt{(CVPR25)}} & 71.0 & 59.0 & 66.7 & 65.0 & 50.5 & 59.1 & 61.2 & 61.2 & 61.7 & -- \\
    +bias &  77.4 &  71.4 &  75.5 &  70.8 &  60.8 &  66.8 &  72.8 &  72.7 &   71.0 & $+9.3$ \\
    DivPrune{\small\texttt{(CVPR25)}} &  74.1 & 59.8 & 67.5 & 68.6 &  51.9 &  60.9 &  63.3 &  63.7 &  63.7 & -- \\
    +bias &  81.2 &  70.6 &  76.9 &  76.1 &  60.8 &  69.0 &  74.3 &  74.5 &       72.9 & $+9.2$ \\
     AnchorPrune{\small\texttt{(ECCV26)}} & 75.6 & 62.4 & 68.3 & 68.4 & 53.2 & 61.5 & 64.7 & 65.5 & 64.9 & -- \\
     +bias &  81.6 &  72.9 &  77.1 &  75.1 &  62.9 &  69.0 &  74.8 &  76.2 &       73.7 & $+8.8$ \\
    EADP{\small\texttt{(ECCV26)}} & 71.9 & 53.1 & 63.3 & 68.0 & 44.4 & 57.2 & 56.9 & 56.3 & 58.9 & -- \\
    +bias &  79.6 &  66.3 &  74.4 &  74.9 &  55.9 &  67.3 &  69.7 &  69.8 &       69.7 & $+10.8$ \\
    FastV{\small\texttt{(ECCV24)}} & 23.7 & 30.7 & 28.4 & 21.6 & 25.6 & 24.1 & 29.1 & 29.2 & 26.6 & -- \\
    +bias &  27.7 &  37.8 &  33.9 &  25.5 &  32.7 &  29.5 &  36.8 &  37.5 &       32.7 & $+6.1$ \\
    SparseVLM{\small\texttt{(ICML25)}} & 30.9 & 31.7 & 31.3 & 26.4 & 25.3 & 24.9 & 29.3 & 29.9 & 28.7 & -- \\
    +bias &  39.4 &  40.7 &  40.7 &  34.3 &  32.4 &  34.3 &  39.5 &  39.1 &       37.6 & $+8.9$ \\
     HAWK{\small\texttt{(CVPR26)}} & 40.4 &  45.7 &  42.7 & 36.0 &  40.0 &  38.0 &  44.9 &  45.7 &  41.7 & -- \\
     +bias &  46.4 &  53.1 &  49.3 &  42.3 &  45.7 &  44.2 &  53.8 &  54.6 &       48.7 & $+7.0$ \\
     VScan{\small\texttt{(TMLR26)}} & 78.7 & 66.9 & 71.7 & 72.3 & 55.5 & 64.0 & 67.9 & 68.2 & 68.2 & -- \\
     +bias &  84.3 &  74.8 &  80.4 &  78.4 &  63.5 &  72.2 &  77.5 &  77.5 &       76.1 & $+7.9$ \\
    N\"uwa{\small\texttt{(ICLR26)}} & 77.5 & 72.5 & 75.6 & 69.7 & 60.3 & 65.8 & 72.8 & 72.3 & 70.8 & -- \\
    +bias  &  81.5 & 74.3 & 77.5 & 73.7 & 61.2 & 68.2 & 74.7 & 75.0 & 73.3 & $+2.5$ \\
    Ours w/o bias & 79.8 & 66.6 & 75.3 & 73.0 & 58.5 & 68.1 & 67.5 & 67.9 & 69.6 & -- \\
    \textbf{CAR} (Ours) &  87.0 & 80.1 & 84.3 & 80.2 & 68.3 & 75.9 & 79.4 & 79.6 & 79.4 & $+9.8$ \\
    \rowcolor{lightgray!75}\multicolumn{11}{c}{\textit{Aggressive Budget, Retain} $\mathbf{9.9\%}$ \textit{of Visual Tokens}} \\
    Random & 23.1 & 17.2 & 21.5 & 20.3 & 14.5 & 18.6 & 17.3 & 18.1 & 18.8 & -- \\
    +bias & 44.0 & 37.6 & 42.9 & 38.3 & 31.7 & 37.0 & 39.0 & 38.4 & 38.6 & $+19.8$ \\
    Pooling & 20.0 & 15.6 & 17.5 & 18.2 & 13.0 & 15.5 & 14.8 & 14.2 & 16.1 & -- \\
    +bias & 58.7 & 46.7 & 53.6 & 51.5 &  38.8 & 46.7 &  48.0 & 48.9 & 49.1 & $+33.0$ \\
    VisionZip{\small\texttt{(CVPR25)}} & 23.3 & 19.2 & 20.5 & 18.8 & 16.0 & 17.7 & 18.4 & 19.5 & 19.2 & -- \\
    +bias & 41.3 & 42.4 & 42.4 & 35.6 & 35.5 & 36.1 & 40.8 & 41.8 & 39.5 & $+20.3$ \\
    DivPrune{\small\texttt{(CVPR25)}} & 27.8 & 19.6 & 24.2 & 24.7 & 15.8 & 20.9 & 21.3 & 21.5 & 22.0 & -- \\
    +bias & 51.5 & 42.5 & 46.9 & 46.1 & 35.5 & 40.8 & 46.3 & 45.4 & 44.4 & $+22.4$ \\
    AnchorPrune{\small\texttt{(ECCV26)}} & 24.4 & 18.0 & 22.0 & 21.4 & 16.2 & 19.4 & 19.7 & 20.2 & 20.2 & -- \\
    +bias & 46.8 & 44.2 & 44.2 & 40.8 & 37.3 & 39.2 & 44.8 & 45.7 & 42.9 & $+22.7$ \\
    EADP{\small\texttt{(ECCV26)}} & 24.1 & 16.5 & 21.2 & 23.4 & 14.6 & 18.6 & 17.4 & 18.2 & 19.2 & -- \\
    +bias & 51.0 & 39.3 & 47.8 & 46.2 & 34.6 & 41.7 & 42.3 & 43.3 & 43.3 & $+24.1$ \\
    FastV{\small\texttt{(ECCV24)}} & 3.2 & 2.7 & 3.2 & 3.3 & 2.9 & 3.5 & 3.6 & 3.3 & 3.2 & -- \\
    +bias & 5.7 & 3.9 & 5.6 & 6.1 & 4.3 & 4.9 & 5.8 & 5.5 & 5.2 & $+2.0$ \\
    SparseVLM{\small\texttt{(ICML25)}} & 5.7 & 5.7 & 5.8 & 3.8 & 4.2 & 3.8 & 5.4 & 5.3 & 5.0 & -- \\
    +bias & 7.8 & 7.4 & 7.7 & 4.8 & 5.1 & 5.1 & 7.0 & 6.8 & 6.5 & $+1.5$ \\
    HAWK{\small\texttt{(CVPR26)}} &  8.1 &  8.6 & 7.5 & 6.9 & 7.0 &  7.0 &  8.5 & 7.5 &  7.6 & -- \\
    +bias & 9.9 & 13.1 & 12.4 & 9.0 & 11.4 & 11.0 & 14.9 & 14.4 & 12.0 & $+4.4$ \\
    VScan{\small\texttt{(TMLR26)}} & 23.8 & 19.8 & 21.4 & 20.4 & 15.5 & 18.0 & 18.5 & 20.5 & 19.7 & -- \\
    +bias & 48.0 & 40.5 & 43.8 & 42.0 & 31.6 & 36.4 & 41.6 & 42.0 & 40.7 & $+21.0$ \\
    N\"uwa{\small\texttt{(ICLR26)}} & 27.5 & 23.7 & 26.2 & 21.8 & 17.4 & 20.5 & 23.0 & 22.0 & 22.8 & -- \\
    +bias & 48.2 & 43.9 & 47.6 & 40.5 & 35.5 & 38.7 & 43.6 & 44.1 & 42.8 & $+20.0$ \\
    Ours w/o bias & 33.9 & 24.5 & 29.2 & 30.2 & 20.9 & 25.4 & 25.2 & 25.0 & 26.8 & -- \\
    \textbf{CAR} (Ours) &  75.4 & 66.7 & 72.2 & 67.2 & 55.9 & 62.3 & 67.8 & 66.5 & 66.8 & $+40.0$ \\
    \bottomrule
    \end{tabular}
  }
  \vspace{-4mm}
\end{table}

\paragraph{Visual attention share across decoder layers (panels b to d).}
At every decoder layer of Qwen2.5-VL-7B we measure $\sigma$, the share of the attention of text
tokens that falls on visual tokens, averaged over heads and over text queries. Panel (b) gives
$\sigma$ in the unpruned model on $100$ RefCOCO val images and $100$ TextVQA images: on both tasks
it peaks in layers $13$ to $19$, the layers where text tokens attend to the image most. Panels (c)
and (d) prune the $100$ RefCOCO val images to a retention ratio of $9.9\%$ and report the ratio
$\sigma'/\sigma$ of Eq.~\ref{eq:blend}, the share after pruning over the share the unpruned model
gives on the same image. They compare CAR with DivPrune, HAWK, uniform random sampling and
average pooling, without any bias in (c) and, in (d), with the uniform bias of
Eq.~\ref{eq:bias-uniform} added to the tokens each method retains. In layers $13$ to $19$, pruning
lowers the share to between $0.51$ and $0.64$ of its unpruned value, and the bias raises it to
between $0.70$ and $0.94$.

\paragraph{Saliency source (panels e and f).}
In panels (e) and (f) we hold the token selection fixed to our weighted coverage selection of
Section~\ref{sec:method:select} and vary only the source that supplies the saliency $w$. For
Qwen2.5-VL-7B the candidates are the $16$ attention heads of the final encoder layer and their
average. We report grounding accuracy on $500$ RefCOCO val images at a $9.9\%$ retention ratio and
question answering accuracy averaged over the eight benchmarks at $128$ tokens. There is significant
variation in importance among different attention heads, with accuracy differences exceeding $30$
percentage points on the grounding task. Furthermore, attention heads exhibit task dependency; the
head performing best on one task does not necessarily excel on another: head $13$ leads in the
grounding task, while head $2$ leads in the question-answering task. For InternVL3.5-8B, the late
layers develop attention sinks~\citep{darcet2024vision, luo2026sink}, where a few sink tokens
absorb most of the attention (Appendix~\ref{sec:appendix:internvl-attn}), which leaves the final
layer the worst source. We therefore compare the layers that supply $w_i$, averaging over the
heads of each, and scan them at an $11.1\%$ retention ratio on $500$ samples of each of the two
question answering benchmarks. The results show that performance on both TextVQA and ChartQA peaks
at the intermediate layers, which the band marks.

\subsection{Hybrid-Attention Models: Qwen3.5-9B and Qwen3.8-27B}
\label{sec:appendix:hybrid}
{
We further evaluate CAR on Qwen3.5-9B and Qwen3.8-27B, retaining $9.9\%$ of the visual tokens on
grounding and $11.1\%$ on VQA (Tables~\ref{tab:qwen35-grounding}--\ref{tab:qwen38-vqa}).

\paragraph{Extending CAR to hybrid attention.}
Most decoder layers of both models are Gated DeltaNet (GDN) layers~\citep{yang2025gated} ($24$ of
$32$ in Qwen3.5-9B, $48$ of $64$ in Qwen3.8-27B). AR uses a logit bias to let the retained tokens
carry the attention mass of the removed tokens, which requires a softmax. We therefore apply AR only in the full-attention layers, whose softmax attention is the same as in the other backbones.
A GDN layer has none: it
writes the tokens one by one into a fixed-size recurrent state. After pruning, fewer visual tokens
are written over a shorter timeline, so both the share of the image in the state and the
proportions among its contents deviate from those of the unpruned model, which a logit bias cannot
correct. We therefore extend the idea of AR to GDN layers: each representative explicitly writes on
behalf of the removed tokens it covers.

\paragraph{GDN representative replay.}
Each removed visual token is represented by the retained token most similar to it, the same
assignment that defines the cells of AR. Within each GDN layer, the projections are computed on the
pruned sequence, expanded to the original length by filling each removed position with those of
its representative, and passed through the convolution and the recurrence; only the outputs at the
retained positions are then kept. Every other component, namely the full-attention layers, the MLPs
and the KV cache, processes only the pruned sequence. Replay is applied only during prefill;
decoding continues from the resulting GDN states without any change. The only overhead is the GDN
scan over the removed positions during prefill, which on Qwen3.5-9B adds less than $2\%$ to the
FLOPs of the vision encoder and the prefill of plain deletion.

\paragraph{Results.}
On grounding (Tables~\ref{tab:qwen35-grounding} and~\ref{tab:qwen38-grounding}), all baselines
nearly collapse on both models: the strongest reaches only $7.4$ on Qwen3.5-9B and $12.8$ on
Qwen3.8-27B, whereas CAR reaches $43.1$ and $41.0$, clearly outperforming all baselines. The
component ablations follow the same trend on both models. The probe-selected head raises WCS from
$6.6$ and $15.1$ to $20.3$ and $23.3$, GDN replay brings the largest gain, to $42.7$ and $40.4$, and
AR adds a small further gain, reaching $43.1$ and $41.0$. Replay also transfers to other selection
methods: on Qwen3.5-9B, combined with replay and AR, the five baselines rise from at most $7.4$ to
between $17.7$ and $43.2$, with EADP on par with CAR. On VQA (Tables~\ref{tab:qwen35-vqa}
and~\ref{tab:qwen38-vqa}), CAR outperforms the strongest baseline, EADP, by $5.0$ and $4.7$ points on
the two models. Most of the gain comes from head selection, while replay and AR bring little
further change.
}

{
\begin{table}[t]
  \centering
  
  \caption{\textbf{Visual grounding performance on Qwen3.5-9B, retaining 9.9\% of visual tokens.}}
  \label{tab:qwen35-grounding}
  \resizebox{\linewidth}{!}{
    \begin{tabular}{l|ccc|ccc|cc|cc}
    \toprule
    \multirow{2}{*}{\textbf{Method}}
    & \multicolumn{3}{c|}{\textbf{RefCOCO}} & \multicolumn{3}{c|}{\textbf{RefCOCO+}}
    & \multicolumn{2}{c|}{\textbf{RefCOCOg}} & \multirow{2}{*}{\textbf{Avg.}} & \multirow{2}{*}{\textbf{Rel.}} \\
    & testA & testB & val & testA & testB & val & test & val & & \\
    \rowcolor{lightgray!75}\multicolumn{11}{c}{\textit{Upper Bound, All Visual Tokens} ($\mathbf{100\%}$)} \\
    \rowcolor{lightgray!25} Qwen3.5-9B & 93.1 & 87.8 & 90.8 & 88.7 & 79.1 & 84.3 & 88.1 & 88.7 & 87.6 & 100.0\% \\
    \rowcolor{lightgray!75}\multicolumn{11}{c}{\textit{Pruning baselines}} \\
    DivPrune{\small\texttt{(CVPR25)}} & 9.6 & 4.8 & 6.7 & 7.5 & 3.7 & 5.3 & 5.7 & 6.1 & 6.2 & 7.0\% \\
    VisionZip{\small\texttt{(CVPR25)}} & 8.7 & 4.2 & 6.5 & 7.6 & 3.5 & 5.1 & 5.4 & 5.8 & 5.9 & 6.7\% \\
    AnchorPrune{\small\texttt{(ECCV26)}} & 10.0 & 5.5 & 7.6 & 7.0 & 3.7 & 5.3 & 5.9 & 5.7 & 6.3 & 7.2\% \\
    EADP{\small\texttt{(ECCV26)}} & 10.6 & 5.1 & 8.4 & 9.9 & 4.4 & 7.1 & 7.2 & 6.8 & 7.4 & 8.5\% \\
    N\"uwa{\small\texttt{(ICLR26)}} & 11.2 & 6.0 & 7.8 & 9.3 & 4.9 & 6.4 & 7.0 & 6.7 & 7.4 & 8.5\% \\
    \rowcolor{lightgray!75}\multicolumn{11}{c}{\textit{CAR components}} \\
    WCS (head avg.) & 9.7 & 5.5 & 7.0 & 8.5 & 4.2 & 6.0 & 5.9 & 5.7 & 6.6 & 7.5\% \\
    WCS (selected head) & 27.8 & 17.5 & 22.6 & 23.5 & 13.9 & 19.2 & 18.8 & 19.4 & 20.3 & 23.2\% \\
    + GDN replay & 49.6 & 42.7 & 46.9 & 41.5 & 34.5 & 38.6 & 43.5 & 44.4 & 42.7 & 48.8\% \\
    + GDN replay + AR (\textbf{CAR}) & 49.7 & 43.6 & 46.8 & 42.0 & 35.6 & 39.2 & 43.5 & 44.6 & 43.1 & 49.3\% \\
    \rowcolor{lightgray!75}\multicolumn{11}{c}{\textit{Baseline selection with GDN replay}} \\
    EADP + replay + AR & 49.1 & 43.8 & 46.8 & 42.2 & 35.2 & 40.1 & 43.9 & 44.6 & 43.2 & 49.3\% \\
    DivPrune + replay + AR & 35.4 & 29.4 & 33.7 & 28.6 & 23.2 & 27.6 & 29.5 & 30.1 & 29.7 & 33.9\% \\
    VisionZip + replay + AR & 33.1 & 31.1 & 32.7 & 27.8 & 25.2 & 26.1 & 30.0 & 30.8 & 29.6 & 33.8\% \\
    AnchorPrune + replay + AR & 21.6 & 19.5 & 21.0 & 16.6 & 13.9 & 15.4 & 17.0 & 16.8 & 17.7 & 20.2\% \\
    N\"uwa + replay + AR & 37.2 & 32.9 & 35.1 & 31.2 & 26.0 & 28.1 & 33.2 & 33.0 & 32.1 & 36.6\% \\
    \bottomrule
    \end{tabular}
  }
\end{table}
}

{
\begin{table}[t]
  \centering
  
  \caption{\textbf{Visual grounding performance on Qwen3.8-27B, retaining 9.9\% of visual tokens.}}
  \label{tab:qwen38-grounding}
  \resizebox{\linewidth}{!}{
    \begin{tabular}{l|ccc|ccc|cc|cc}
    \toprule
    \multirow{2}{*}{\textbf{Method}}
    & \multicolumn{3}{c|}{\textbf{RefCOCO}} & \multicolumn{3}{c|}{\textbf{RefCOCO+}}
    & \multicolumn{2}{c|}{\textbf{RefCOCOg}} & \multirow{2}{*}{\textbf{Avg.}} & \multirow{2}{*}{\textbf{Rel.}} \\
    & testA & testB & val & testA & testB & val & test & val & & \\
    \rowcolor{lightgray!75}\multicolumn{11}{c}{\textit{Upper Bound, All Visual Tokens} ($\mathbf{100\%}$)} \\
    \rowcolor{lightgray!25} Qwen3.8-27B & 90.9 & 86.7 & 89.9 & 86.3 & 77.7 & 83.6 & 88.8 & 88.7 & 86.6 & 100.0\% \\
    \rowcolor{lightgray!75}\multicolumn{11}{c}{\textit{Pruning baselines}} \\
    DivPrune{\small\texttt{(CVPR25)}} & 17.8 & 10.2 & 14.2 & 15.7 & 8.9 & 11.9 & 12.1 & 12.1 & 12.8 & 14.8\% \\
    VisionZip{\small\texttt{(CVPR25)}} & 8.6 & 4.7 & 6.4 & 7.4 & 3.7 & 5.0 & 5.4 & 4.8 & 5.7 & 6.6\% \\
    AnchorPrune{\small\texttt{(ECCV26)}} & 16.4 & 10.9 & 13.3 & 13.8 & 7.9 & 11.3 & 11.9 & 12.2 & 12.2 & 14.1\% \\
    EADP{\small\texttt{(ECCV26)}} & 18.0 & 10.1 & 11.1 & 15.4 & 8.0 & 11.8 & 11.9 & 11.3 & 12.2 & 14.1\% \\
    N\"uwa{\small\texttt{(ICLR26)}} & 12.2 & 7.0 & 8.7 & 9.6 & 5.6 & 6.9 & 7.9 & 7.3 & 8.2 & 9.4\% \\
    \rowcolor{lightgray!75}\multicolumn{11}{c}{\textit{CAR components}} \\
    WCS (head avg.) & 20.5 & 12.2 & 16.9 & 17.8 & 10.5 & 15.1 & 13.8 & 14.2 & 15.1 & 17.5\% \\
    WCS (selected head) & 30.6 & 21.2 & 25.3 & 26.9 & 17.4 & 22.7 & 20.9 & 21.2 & 23.3 & 26.9\% \\
    + GDN replay & 47.9 & 39.7 & 44.8 & 41.4 & 32.4 & 37.3 & 40.4 & 39.3 & 40.4 & 46.7\% \\
    + GDN replay + AR (\textbf{CAR}) & 48.0 & 40.6 & 45.9 & 41.8 & 32.7 & 38.3 & 40.6 & 40.3 & 41.0 & 47.4\% \\
    \bottomrule
    \end{tabular}
  }
\end{table}
}

{
\begin{table}[t]
  \centering
  
  \caption{\textbf{VQA performance on Qwen3.5-9B, retaining 11.1\% of visual tokens.}}
  \label{tab:qwen35-vqa}
  \resizebox{\linewidth}{!}{
    \begin{tabular}{l|cccccccc|cc}
    \toprule
    \textbf{Method} & \textbf{TextVQA} & \textbf{ChartQA} & \textbf{AI2D} & \textbf{OCRBench} & \textbf{HallBench} & \textbf{MME} & \textbf{MMB-EN} & \textbf{MMB-CN} & \textbf{Acc.} & \textbf{Rel.} \\
    \rowcolor{lightgray!75}\multicolumn{11}{c}{\textit{Upper Bound, All Visual Tokens} ($\mathbf{100\%}$)} \\
    \rowcolor{lightgray!25} Qwen3.5-9B & 84.6 & 87.4 & 88.0 & 889 & 57.2 & 2365.4 & 85.8 & 85.8 & 87.0 & 100.0\% \\
    \rowcolor{lightgray!75}\multicolumn{11}{c}{\textit{Pruning baselines}} \\
    DivPrune{\small\texttt{(CVPR25)}} & 68.5 & 34.4 & 74.1 & 411 & 43.3 & 1981.2 & 75.9 & 74.1 & 63.8 & 73.1\% \\
    VisionZip{\small\texttt{(CVPR25)}} & 52.7 & 40.2 & 71.2 & 202 & 41.2 & 1715.7 & 72.6 & 71.3 & 56.9 & 65.5\% \\
    AnchorPrune{\small\texttt{(ECCV26)}} & 57.2 & 27.9 & 70.5 & 278 & 44.3 & 1785.8 & 68.5 & 65.6 & 56.4 & 65.0\% \\
    EADP{\small\texttt{(ECCV26)}} & 64.1 & 39.7 & 74.2 & 402 & 42.9 & 2059.8 & 78.1 & 75.7 & 64.7 & 74.0\% \\
    N\"uwa{\small\texttt{(ICLR26)}} & 60.3 & 52.4 & 74.9 & 233 & 45.4 & 1994.7 & 77.6 & 76.4 & 63.7 & 73.2\% \\
    \rowcolor{lightgray!75}\multicolumn{11}{c}{\textit{CAR components}} \\
    WCS (head avg.) & 66.8 & 41.4 & 74.3 & 382 & 42.9 & 2007.2 & 77.6 & 76.3 & 64.7 & 74.1\% \\
    WCS (selected head) & 69.0 & 48.7 & 77.0 & 448 & 46.1 & 2120.0 & 79.0 & 78.6 & 68.6 & 78.6\% \\
    + GDN replay & 70.1 & 46.9 & 76.6 & 521 & 45.3 & 2111.2 & 78.9 & 76.3 & 69.0 & 78.9\% \\
    + GDN replay + AR (\textbf{CAR}) & 70.3 & 47.0 & 76.9 & 513 & 46.9 & 2144.9 & 79.7 & 77.9 & 69.7 & 79.8\% \\
    \bottomrule
    \end{tabular}
  }
\end{table}
}

{
\begin{table}[t]
  \centering
  
  \caption{\textbf{VQA performance on Qwen3.8-27B, retaining 11.1\% of visual tokens.}}
  \label{tab:qwen38-vqa}
  \resizebox{\linewidth}{!}{
    \begin{tabular}{l|cccccccc|cc}
    \toprule
    \textbf{Method} & \textbf{TextVQA} & \textbf{ChartQA} & \textbf{AI2D} & \textbf{OCRBench} & \textbf{HallBench} & \textbf{MME} & \textbf{MMB-EN} & \textbf{MMB-CN} & \textbf{Acc.} & \textbf{Rel.} \\
    \rowcolor{lightgray!75}\multicolumn{11}{c}{\textit{Upper Bound, All Visual Tokens} ($\mathbf{100\%}$)} \\
    \rowcolor{lightgray!25} Qwen3.8-27B & 83.0 & 88.2 & 88.2 & 856 & 57.7 & 2179.9 & 83.7 & 81.0 & 84.6 & 100.0\% \\
    \rowcolor{lightgray!75}\multicolumn{11}{c}{\textit{Pruning baselines}} \\
    DivPrune{\small\texttt{(CVPR25)}} & 62.8 & 32.8 & 72.0 & 358 & 36.0 & 1720.1 & 70.2 & 67.0 & 57.8 & 68.0\% \\
    VisionZip{\small\texttt{(CVPR25)}} & 47.1 & 36.4 & 69.4 & 165 & 36.7 & 1546.2 & 68.3 & 65.2 & 52.1 & 61.6\% \\
    AnchorPrune{\small\texttt{(ECCV26)}} & 55.2 & 34.4 & 72.8 & 267 & 38.6 & 1619.2 & 68.7 & 64.7 & 55.3 & 65.3\% \\
    EADP{\small\texttt{(ECCV26)}} & 64.1 & 34.9 & 71.9 & 359 & 39.2 & 1792.7 & 73.5 & 70.5 & 60.0 & 70.7\% \\
    N\"uwa{\small\texttt{(ICLR26)}} & 54.2 & 41.6 & 70.1 & 223 & 39.0 & 1550.7 & 66.1 & 63.4 & 54.3 & 64.3\% \\
    \rowcolor{lightgray!75}\multicolumn{11}{c}{\textit{CAR components}} \\
    WCS (head avg.) & 64.0 & 39.0 & 73.3 & 324 & 40.4 & 1737.7 & 74.1 & 70.5 & 60.1 & 70.9\% \\
    WCS (selected head) & 67.5 & 59.6 & 76.1 & 405 & 42.5 & 1887.0 & 73.7 & 68.5 & 65.3 & 76.9\% \\
    + GDN replay & 67.8 & 54.6 & 74.1 & 457 & 39.1 & 1927.1 & 74.0 & 69.7 & 65.2 & 76.4\% \\
    + GDN replay + AR (\textbf{CAR}) & 67.9 & 52.4 & 73.5 & 444 & 41.6 & 1880.5 & 74.5 & 69.4 & 64.7 & 76.2\% \\
    \bottomrule
    \end{tabular}
  }
\end{table}
}

\subsection{A Single Configuration for Grounding and Question Answering}
\label{sec:appendix:unified}
{
The deployed configurations of CAR are task-specific and assume that the task family is known at
inference. They differ in three components: grounding uses the spatial kernel, the uniform bias and the
saliency of the head that the probe selects on grounding, whereas VQA uses no spatial kernel, the
capped bias and the saliency of the head selected on VQA (Appendix~\ref{sec:appendix:probe}).
On Qwen2.5-VL-7B, we evaluate one configuration shared by grounding and question answering, and
test it on MM-GCoT~\citep{wu2026grounded}, whose queries ask both what an object is and where it is. This
unified configuration fixes the spatial kernel and the capped bias: the kernel gains $4.3$ on VG
and costs $0.2$ on VQA (Table~\ref{tab:abl-comp}), and the capped bias has the smallest loss across
the two tasks (Table~\ref{tab:abl-bias}). Its saliency averages the saliency of the two heads that
the label-free head probe selects for VQA (head $2$) and VG (head $13$). The probe runs once per
backbone on the probe set, so no task label is needed at inference.

{
\begin{table}[t]
  \centering
  
  \caption{\textbf{Single vs.\ task-specific configurations on VQA with Qwen2.5-VL-7B} ($128$
  tokens retained). The unified configuration uses the spatial kernel, the capped bias and the
  saliency averaged over heads $2$ and $13$.}
  \label{tab:unified-config-vqa}
  \resizebox{\linewidth}{!}{
    \begin{tabular}{l|cccccccc|c}
    \toprule
    \textbf{Method} & \textbf{TextVQA} & \textbf{ChartQA} & \textbf{AI2D} & \textbf{OCRBench} & \textbf{HallBench} & \textbf{MME} & \textbf{MMB-EN} & \textbf{MMB-CN} & \textbf{Acc.} \\
    \midrule
    \rowcolor{lightgray!25} Qwen2.5-VL-7B & 84.3 & 87.1 & 81.6 & 867 & 56.2 & 2342.3 & 83.1 & 82.9 & 84.9 \\
    HAWK{\small\texttt{(CVPR26)}} & 79.9 & 71.3 & 77.5 & 705 & 48.8 & 2225.6 & 79.0 & 78.6 & 77.1 \\
    \textbf{CAR}, task-specific & 78.8 & 76.1 & 79.4 & 731 & 49.0 & 2314.3 & 81.1 & 81.3 & 79.3 \\
    \textbf{CAR}, unified & 76.2 & 75.6 & 79.6 & 695 & 48.5 & 2262.0 & 80.3 & 80.9 & 78.0 \\
    \bottomrule
    \end{tabular}
  }

  \vspace{2mm}
  \caption{\textbf{Single vs.\ task-specific configurations on visual grounding with Qwen2.5-VL-7B}
  ($9.9\%$ of the visual tokens retained). The unified configuration is the same as in
  Table~\ref{tab:unified-config-vqa}.}
  \label{tab:unified-config-vg}
  \resizebox{\linewidth}{!}{
    \begin{tabular}{l|ccc|ccc|cc|c}
    \toprule
    \multirow{2}{*}{\textbf{Method}}
    & \multicolumn{3}{c|}{\textbf{RefCOCO}} & \multicolumn{3}{c|}{\textbf{RefCOCO+}}
    & \multicolumn{2}{c|}{\textbf{RefCOCOg}} & \multirow{2}{*}{\textbf{Avg.}} \\
    & testA & testB & val & testA & testB & val & test & val & \\
    \midrule
    \rowcolor{lightgray!25} Qwen2.5-VL-7B & 92.2 & 84.8 & 89.4 & 87.6 & 74.1 & 82.8 & 86.7 & 86.7 & 85.5 \\
    N\"uwa{\small\texttt{(ICLR26)}} & 27.5 & 23.7 & 26.2 & 21.8 & 17.4 & 20.5 & 23.0 & 22.0 & 22.8 \\
    \textbf{CAR}, task-specific & 75.4 & 66.7 & 72.2 & 67.2 & 55.9 & 62.3 & 67.8 & 66.5 & 66.8 \\
    \textbf{CAR}, unified & 71.9 & 62.7 & 69.4 & 64.3 & 52.2 & 59.7 & 65.6 & 64.3 & 63.8 \\
    \bottomrule
    \end{tabular}
  }
\end{table}
}

\paragraph{Grounding and question answering.}
Against the task-specific configurations, the unified configuration loses $3.0$ points on VG and
$1.3$ on VQA (Tables~\ref{tab:unified-config-vqa} and~\ref{tab:unified-config-vg}). It still clearly outperforms the best VG baseline
($63.8$ vs.\ $22.8$) and remains above the best VQA baseline ($78.0$ vs.\ $77.1$).

{
\begin{table}[t]
  \centering
  
  \caption{\textbf{MM-GCoT on Qwen2.5-VL-7B} ($9.9\%$ of the visual tokens retained). A-Acc: answer
  accuracy; G-Acc: box IoU $>0.5$; Cons.: answer-box consistency. VQA/VG config.: the task-specific configurations.}
  \label{tab:mmgcot}
  \resizebox{0.85\linewidth}{!}{
    \begin{tabular}{l|ccc|ccc}
    \toprule
    \multirow{2}{*}{\textbf{Method}} & \multicolumn{3}{c|}{\textbf{Answer, then grounding}} & \multicolumn{3}{c}{\textbf{Single query}} \\
    & A-Acc & G-Acc & Cons. & A-Acc & G-Acc & Cons. \\
    \midrule
    \rowcolor{lightgray!25} Qwen2.5-VL-7B & 70.9 & 63.4 & 56.1 & 66.1 & 70.3 & 56.9 \\
    VisionZip{\small\texttt{(CVPR25)}} & 58.8 & 17.8 & 18.9 & 54.5 & 19.1 & 18.7 \\
    DivPrune{\small\texttt{(CVPR25)}} & 63.6 & 19.9 & 20.0 & 58.8 & 21.2 & 20.9 \\
    AnchorPrune{\small\texttt{(ECCV26)}} & 61.0 & 17.6 & 19.0 & 56.1 & 20.3 & 18.3 \\
    EADP{\small\texttt{(ECCV26)}} & 58.5 & 16.4 & 17.6 & 52.0 & 18.5 & 16.9 \\
    N\"uwa{\small\texttt{(ICLR26)}} & 61.2 & 18.0 & 17.9 & 55.3 & 17.6 & 16.6 \\
    \midrule
    \textbf{CAR}, VQA config. & 63.8 & 25.6 & 24.0 & 61.1 & 30.9 & 27.3 \\
    \textbf{CAR}, VG config. & 64.3 & 41.0 & 38.1 & 55.4 & 47.8 & 38.1 \\
    \textbf{CAR}, unified & 62.1 & 38.1 & 37.3 & 58.7 & 43.8 & 37.4 \\
    \textbf{CAR}, unified, MM-GCoT probe & 64.5 & 42.3 & 39.7 & 59.0 & 48.4 & 39.6 \\
    \bottomrule
    \end{tabular}
  }
\end{table}
}

\paragraph{A query that asks both what and where.}
In MM-GCoT, each question asks for the attribute or category of an object and for its box. We
evaluate it in two settings (Table~\ref{tab:mmgcot}): the official answer-then-grounding protocol, which asks for the
answer and then the box, and a single query asking for both. The unified configuration uses the
saliency averaged over the heads selected by the VQA and VG probes, without any MM-GCoT data. It
substantially outperforms the best baseline in localization (G-Acc $38.1$ vs.\ $19.9$ under answer-then-grounding and $43.8$ vs.\ $21.2$ with a single query) and in answer-box consistency ($37.3$
vs.\ $20.0$ and $37.4$ vs.\ $20.9$), with comparable answer accuracy.

We further run the head probe on $500$ unlabeled questions from the MM-GCoT training set and
compute the saliency from the head it selects. This variant gives the best pruned
results on all three metrics under answer-then-grounding ($64.5$ A-Acc, $42.3$ G-Acc, $39.7$
Cons.) and the best G-Acc and consistency with a single query. It improves G-Acc over the unified
configuration by $4.2$ and $4.6$ points and also exceeds the VG configuration in both settings.

\paragraph{Guideline.}
(i) When the task family is known, use its task-specific configuration. (ii) When it is unknown or
the query is mixed, use the unified configuration. (iii) For a new task family with unlabeled
inputs, a one-time probe on these inputs can further improve accuracy.
}

\subsection{Total Output Error and Output KL}
\label{sec:appendix:total-kl}
\begin{figure}[h]
  \centering
  \includegraphics[width=\linewidth]{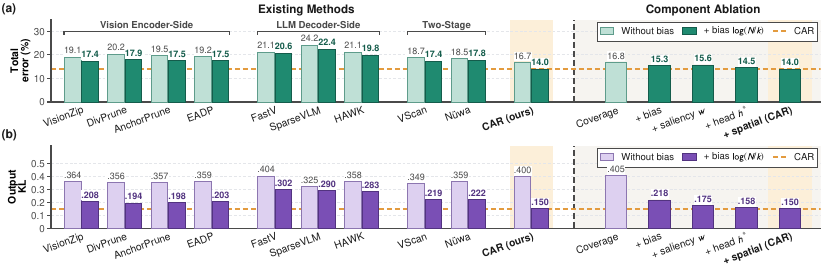}
  
  \caption{\textbf{Total output error and output KL on grounding} for existing methods and the
  ablation of Table~\ref{tab:abl-comp} (Qwen2.5-VL-7B, $9.9\%$). (a) Total error
  $\sum\lVert O'-O\rVert^{2}/\sum\lVert O\rVert^{2}$ (\%) over layers $13$--$19$. (b) KL divergence
  between the output distributions of the unpruned and pruned models. Lower is better.}
  \label{fig:total-kl}
\end{figure}

{
For every configuration of Figure~\ref{fig:error-ablation} (Qwen2.5-VL-7B, grounding, $9.9\%$
retention), we report the total error
$\sum\lVert O'-O\rVert^2/\sum\lVert O\rVert^2$ (\%) over layers $13$--$19$, where
$\lVert O'-O\rVert^2=\lVert\mathcal{E}_{\mathrm{sum}}\rVert^2+\lVert\mathcal{E}_{\mathrm{alloc}}\rVert^2+2\langle\mathcal{E}_{\mathrm{sum}},\mathcal{E}_{\mathrm{alloc}}\rangle$
is the sum of the image summary error, the attention allocation error and their cross term. We also
report the KL divergence between the output distributions of the unpruned and pruned models over the vocabulary
(Figure~\ref{fig:total-kl}).

\paragraph{Total Error.}
The bias lowers the total error of every method. CAR has the lowest total error both without the
bias ($16.7$, against $18.5$--$24.2$ for the existing methods) and with it ($14.0$, against
$17.4$--$22.4$). In the component ablation, the total error decreases from $16.8$ for coverage
selection alone to $14.0$ for CAR, mainly through the bias ($-1.5$) and the selected head ($-1.1$).

\paragraph{Output KL.}
The bias also lowers the KL of every method. With the bias, CAR has the lowest KL ($0.150$, against
$0.194$--$0.302$ for the existing methods). In the component ablation, each component lowers the KL
from $0.405$ for coverage selection alone: the bias to $0.218$, the saliency weights to $0.175$, the
selected head to $0.158$, and the spatial kernel to $0.150$ (CAR). Thus, CAR lowers not only the two
terms of Figure~\ref{fig:error-ablation} but also their total and the shift in the final output.
}

\subsection{Consistency between the Theoretical and Practical Optimization Objectives}
\label{sec:appendix:enc-dec}
{
Proposition~\ref{prop:bound} bounds the image summary error with the similarity and attention of
the decoder, whereas CAR selects tokens with those of the encoder. We relate the two on
Qwen2.5-VL-7B, measuring the decoder quantities in the unpruned model on $300$ grounding images from
RefCOCO/+/g val and $300$ VQA images from TextVQA and ChartQA.

\paragraph{Similarity.}
In layers $13$--$19$, where the text attends most to the image, the Spearman correlation between
the encoder cosine similarity of token pairs and their decoder similarity is
$0.54$/$0.63$/$0.68$ for values/keys/hidden states on grounding and $0.38$/$0.52$/$0.53$ on VQA.
Tokens similar in the encoder thus largely remain similar in the decoder.

\paragraph{Saliency.}
The head-averaged encoder saliency correlates moderately with the decoder attention in the shallow
layers (Spearman $0.33$--$0.37$ on grounding, $0.44$ on VQA) and less in deeper layers, where
attention depends more on the query. CAR does not need the saliency to match the decoder attention
token by token, since the head probe (Section~\ref{sec:method:head}) selects the head by the
agreement of pruned and unpruned outputs. The selected head focuses on task-relevant content: on
RefCOCO/+/g, a token on the referred object receives on average $1.30\times$ the saliency of the
other tokens.

\paragraph{Coverage Objective and Summary Error.}
We replace $10$--$75\%$ of the tokens selected by CAR with random discarded tokens and measure the
encoder objective $F$ of Eq.~\ref{eq:fl-ours} and the image summary error in layers $13$--$19$
(grounding, $9.9\%$ retention; Table~\ref{tab:app-cov-align}). As $F$ decreases, the summary error
increases monotonically. The Spearman correlation between them, computed within each image, has a
median of $-0.52$. A higher coverage objective thus corresponds to a lower image summary error, the
quantity that the bound controls.
}

\begin{table}[h]
  \centering
  \small
  
  \caption{\textbf{Coverage objective $F$ and image summary error $\mathcal{E}_{\mathrm{sum}}$} as the tokens of CAR are
  replaced with random discarded ones (Qwen2.5-VL-7B, grounding, $9.9\%$).}
  \label{tab:app-cov-align}
  {
  \setlength{\tabcolsep}{3.5pt}
  \begin{tabular}{l|cccccc}
    \toprule
    Retained set & CAR & 10\% replaced & 25\% replaced & 50\% replaced & 75\% replaced & Random \\
    \midrule
    $F$ & $0.673$ & $0.653$ & $0.625$ & $0.576$ & $0.538$ & $0.509$ \\
    Image summary error (\%) & $10.28$ & $10.62$ & $11.14$ & $12.17$ & $12.71$ & $13.57$ \\
    \bottomrule
  \end{tabular}}
\end{table}

\section{Efficiency Analysis}
\label{sec:appendix:efficiency}

\subsection{Compatibility with FlashAttention}
\label{sec:appendix:flashattn}

To maintain compatibility with optimized attention kernels such as FlashAttention
\citep{dao2024flashattention2}, which never materialize the attention matrix, we propose
Post-Softmax Recombination (PSR) to apply the logit bias of Eq.~\ref{eq:pruned} without
modifying the kernel. PSR builds on a standard practice of attention kernels: besides the
attention output, the kernel returns the log-sum-exp of the logits, which suffices to
add further terms to the softmax outside the kernel
\citep{dao2024flashattention2,ye2025flashinfer}. PSR therefore runs the unmodified kernel
and adds the effect of the bias in such a recombination.

\begin{figure}[t]
\centering
\includegraphics[width=\linewidth]{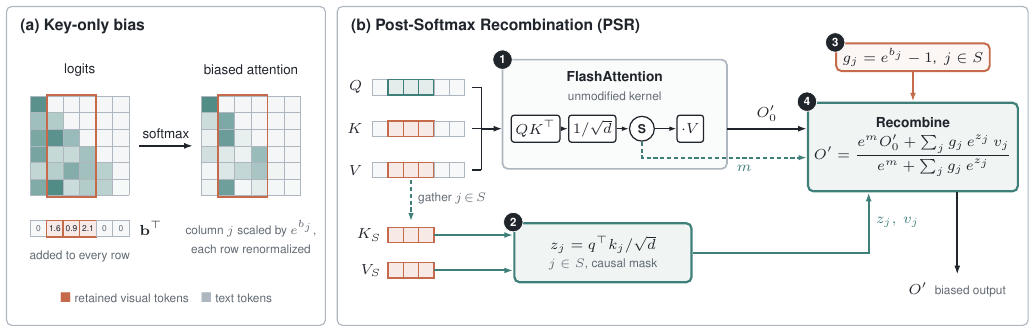}
\caption{(a) A key-only bias rescales the retained columns of the attention.
(b) The four steps of PSR around an unmodified FlashAttention.}
\label{fig:psr}
\end{figure}

This recombination applies because our bias is key-only: $b_j$ depends on the retained
token $j\in S$ but not on the query, and text
tokens receive $b_t=0$. As shown in Figure~\ref{fig:psr}(a), the bias thus scales each
retained column of the attention by $e^{b_j}$ and renormalizes each row, which only adds
extra terms for the $k$ retained tokens to the softmax.

Figure~\ref{fig:psr}(b) shows the four steps of PSR for a query $q$ with logits $z_u$.
In step~1, we call the unmodified FlashAttention, which returns the unbiased output
$O'_0$, that is, Eq.~\ref{eq:pruned} with $b=0$, together with the log-sum-exp $m$ of the
logits:
\begin{equation}
O'_0,\;m=\mathrm{FlashAttn}(Q,K,V),
\qquad
m=\log\sum_{u\in T\cup S}e^{z_u} ,
\label{eq:psr-kernel}
\end{equation}
so that $e^{m}$ and $e^{m}O'_0$ are the denominator and numerator of the unbiased softmax.
In step~2, we gather the keys $K_S$ and values $V_S$ of the retained tokens and compute
their logits $z_j=q^{\top}k_j/\sqrt{d_h}$ under the kernel's causal mask. In step~3, we
write $e^{b_j}=1+g_j$ for $j\in S$, so that the bias adds only the extra weight
$g_j e^{z_j}$ of each retained token to the softmax. In step~4, we recombine these
quantities into the biased output:
\begin{equation}
O'=\frac{e^{m}\,O'_0+\sum_{j\in S}g_j\,e^{z_j}\,v_j}{e^{m}+\sum_{j\in S}g_j\,e^{z_j}} .
\label{eq:psr}
\end{equation}
Since the extra terms involve only the retained tokens, PSR adds computation only on
these $k$ columns.

\subsection{Complexity Analysis}
\label{sec:appendix:eff-complexity}

Our method adds computation in two places. (1) The first is the token selection that runs after the
vision encoder and before the language model. It computes the similarity matrix with a matrix
multiplication of complexity $O(N^{2}d)$, which dominates this stage. The greedy pass performs $k$
steps and, since the gain of a candidate is a sum over all the visual tokens, every step rescans
the matrix, $O(kN^{2})$. The saliency requires one explicit attention of the head $h^{*}$ over the
patches, $O(P^{2}d_h)$. For the per-cell bias, the values of the bias require the assignment of the
discarded tokens to the retained ones, $O((N-k)kd)$. (2) The second is the prefill, where the bias
is added to the logits at every decoder layer: the recombination of Eq.~\ref{eq:psr} acts on the
retained columns of each layer, $O(Lnkd)$.

However, this overhead is negligible compared to the total computational cost of the model. As
shown in Table~\ref{tab:complexity}, these steps require $24.2$ GFLOPs for Qwen2.5-VL-7B, only
$0.6\%$ of the $3.8$ TFLOPs the pruned model spends, and the decode time is unchanged, $27.3$
against $27.1$\,ms in Table~\ref{tab:efficiency}. The greedy pass is a sequence of $k$ steps, so the
selection takes $13.9$\,ms against $1.6$\,ms for top-$k$ selection by the saliency, which is why
our prefill is longer than that of the methods built on it.

We further isolate the cost of the bias itself. The two configurations of
Table~\ref{tab:bias-overhead} use the same selection and retain the same tokens, so they differ
only in whether the bias is applied. Applying it adds $5.0$\,ms to the prefill and $5.6$\,ms to
the end-to-end latency, and leaves the peak memory unchanged. The decode time differs by
$0.1$\,ms per token, less than the variation between repeated runs, because the bias enters the
prefill only and the decoding steps run the unmodified kernel. Of the $5.0$\,ms, the assignment
of the discarded tokens takes $0.5$\,ms and the recombination of Eq.~\ref{eq:psr} across the
$28$ layers takes the remaining $4.5$\,ms. The bias therefore adds $1.9\%$ to the end-to-end
latency, all of it in the prefill.

\begin{table}[t]
  \centering
  \small
  \caption{\textbf{Cost of the bias in isolation} (Qwen2.5-VL-7B, $128$ of $1{,}296$ visual
  tokens). The two configurations retain the same tokens and differ only in the bias.}
  \label{tab:bias-overhead}
  \begin{tabular}{l|cc|c}
  \toprule
  & \textbf{w/o bias} & \textbf{+ bias (ours)} & \textbf{Difference} \\
  \midrule
  Prefill (ms) & 241.6 & 246.6 & +5.0 \\
  Decode (ms/token) & 27.2 & 27.3 & +0.1 \\
  Latency (ms) & 299.1 & 304.7 & +5.6 \\
  \midrule
  Peak memory (GB) & 15.82 & 15.82 & ${<}0.01$ \\
  \bottomrule
  \end{tabular}
  \vspace{-2mm}
\end{table}

We measure the efficiency of Table~\ref{tab:efficiency} on the first $200$ questions of TextVQA
using one RTX $3090$ with batch size $1$ in \texttt{bfloat16}. The language model and the vision
encoder run FlashAttention-2 for every method, and our bias is injected through PSR
(Appendix~\ref{sec:appendix:flashattn}). Its FLOPs column is theoretical, obtained by applying
$4nd^{2}+2n^{2}d+2ndm$ to the token count measured at each layer of the vision encoder and the
language model. {The prefill time and end-to-end latency cover the whole forward pass and
generation call, including token selection, bias preparation and the associated tensor transfers;
only CPU preprocessing and the initial CPU-to-GPU input transfer are excluded, for all methods
alike. The FLOPs count the vision encoder and the language model for all methods alike and exclude
the selection step of every method; Table~\ref{tab:complexity} gives the cost of the additional
steps of CAR.}

\begin{table}[t]
  \centering
  \small
  \caption{\textbf{Cost of the steps our method adds} (Qwen2.5-VL-7B, $128$ of $1{,}296$ visual
  tokens). $N$ visual tokens, $k$ retained, $d$ the hidden size, $P=4N$ patches before the merger,
  $d_h$ the encoder head dimension, $n$ the prompt length, $L$ the decoder layers.}
  \label{tab:complexity}
  \begin{tabular}{l|cc}
  \toprule
  \textbf{Step} & \textbf{Complexity} & \textbf{FLOPs} \\
  \midrule
  Saliency from the probed head & $O(P^{2}d_h)$ & $4.3$\,G \\
  Similarity matrix & $O(N^{2}d)$ & $12.0$\,G \\
  Spatial kernel (grounding only) & $O(N^{2})$ & ${<}0.1$\,G \\
  Greedy coverage, $k$ steps & $O(kN^{2})$ & $0.6$\,G \\
  Assignment for the bias & $O((N-k)kd)$ & $1.1$\,G \\
  Bias in prefill, $L$ layers & $O(Lnkd)$ & $6.2$\,G \\
  \midrule
  \textbf{Total} & & $\mathbf{24.2}$\,\textbf{G} \\
  \bottomrule
  \end{tabular}
  \vspace{-2mm}
\end{table}

\paragraph{Selection Cost at Large $N$.}
The greedy pass rescans the $N\times N$ similarity matrix at every step, so it costs $O(kN^{2})$,
which becomes noticeable when $N$ is large, as for Qwen3-VL-8B selecting jointly over $32$ video
frames ($N=11{,}520$) or InternVL3.5-8B with up to $24$ tiles ($N=6{,}144$). We change only how the
gains are computed. When a token is selected, only the few tokens that are similar to it and not
yet well covered, about a dozen at the median in our measurements, gain coverage, so each step
updates the gains by their contribution alone instead of rescanning the matrix. This lowers the
selection time on video from $5.07$\,s to $0.61$\,s and the extra peak memory from $1.52$ to
$1.19$\,GB, and on InternVL3.5-8B with $24$ tiles from $814$ to $279$\,ms. For $N<3{,}000$ the
fixed cost of each incremental step dominates, so there we recompute all gains at every step.

{
\paragraph{End-to-End Latency across Input Sizes.}
Following the protocol of Table~\ref{tab:efficiency}, we evaluate the end-to-end latency of CAR
across input sizes and retention ratios (Table~\ref{tab:latency-scaling}). Each model uses the two
retention ratios at which its accuracy is reported, and for $N\geq3{,}000$ selection uses the
incremental implementation above. CAR is faster than the unpruned model in every setting, by
$1.71$--$2.18\times$ at the low ratio and by $1.30$--$1.57\times$ at the high ratio, with all
selection overhead included.
}

{
\begin{table}[t]
  \centering
  
  \caption{End-to-end speedup of CAR over the unpruned model across input sizes and retention
  ratios ($N$: number of visual tokens). Selection overhead is included.}
  \label{tab:latency-scaling}
  \resizebox{\linewidth}{!}{
    \begin{tabular}{l|c|r|r|cc}
    \toprule
    \textbf{Model (benchmark)} & \textbf{Ratio (low / high)} & \textbf{$N$} & \textbf{Unpruned (ms)}
    & \textbf{Speedup, low ratio} & \textbf{Speedup, high ratio} \\
    \midrule
    Qwen2.5-VL-7B (TextVQA) & $9.9\%$ / $19.8\%$ & $1{,}296$ & $553$ & $1.73\times$ & $1.57\times$ \\
    Qwen2.5-VL-7B (TextVQA) & $9.9\%$ / $19.8\%$ & $2{,}916$ & $1{,}162$ & $1.76\times$ & $1.42\times$ \\
    Qwen2.5-VL-7B (TextVQA) & $9.9\%$ / $19.8\%$ & $3{,}136$ & $1{,}246$ & $1.71\times$ & $1.34\times$ \\
    Qwen2.5-VL-7B (TextVQA) & $9.9\%$ / $19.8\%$ & $4{,}096$ & $1{,}608$ & $1.73\times$ & $1.34\times$ \\
    InternVL3.5-8B (OCRBench, $24$ tiles) & $11.1\%$ / $22.2\%$ & $6{,}144$ & $2{,}139$ & $2.18\times$ & $1.47\times$ \\
    Qwen3-VL-8B (Video-MME, $32$ frames) & $9.9\%$ / $25\%$ & $11{,}520$ & $4{,}607$ & $1.95\times$ & $1.30\times$ \\
    \bottomrule
    \end{tabular}
  }
\end{table}
}

\section{Qualitative Results}
\label{sec:appendix:qualitative}

In this section we supplement the results with four groups of visualizations: the tokens different
selection rules retain, the effect of the attention bias, the saliency read from different encoder
layers of InternVL3.5-8B, and failure cases of the deployed configuration.

\paragraph{Retained Tokens and Cells.}
\label{sec:appendix:qual-cells}

\begin{figure}[t]
  \centering
  \includegraphics[width=0.9\linewidth]{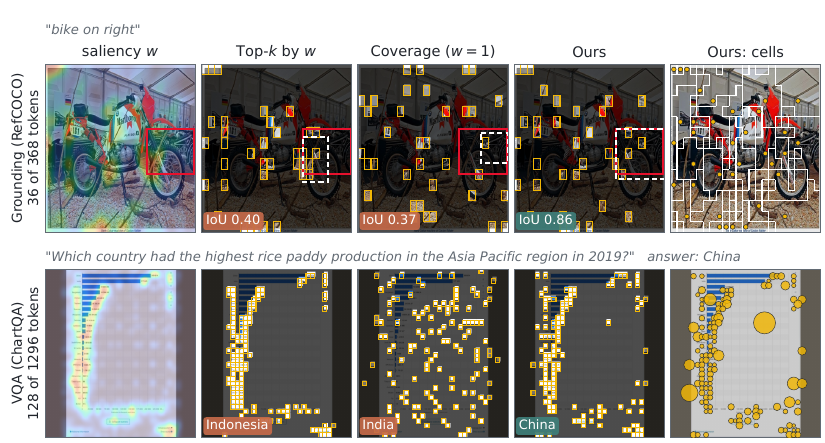}
  \caption{\textbf{Retained tokens under three selection rules.} Qwen2.5-VL-7B on RefCOCO at
  $9.9\%$ of the tokens and on ChartQA at $128$ tokens. Discarded tokens are darkened and retained ones outlined; red is the ground truth,
  dashed the prediction, and the label gives the IoU or the answer. The last column shows which
  retained token represents each discarded one: cell boundaries on grounding, and on VQA each
  representative with an area proportional to the tokens it represents.}
  \label{fig:qual-cells}
\end{figure}

Figure~\ref{fig:qual-cells} compares three selection rules: ranking tokens by $w$ (the
importance-ranking rule of Table~\ref{tab:abl-select}), facility location with $w=1$, and our
saliency-weighted facility location. Ranking by $w$ concentrates the retained tokens on
the regions of highest saliency, where the tokens a task requires may be left unrepresented, while
facility location with $w=1$ distributes them uniformly over the image and disregards the
saliency. Our rule covers the image and at the same time allocates more tokens to the salient
regions, and it answers both questions correctly. The last column shows the assignment of each
discarded token to its representative, which induces a partition of the image into cells. Every
region of the image is thereby represented, and the cells are larger in the uniform regions and
smaller where the saliency concentrates. The number of tokens each retained token represents
is the count $|C_j|$ on which the per-cell bias of Section~\ref{sec:method:bias} is defined.

\paragraph{Effect of the Attention Bias.}
\label{sec:appendix:qual-bias}

\begin{figure}[t]
  \centering
  \includegraphics[width=\linewidth]{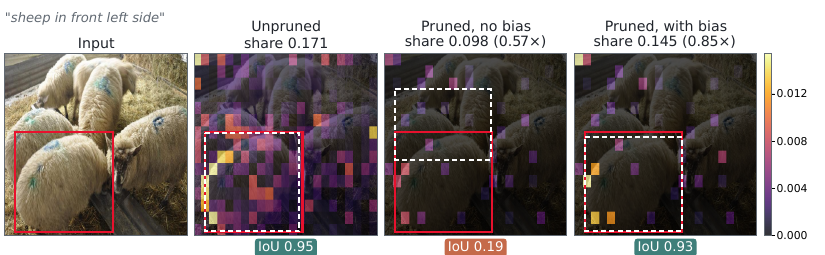}
  \caption{\textbf{Effect of the attention bias under a fixed retained set} (Qwen2.5-VL-7B,
  RefCOCO at $9.9\%$). Color encodes the attention a visual token receives from the text, averaged
  over heads, text queries and decoder layers $13$ to $19$. Discarded tokens are darkened, the
  ground truth is in red and the prediction is dashed. \emph{Share} denotes the fraction of the
  text attention falling on the image, with its ratio to the unpruned model in brackets.}
  \label{fig:qual-bias}
\end{figure}

With the retained set held fixed, Figure~\ref{fig:qual-bias} compares the attention that each
retained token receives in the pruned model without and with the bias. Without bias, the retained
tokens receive $0.57$ of the share they receive in the unpruned model, and the predicted box
covers the wrong sheep. The bias restores the share to $0.85$, the retained tokens on the referred
sheep are attended again, and the predicted box returns to the target.

\paragraph{Saliency Layers on InternVL3.5-8B.}
\label{sec:appendix:qual-layers}

\begin{figure}[t]
  \centering
  \includegraphics[width=0.9\linewidth]{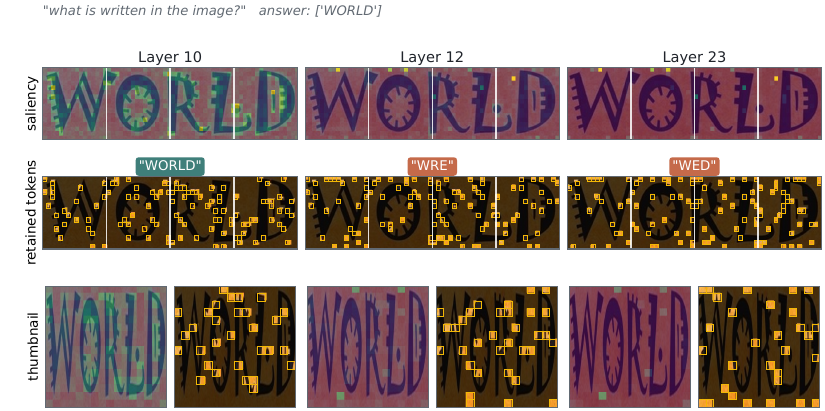}
  \caption{\textbf{Saliency from three encoder layers of InternVL3.5-8B and the tokens it
  retains.} OCRBench, $142$ of $1{,}280$ tokens kept; the image is cut into $4{\times}1$ tiles
  (white lines) plus a thumbnail tile, shown in the bottom row as saliency and retained tokens.
  The label above the second row is the answer. The attention sinks grow with depth: for this image, the class token
  receives $3\%$, $19\%$ and $36\%$ of the encoder attention at layers $10$, $12$ and $23$, and the
  top $1\%$ of patches receive $4\%$, $8\%$ and $25\%$.}
  \label{fig:qual-layers}
\end{figure}

Figure~\ref{fig:qual-layers} shows the attention of three encoder layers of InternVL3.5-8B and the
tokens retained when the saliency is read from each of them. The attention sinks intensify with
depth: at layer $23$ a few tokens absorb most of the attention, so the saliency carries little
information about the content and the retained tokens no longer cover what the task needs. They
spread over the background, the letters keep few of them and the model misreads the word. At layer
$10$, the layer the probe selects on VQA, the saliency follows the strokes of the letters, the
retained tokens sit on them and the model reads the word. Layer $12$, where the concentration sets
in, falls in between.

\paragraph{Failure Cases.}
\label{sec:appendix:qual-failure}

\begin{figure}[t]
  \centering
  \includegraphics[width=0.8\linewidth]{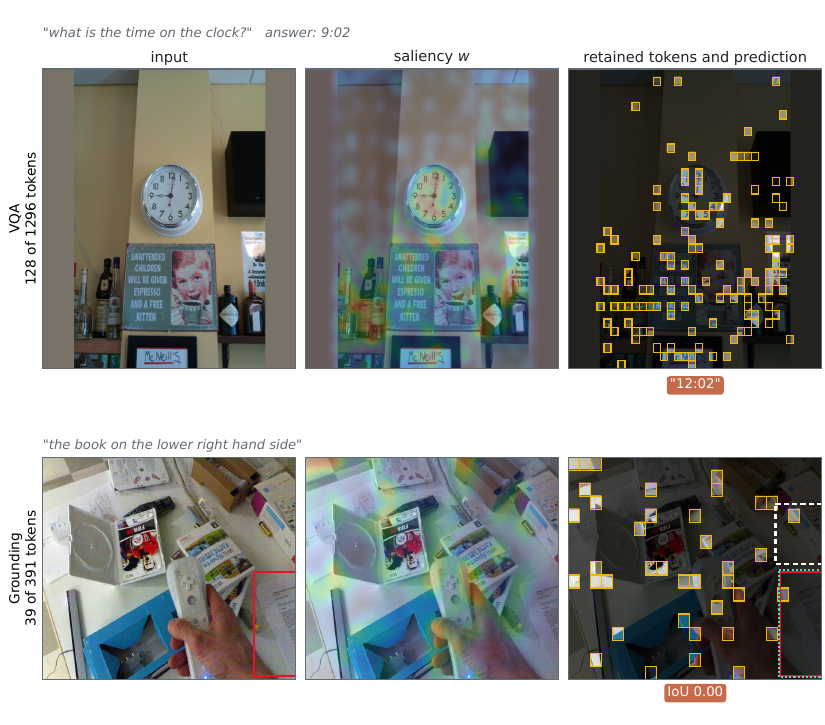}
  \caption{\textbf{Two failure cases} (Qwen2.5-VL-7B; the unpruned model answers both correctly).
  Left to right: the input, the saliency $w$, and the retained tokens. Top, VQA: the clock is
  retained and salient, yet the position of the hands is not resolved. Bottom, grounding: the
  ground truth is in red, our prediction is the white dashed box and the unpruned prediction the
  dotted one; the saliency misses the referred book, which keeps a single token.}
  \label{fig:qual-failure}
\end{figure}

Figure~\ref{fig:qual-failure} presents one case of each failure mode. The first is the loss of
fine detail under aggressive pruning: the target region is retained and attended, yet the detail
on which the answer depends is no longer resolved. In the VQA case the clock is retained, while
the position of its hands is not, and the model reports the wrong time. The second is the saliency
source, which does not always respond to the region the task requires. In the grounding case the
saliency is concentrated on the game boxes and assigns little weight to the referred book, so a
single token is retained there and the prediction falls elsewhere.

\end{document}